\documentclass[10pt,twocolumn,letterpaper]{article}

\usepackage{iccv}
\usepackage{times}
\usepackage{epsfig}
\usepackage{graphicx}
\usepackage{amsmath}
\usepackage{amssymb}
\usepackage{subcaption}
\usepackage{color}
\usepackage{booktabs}
\usepackage{mathastext}
\usepackage{multirow}
\usepackage{comment}
\usepackage{diagbox}
\usepackage{bbding}
\usepackage{appendix}
\usepackage{eso-pic}
\usepackage[normalem]{ulem}

\usepackage[pagebackref=true,breaklinks=true,letterpaper=true,colorlinks,bookmarks=false]{hyperref}

\iccvfinalcopy 

\def\iccvPaperID{1295} 

\ificcvfinal\fi

\begin{document}

\title{FemtoDet: An Object Detection Baseline for Energy Versus Performance Tradeoffs}

\vspace{-5pt}
\author{
Peng Tu$^{1}$, Xu Xie$^{1}$, GUO AI$^{1}$, Yuexiang Li $^{2}$, Yawen Huang$^{2*}$, Yefeng Zheng$^{2}$ \\
$^1$ MicroBT Inc. \ $^2$ Jarvis Lab, Tencent \\
}

\maketitle

\ificcvfinal\thispagestyle{empty}\fi

\let\thefootnote\relax\footnotetext{$^*$Corresponding author; First author's Email: yh.peng.tu@gmail.com}

\vspace{-5pt}
\begin{abstract}
   Efficient detectors for edge devices are often optimized for parameters or speed count metrics, which remain in weak correlation with the energy of detectors.
   However, some vision applications of convolutional neural networks, such as always-on surveillance cameras, are critical for energy constraints.  
   This paper aims to serve as a baseline by designing detectors to reach tradeoffs between energy and performance from two perspectives: 
   1) We extensively analyze various CNNs to identify low-energy architectures, including selecting activation functions, convolutions operators, and feature fusion structures on necks. These underappreciated details in past work seriously affect the energy consumption of detectors;
   2) To break through the dilemmatic energy-performance problem, we propose a balanced detector driven by energy using discovered low-energy components named \textit{FemtoDet}. 
   In addition to the novel construction, we improve FemtoDet by considering convolutions and training strategy optimizations.
   Specifically, we develop a new instance boundary enhancement (IBE) module for convolution optimization to overcome the contradiction between the limited capacity of CNNs and detection tasks in diverse spatial representations, and propose a recursive warm-restart (RecWR) for optimizing training strategy to escape the sub-optimization of light-weight detectors by considering the data shift produced in popular augmentations.
   As a result, FemtoDet with only 68.77k parameters achieves a competitive score of 46.3 AP50 on PASCAL VOC and 1.11 W $\&$ 64.47 FPS on Qualcomm Snapdragon 865 CPU platforms.
   Extensive experiments on COCO and TJU-DHD datasets indicate that the proposed method achieves competitive results in diverse scenes.
\end{abstract}

\vspace{-5pt}
\section{Introduction}
The deployment of efficient convolutional neural networks (CNNs) enabled immense progress \cite{liu2016ssd, redmon2016you, redmon2017yolo9000, redmon2018yolov3, bochkovskiy2020yolov4, ge2021yolox, xu2022pp} in vision detectors for edge devices, in which they consistently reduce parameters and speed counts for improving accuracy.
However, these metrics are not correlated well with the efficiency of the models in terms of energy.
Evaluation metrics, such as parameters, do not take into account the energy cost of models, resulting in a nontrivial effect on the energy cost of detectors. 
Compared with the same architecture, the parameters of models are positively correlated with their energy cost (shown in Table \ref{energy_convolution_operators}). 
However, in the case of equal model parameters, their energy consumption may be negatively correlated or even irrelevant to the model parameters (shown in Table \ref{energy_activation_functions}).
Considering that various activation functions, convolution operators, and feature fuse structures may not increase model parameters, but generate more energy costs.
Similarly, the speed count is also not well correlated with energy, as it can be optimized by the degree of parallelism.
These disconnections will leave customized efficient detectors unavailable, when they are deployed under severe energy constraints like the always-on surveillance cameras.


This paper aims to reduce the energy costs of efficient object detectors, while improving their performances by achieving energy versus performance tradeoffs. 
Specifically, these bottlenecks fall into several categories: 

1) \textit{Detector components with unknown energy}.
Most current object detection methods focus either on latency-oriented \cite{cai2021yolobile, shafiee2017fast, yu2021pp} or accuracy-oriented \cite{ge2021yolox, tian2019fcos, ren2015faster, sun2021sparse} works.
There are very limited works toward the energy cost of the detector components, which is the first obstacle to designing energy-performance-balanced detectors.
To identify energy-efficient components of detectors, we followed \cite{yang2019ecc} to benchmark their energy metrics from three types of structures, \textit{i.e.}, activation functions, convolution operators, and necks of detectors.
The following finding is acquired: 
Firstly, some activation functions have been widely used, because of their ability to improve models without adding more parameters.
They have increased costs, yet received very limited attention.
As shown in Table \ref{energy_activation_functions}, we set ReLU \cite{agarap2018deep} based models as a baseline.
When replacing the ReLU with the GELU \cite{hendrycks2016gaussian}, 4.40\% performance is gained, while 12.50\% energy cost is increased, and the corresponding factor of the mean energy versus the performance tradeoffs (Eq. \ref{mEPT}) drops 7.16\%. 
In addition, although large kernel convolutions \cite{liu2022convnet} can improve models (Table \ref{energy_convolution_operators}), the increased energy cost is unacceptable.
As seen in Table \ref{energy_convolution_operators}, the large kernel convolutions are equipped to enable the model to rise by around 1.87\% in performance, but with more energy costs (16.20\%) than in the small kernel size of convolutions (shown in the second row of Table \ref{energy_convolution_operators}). 
Finally, the standard FPN \cite{lin2017feature, liu2018path, xie2022latent,ghiasi2019fpn, tan2020efficientdet} in the detectors also causes significant energy costs.
We observe that this situation is happened due to multi-feature fusion paths following either bottom-to-up or top-to-down way, resulting in frequent data reading in memory for data coverage. 
However, it is shown that the up-down fusion between multi-features may not necessary.
As shown in Table \ref{energy_fpn}, we propose a simple SharedNeck for replacing FPN to reduce around 5.77\% energy costs, while obtaining 6.25\% performance gains.
SharedNeck uses an adaptive fusion between multi-features by learning convolution, instead of feature-by-feature bottom-to-up or top-to-down fusions, as operated in FPN. 
Based on these analyses, we further build upon a low-energy detector, named FemtoDet.
Surprisingly, FemtoDet just has 68.77k parameters and 1.1W power on the  platform.
2) \textit{Optimization for CNN} is another bottleneck, since obtaining a favorable detector is very challenging, especially for a small number of parameters.
Light-weight detectors are restricted by their limited capacities \cite{ding2022slimyolov4}, which will cause obfuscation on feature maps of the instances' boundaries of interesting objects, as shown in Figure \ref{rgb_baseline_femtodet} (b).
Ambiguous instance boundaries on the features may raise the risk of false detection in the models.
For this problem, we propose a novel instance boundary enhancement module (IBE), which emphasizes the potential of the object boundary information:
The parameter reuse mechanism is first applied to integrate local descriptors with convolution operation to make robust and diverse instance boundary feature representation.
But this operation would cause feature un-alignment between normal features and integrated features.
Then, in order to sufficiently exploit these two types of features, we design dual-normalization in IBE, as to re-align the features (Fig. \ref{unaligment_features} of the Appendix).
IBE employs a shared convolution yet an independent batch normalization layer for separate normalization is also adopted.
Finally, we add normal and integrated features together behind the dual-normalization layer.
IBE provides a novel form of parameter reuse to generate new local descriptors from shared convolution operators.
This method can capture object boundary information by integrating gradient cues around the learned convolution.
After trained, IBE modules can be folded into simple convolution operators, thus requiring no extra computations anymore;
In addition, data augmentation is a common approach for efficient detector training. 
Using well-designed strong augmentations is conducive to improving the generalization of models \cite{redmon2018yolov3, ge2021yolox}.
However, how to prevent data shifts between training and validation images has not been explored, while training data suffers strong augmentations.
It is found that the produced data shifts by strong augmentations would prevent light-weight detectors from moving toward global optimization.
A common view is that strong image augmentations can effectively encourage networks to learn diverse features \cite{gontijo2020tradeoffs}.
But for light-weight detectors, these diverse features cannot assist the models in making better generalizations in the validation set.
In other words, light-weight detectors are more susceptible to these ignored data shifts because of their limited capacities.
Furthermore, we propose an effective training strategy, namely recursive warm-restart (RecWR), to adapt these diverse features to improve the model's generalization.
RecWR works based on multi-stage training while gradually weakening data augmentation strength.
This method can help the limited-capacity detectors jump out of local optima under the assistance of high-dimension $\&$ diverse features.
The effectiveness of the IBE and RecWR has been evaluated on PASCAL VOC dataset.
And the experimental results show that IBE can improve FemtoDet $\sim$ $7.72\%$ performance, while incurring no extra parameter burdens, as compared to the original architecture; 
RecWR can improve FemtoDet $\sim$ $6.19\%$ performance by gradually weakening the data augmentation strength in multi-stage learning.
By jointly training FemtoDet with IBE and RecWR, our proposed methodology can go over the YOLOX \cite{ge2021yolox} by $51.34\%$ in performance when using the same-level parameters.

Something worth mentioning is that FemtoDet is specially designed for hierarchical intelligent chips to enable always-on alerts: the always-on low power, high-recall, and decent accuracy - it achieved 85.8 AR20 $\&$ 76.3 AP20 (Table \ref{results_ap_ar_20}) while performing pedestrian detection on TJU-DHD. 
In addition, FemtoDet is poor in detecting small objects but is excellent in detecting medium and large objects, which they are more interested in, with 88.8 AR20-m $\&$ 94.1 AP20-m / 95.3 AR20-l $\&$ 98.6 AP20-l (Table \ref{results_ap_ar_20}) on TJU-DHD. 
Information can be passed to other models after identifying possible objects of interest, and high-precision robust models are then initiated for accurate recognition. 
The always-on smart products have a wide range of applications, e.g., in-home monitoring or robots. 
Hence, the loose metric (e.g., AP50 or AP20) and data scenario with moderate difficulty (e.g., VOC) can well reflect the application ability of FemtoDet.
Further, the experiments on COCO datasets have verified that the proposed method is suitable for diverse scenarios resulting in competitive results.

\vspace{-5pt}
\section{Related Works}
\subsection{Object Detection}
Object detection is a classic computer vision task for identifying the category and location of object in pictures or videos.
The existing object detectors can be divided into two categories: two-stage detectors and one-stage detectors.
Two-stage detectors \cite{ren2015faster, sun2021sparse, he2017mask, lin2017feature} are anchor-based devices that generate region proposals from the image and then produce the final prediction boxes from these proposals.
Further, FPN \cite{lin2017feature} improves the two-stage detectors by fusing multi-level features.
Despite their higher accuracy compared to the one-stage detectors, two-stage detectors are still difficult to achieve low latency when deployed on edge devices.

Specifically, there are two types of one-stage detectors: anchor-based and anchor-free, which depends on anchor priors injecting into whole images to enable box regression. 
SSD \cite{liu2016ssd}, as a classical anchor-based one-stage detector, discretizes the output space of bounding boxes into a set of default anchors with different aspect ratios and scales per feature map location. 
This operation helps detecting tiny objects. 
Another typical anchor-based detector is the YOLO series \cite{redmon2017yolo9000, redmon2018yolov3, bochkovskiy2020yolov4}. 
YOLOv2 \cite{redmon2017yolo9000} explores bag of freebies to improve the performances of one-stage detectors. 
YOLOv3 \cite{redmon2018yolov3} proposes cross-scale features and novel nms (Non-Maximum Suppression) to obtain more confident predictions. 
YOLOv4 \cite{bochkovskiy2020yolov4} finds that the repeated gradient of network optimization aggravates detector latency, and thus designs Cross-Stage-Partial-connection modules to reduce the latency of detectors while maintaining detection performance. 
Anchor-free detectors \cite{redmon2016you, ge2021yolox, law2018cornernet, tian2019fcos} aim to eliminate the pre-defined set of anchor boxes. YOLO \cite{redmon2016you}, as an anchor-free one-stage detector, divides an image into multiple grids and predict boxes at grids near the object's center. 
CornerNet \cite{law2018cornernet} detects an object as a pair of key points (through the top-left corner and bottom-right corner of the bounding box).
A single convolutional network was leveraged in CornerNet to predict a heatmap for all top-left corners of instances of the same object category, a heatmap for all bottom-right corners, and an embedding vector for each detected corner.
Indeed, CornerNet is a new pipeline for object detection.
FCOS \cite{tian2019fcos} eliminates the anchor setting by proposing a fully convolutional one-stage object detector, which can solve object detection in a per-pixel prediction manner.
The anchor-free detectors solve the problem happened in anchor-based detectors, which reduce the memory cost and increase accuracy of the bounding box.

Both one-stage and two-stage object detection methods have obtained high performance on many challenging public datasets, \textit{i.e.}, COCO and TJU-DHD \cite{pang2020tju}.
These methods provide the accuracy-oriented detectors.
However, the critical problem for detectors serving edge devices is detection latency and its power.
In other words, detectors obtain higher performance in challenging scenes, which is not a necessary option for detectors deployed on edge devices.
For the detection latency, a lot of effort \cite{cai2021yolobile, shafiee2017fast, yu2021pp} have been devoted to achieving the balance between accuracy and efficiency. 
FastYOLO \cite{shafiee2017fast} is an optimized architecture extended from YOLOv2 with fewer parameters and $2\%$ performance drop, which makes FastYOLO run at an average of $\sim$18FPS on an Nvidia Jetson TX1 embedded system. 
YOLObite \cite{cai2021yolobile} focuses on pruning redundancy object detectors to be real-time by designing in a compression-compilation way. 
NanoDet uses ShuffleNetV2 \cite{ma2018shufflenet} as its backbone to make the model lighter, and further uses ATSS \cite{zhang2020bridging} and GFL \cite{li2020generalized} to improve its accuracy.
It is worth noting that NanoDet also reached $\sim$60FPS on an ARM CPU.
These light-weight detectors are latency-oriented.
Actually, neither accuracy-oriented nor latency-oriented detectors take their energy cost into account.
High energy cost detectors are unfriendly to devices deployed at edges.

This paper aims to develop a light-weight detector that can achieve tradeoffs between energy and performance.

\subsection{Energy-Oriented Convolutional Neural Networks}
Besides the manual design of deep neural networks, a great deal of work has been done to improve the effectiveness and efficiency of deep neural networks via network pruning \cite{yu2019autoslim}, quantization \cite{habi2020hmq, bhalgat2020lsq+}, architecture search \cite{guo2020single, xu2022analyzing}.
ECC \cite{yang2019ecc} proposes the end-to-end classifier training framework, which provides quantitative energy consumption guarantees by using weighted sparse projection and input masking.
Zhang $et~al.$ \cite{zhang2019distilling} specialized in video broadcast and leveraged distillation techniques to reduce memory consumption by approximating the video frame data.
MIME \cite{bhattacharjee2022mime} is an algorithm-hardware co-design approach that reuses the weight parameters of a trained parent task and learns task-specific threshold parameters for inferring on multiple child tasks.
However, the above works are all in service for image classification, and we can observe that fewer parameters represent lower energy costs in the same architecture.
Object detection on edge devices is still a challenging topic, considering that the conflict between the limited capacity of CNNs and diverse spatial representations.
This paper is the first work to provide a systematized solution for energy-oriented light-weight detectors.

\vspace{-5pt}
\section{FemtoDet}
This section will be divided into two subsections to describe how FemtoDet achieves energy versus performance tradeoffs. 
1) \textit{Benchmarks for Low-energy Detectors}: We provide benchmarks for designing low-energy detectors, including the exploring of activation functions, convolution operators, and neck of detectors. 
2) \textit{Energy-Oriented FemtoDet}: Based on the benchmarks for designing low-energy detectors, an energy-oriented light-weight detector named FemtoDet is provided, which stacks layers from depthwise separable convolutions (DSC), BN and ReLU. 
FemtoDet involves only 68.77k parameters and power of 1.11W on the Qualcomm Snapdragon 865 CPU platform.
Moreover, FemtoDet can be optimized with two designs:
Firstly, instance boundary enhancement (IBE) modules are used to improve the DSC in FemtoDet, overcoming the bottleneck of light-weight models' representation optimization (\textit{i.e.}, models will learn obfuscation features because of their limited capacity, as shown in Fig. \ref{rgb_baseline_femtodet} (b)).
Second, the recursive warm-restart (RecWR) training strategy is a multi-stage recursive warm-restart learning procedure, which can surmount data shifts produced from strong data augmentations. 

\vspace{-5pt}
\subsection{Benchmarks for Low-energy Detectors' Designing} 
\subsubsection{Evaluation Metrics}
Top1-Acc (top1 accuracy, for image classification), and AP (average precision, for object detection) are widely used measures to evaluate the performance of CNN.
In addition to the commonly used metrics $\mathrm{top1-Acc}$ and $\mathrm{AP}$, we propose the $\mathrm{Power}$ (energy costs) and $\mathrm{mEPT}$ (mean energy versus performance tradeoffs) to comprehensively evaluate the energy costs of the models and their ability to achieve energy versus performance tradeoffs. 
Specifically, $\mathrm{Power}$ and $\mathrm{mEPT}$ are represented by
\vspace{-5pt}
\begin{equation}
    \mathrm{mEPT} = (\frac{1}{N} \sum_{i}^{N} {P})/(\mathrm{Power}),
    \label{mEPT}
\end{equation}
\vspace{-5pt}
\begin{equation}
\mathrm{s.t.~~Power} = \frac{1}{{T}_{time}} \sum_{i}^{N} (\varepsilon(\mathbf{W}, {x}_{i})-\varepsilon(\hat{\mathbf{W}}, {x}_{i})),
\label{mE}
\end{equation}
where $\mathrm{{T}_{time}}$ indicates the total times required to evaluate ${N}$ images, $\varepsilon(\mathbf{W}, {x}_{i})$ represents the energy costs while models evaluate ${i}$-th image ${x}_{i}$ with trained parameters $\mathbf{W}$.
$\varepsilon(\hat{\mathbf{W}}, {x}_{i})$ is similar to $\varepsilon(\mathbf{W}, {x}_{i})$.
The difference is that we set a value of 1 for each layer of channels, where the model parameter is $\hat{\mathbf{W}}$. 
$\varepsilon(\hat{\mathbf{W}}, {x}_{i})$ indicates that the model is an empty state.
${P}$ denotes the model performance (top1-Acc for image classification while evaluating activation functions and convolution operators, AP50 for object detection while evaluating necks of detectors in Sec. \ref{sec311}) in image ${x}_{i}$. 
For the power metric, the value is expected to be as small as possible. For the mEPT metric, the result is expected as greater as possible.

\vspace{-5pt}
\subsubsection{Components to Be Evaluated}\label{sec311}
To identify low-energy components of detectors, we followed \cite{yang2019ecc} to benchmark their energy costs from three types of structures: activation functions, convolution operators and necks of detectors:

\textit{Activation Functions}. 
Rectified Linear Unit (ReLU) \cite{redmon2016you, liu2016ssd}, GELU \cite{hendrycks2016gaussian}, Swish \cite{ge2021yolox}, SiLU \cite{elfwing2018sigmoid}, and \textit{etc.} are widely used for object detection because they have fewer parameters and flops.
We calculate the energy costs of different activation functions in the same architecture to explore which activation function is more friendly for designing an energy-oriented detector.

\textit{Convolution Operators} can automatically learn filer weights while CNNs are trained to summarize helpful information, including horizontal, vertical, edged, diagonal, and \textit{etc.} features from images.
Vanilla convolutions (vanConv) \cite{krizhevsky2012imagenet}, depthwise separable convolutions (DSC) \cite{howard2017mobilenets} and their large kernel size versions \cite{liu2022convnet} are widely used. 
Ding \textit{et~al.} \cite{ding2022scaling} pointed out that when CNNs are constructed based on enough large convolution kernels, results can be comparable to the transformer \cite{liu2021swin}. 
Here, we set up energy consumption comparison experiments with different convolutions (vanCon and DSC) and multi-scale convolution kernel sizes (3$\times$3, and 5$\times$5) on different network structures (ResNet and MobileNetV2).
It is important that for each experiment case, the convolution operator type or kernel size would be the only variable.

\textit{Necks of Detectors}: 
Feature pyramid networks (FPN) are essential components for two-stage or one-stage detectors.
FPN-based necks of detectors fuse multiple low-resolution and high-resolution feature inputs for better representations, leading to a series of studies for designing manually complex fusion methods \cite{lin2017feature, liu2018path, xie2022latent}.
However, they only presented favorable results brought by FPN but ignored its energy costs.
We form a series of experiments on FemtoDet by using different neck types, \textit{e.g.} FPN \cite{lin2017feature}, PAN \cite{liu2018path}, and our proposed SharedNeck (details can be found in Sec. \ref{building_femtodet}) to study corresponding metrics (including energy costs, parameters, and detection precision on PASCAL VOC) changes.

All related experiments can be found in Sec. \ref{benchmarks}.
We can observe that DSC and ReLU are more energy friendly than vanCon and other activation functions.
In addition, by comparing with the necks of detectors such as FPN and PAN, the proposed SharedNeck is more suitable for light-weight detectors.
\vspace{-5pt}


\vspace{-5pt}
\subsection{Energy-Oriented FemtoDet}
\subsubsection{Building FemtoDet}
\begin{figure}[t]
\vspace{-5pt}
\centering
\setcounter{subfigure}{0}
\subfloat[FPN]{\includegraphics[width=4cm, height=1.6cm]{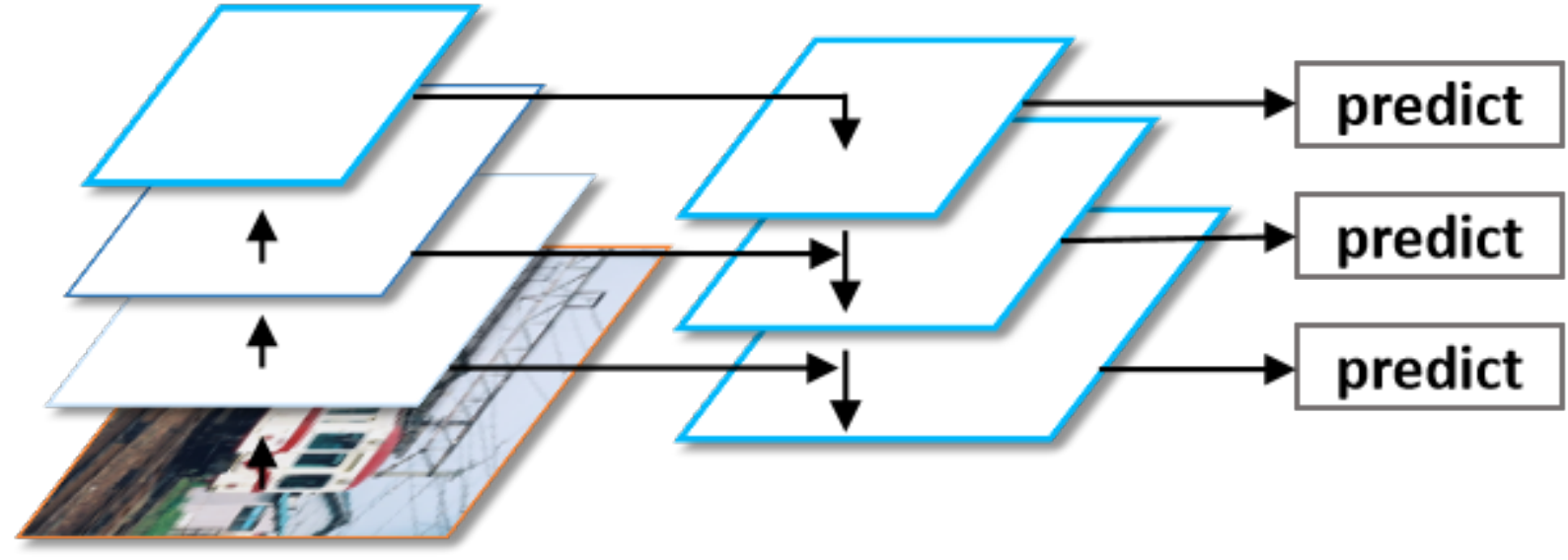}}\hspace{0.1mm}
\subfloat[SharedNeck]{\includegraphics[width=4cm, height=1.6cm]{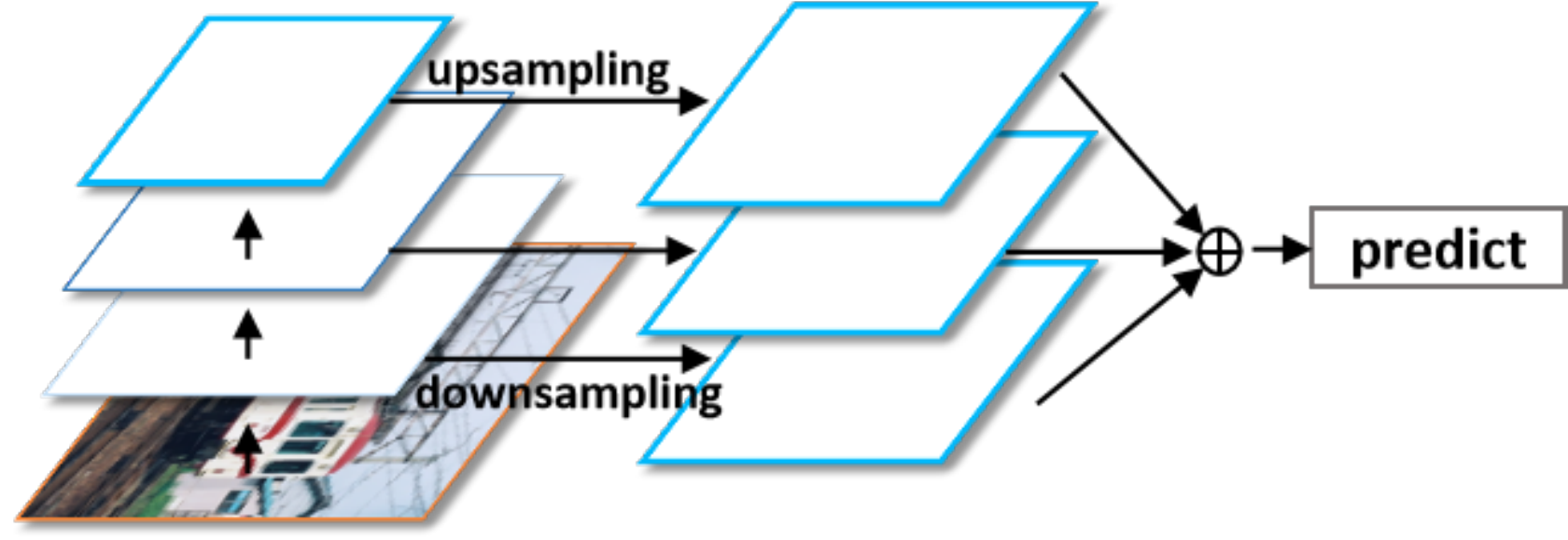}}\\\vspace{0.1mm}
\subfloat[PAN]{\includegraphics[width=6cm, height=1.6cm]{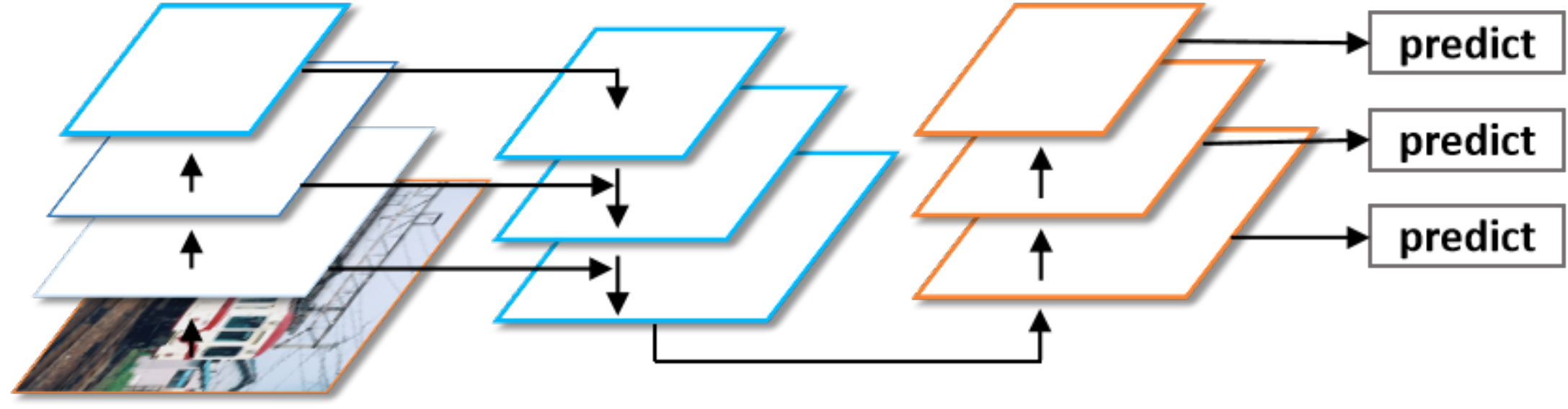}}\\\vspace{0.1mm}
\caption{
(a) FPN: A top-down multi-scale feature fusion architecture that outputs multiple predictions;
(b) SharedNeck: Simple adaptive multi-scale feature fusion architecture with single prediction, enabling tradeoffs between energy and performance;
(c) PAN \cite{liu2018path}: Exploring extra bottom-to-up multi-scale feature fusions with many predictions based on FPN.}
\label{sharedneck_and_others}
\vspace{-5pt}
\end{figure}

{\label{building_femtodet}}
\textit{Backbone}: The backbone of FemtoDet contains an initial full convolution layer with 8 filters.
We use ReLU as the non-linearity and BN as the batch normalization.
From the second layer beginning, all the convolution operators use DSC.
The reason is that we follow the results of our benchmarks (the results can be seen in Sec. \ref{benchmarks}) to select the energy friendly components.
The whole backbone is described in Appendix Table \ref{backbone_arch};
\textit{Neck}: We customize a neck for FemtoDet to achieve better tradeoffs between energy and performance.
In detail, SharedNeck first aligns the scale information of inputs from backbones, and then merges these alignment features with elements addition.
Finally, a DSC performs adaptive multi-scale information fusion between merged features.
The implementation details of SharedNeck, and how it differs from other necks are shown in Fig. \ref{sharedneck_and_others};
\textit{Head and Training Loss}: Here, we use the decouple head of YOLOX as the head of our proposed detector with the same training loss as YOLOX.
\vspace{-5pt}

\vspace{-5pt}
\subsubsection{Instance Boundary Enhancement Module}
\begin{figure}[t]
\centering
\subfloat{\includegraphics[width=2.7cm, height=2cm]{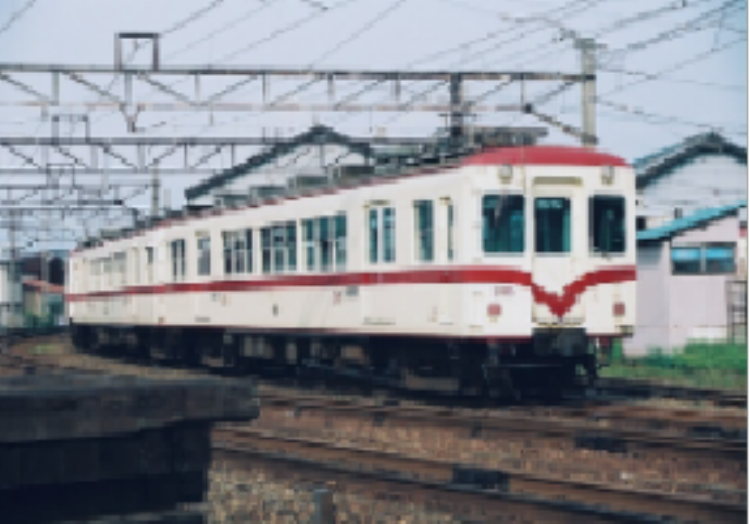}}\hspace{0.1mm}
\subfloat{\includegraphics[width=2.7cm, height=2cm]{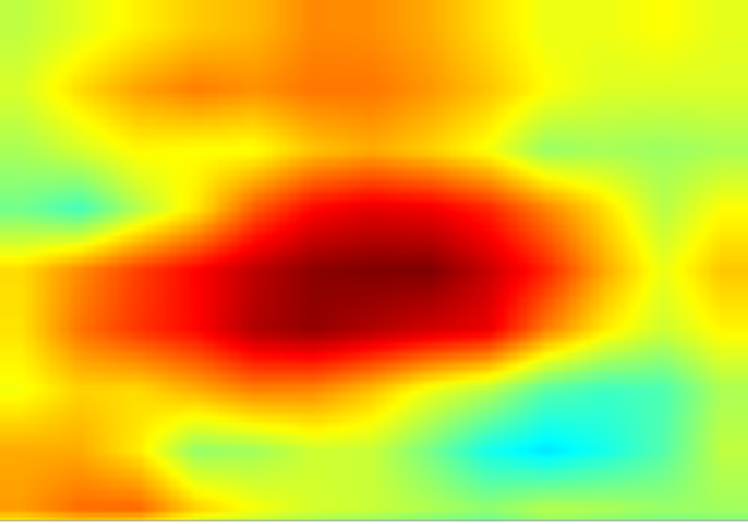}}\hspace{0.1mm}
\subfloat{\includegraphics[width=2.7cm, height=2cm]{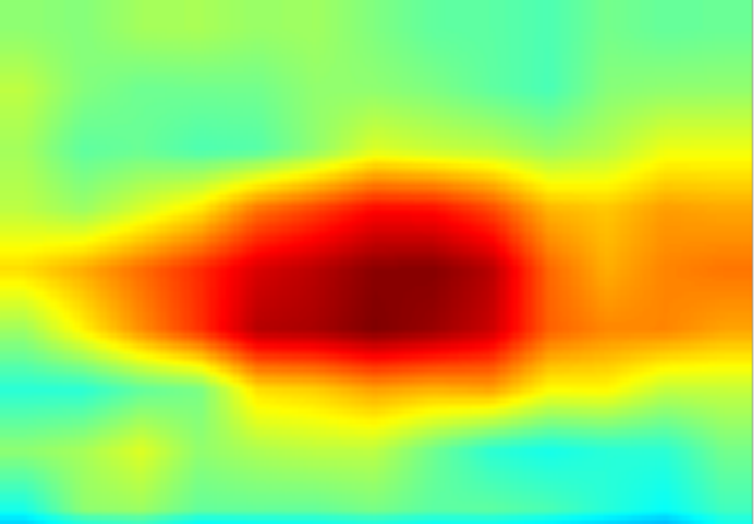}}\vspace{0.5mm}

\subfloat{\includegraphics[width=2.7cm, height=2cm]{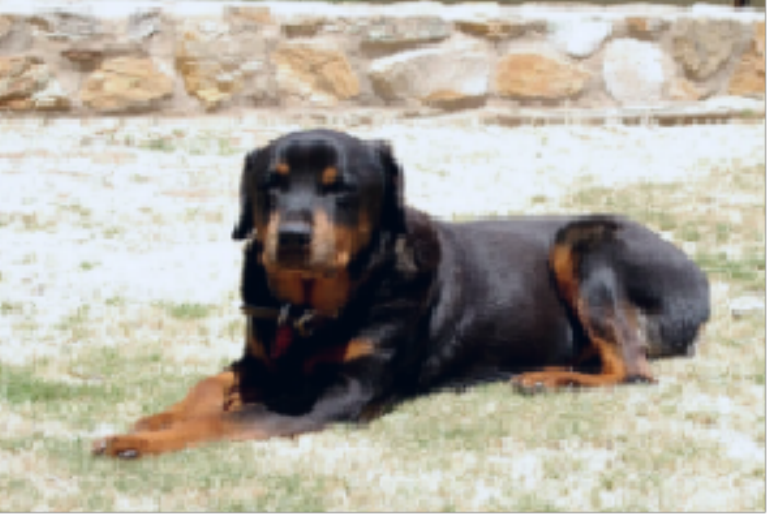}}\hspace{0.1mm}
\subfloat{\includegraphics[width=2.7cm, height=2cm]{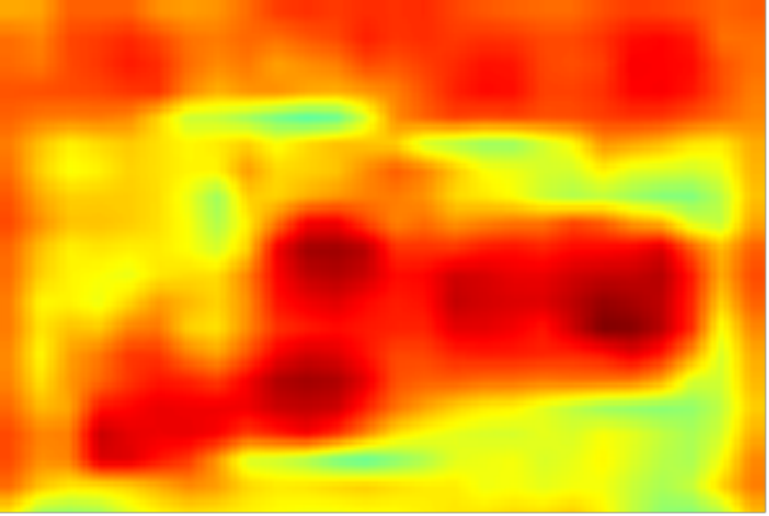}}\hspace{0.1mm}
\subfloat{\includegraphics[width=2.7cm, height=2cm]{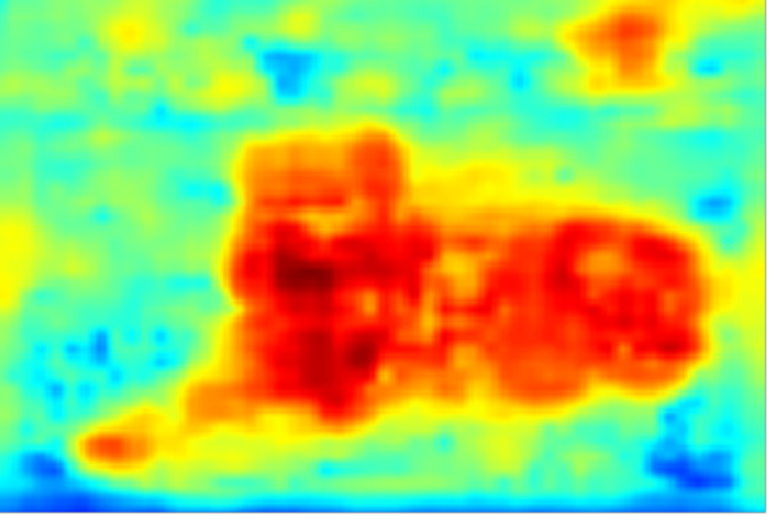}}\vspace{0.5mm}

\setcounter{subfigure}{0}
\subfloat[Inputs]{\includegraphics[width=2.7cm, height=2cm]{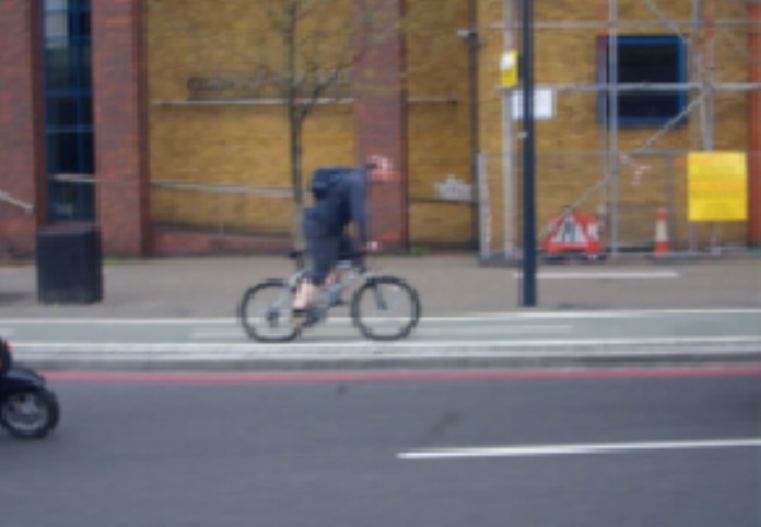}}\hspace{0.1mm}
\subfloat[${\mathrm{FemtoDet}}^{*}$]{\includegraphics[width=2.7cm, height=2cm]{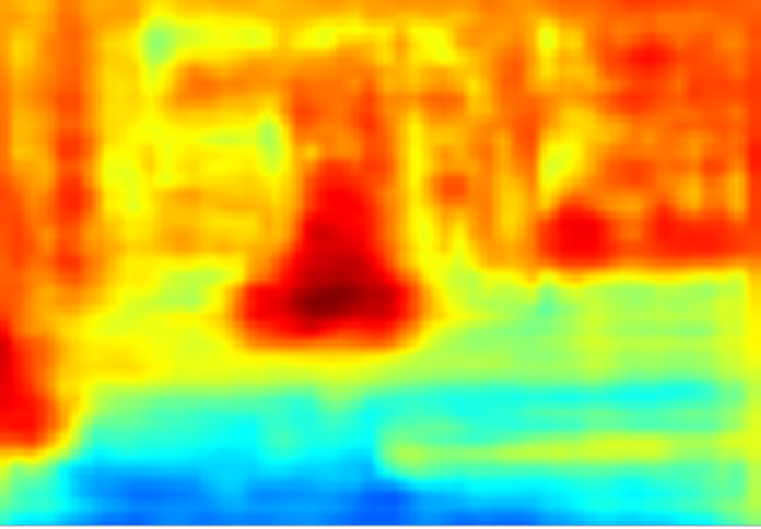}}\hspace{0.1mm}
\subfloat[$\mathrm{FemtoDet}$]{\includegraphics[width=2.7cm, height=2cm]{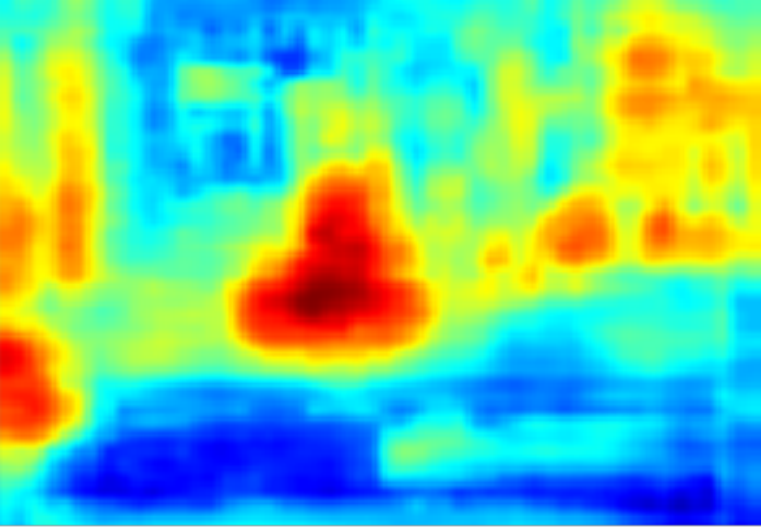}}\vspace{-1mm}
\caption{Feature visualization of the trained light-weight detector.
(a) Inputs: RGB images from PASCAL VOC;
(b) ${\mathrm{FemtoDet}}^{*}$: Trained FemtoDet detector consisting of pure depthwise separable convolution;
(c) FemtoDet: Trained FemtoDet detector, which was trained with the IBE modules.}
\label{rgb_baseline_femtodet}
\vspace{-5pt}
\end{figure}

Optimizing light-weight detectors is a notoriously challenging problem.
The reason for this is that limited by their representation, features learned by detectors are diffuse, as shown in Fig. \ref{rgb_baseline_femtodet} (b).
Instance boundary enhancement (IBE) modules are designed to improve the DSC in FemtoDet, thus overcoming the bottleneck of light-weight models' representation optimization.
IBE is similar to modules introduced in \cite{yu2019autoslim, su2021pixel, ying2022mocopnet}, except that our modules are designed for convolutional layers that are factorized into depthwise and pointwise layers.
We further introduce a dual-normalization mechanism, such that the IBE modules can be used for object detection, while the modules in \cite{yu2019autoslim, su2021pixel, ying2022mocopnet} just working for low-level tasks.
Based on the DSC block, the basic block of IBE includes 3$\times$3 depthwise convolution followed by 1$\times$1 convolution. 
IBE enhances DSC by designing a new local descriptor, semantic projector, and dual-normalization layer. 
Specifically, the 1$\times$1 local descriptor is a parameter reusing mechanism generated by the linear transformation of the integrating gradient cues around shared depthwise convolution.
Therefore, the object boundary information can be found in the local descriptors. 
Previously, the local descriptor is known as the difference convolution \cite{yu2019autoslim, su2021pixel, ying2022mocopnet} used for low-level tasks. 
Although the idea is easy to get that using these object boundary information to enhance the noisy feature representation (shown in Fig. \ref{rgb_baseline_femtodet} (b)) of the above standard operators, such as depthwise convolution, in practice, the original difference convolution architecture cannot serve for high-level semantic tasks. 
We observe that features obtained by the difference convolution are unaligned with the features obtained by standard convolution (the figure shown in Fig. \ref{unaligment_features} of the Appendix).
On the one hand, the difference convolution captures diverse information about object boundaries by integrating gradient cues around 3$\times$3 or with larger convolution.
On the other hand, high-level tasks encourage standard convolution to summarize the abstract semantic information of images.
That is the reason that the classical difference convolution cannot serve high-level tasks.
To simultaneously solve this problem, we propose a semantic projector and a dual-normalization layer. 
The semantic projector is the linear transformation of depthwise convolution to transfer operators related to semantic information extraction;
The dual-normalization layer consists of two independent batch normalization modules specifically designed for aligning the unaligned features.
After that, we incorporate the object boundary cues from the feature addition between the local descriptor and depthwise convolution to guide the model in learning an effective representation of the instance (refined results can be seen in Fig. \ref{rgb_baseline_femtodet} (c)).
The description of IBE modules is given in Fig. \ref{ibe} of the Appendix.
For the depthwise convolution of kernel size 3, $C_{in}$ denotes the input channel dimension, $C_{out}$ represents the output channel dimension, the weight matrix can be denoted as $\mathbf{W}_{conv} \in \mathbb{R}^{1 \times C_{in} \times 3 \times 3}$, and the bias is represented as $\mathbf{b} \in \mathbb{R}^{D}$.
The local descriptor of kernel size 1 can be generated from the depthwise convolution by reusing parameters.
Its weight matric is denoted as $\mathbf{W}_{des}$, which is integrating the gradient cues around $\mathbf{W}_{conv}$:
\begin{equation}
    \mathbf{W}_{des} = -\theta_{1} . \sum_{{p}_{n} \in \mathbf{R}} \mathbf{W}_{conv}({p}_{n}),
    \label{local_descriptor}
\end{equation}
where $\mathbf{W}_{des} \in \mathbb{R}^{C_{out} \times C_{in}}$, $\theta_{1} \in [0, 1], \in \mathbb{R}^{1 \times 1}$ is a learnable parameter projection factor. 
$\mathbf{R}$ represents the $3 \times 3$ convolution in $\mathbf{W}_{conv}$, and ${p}_{n}$ is the ${n}$-the weight value. 
Integrating the gradient cues around $\mathbf{W}_{conv}$ can help the local descriptors capture the object boundary information effectively. 
The semantic projector is generated from the depthwise convolution, while diverse semantic presentations can be obtained through a learnable linear transformation, shown as follows:
\begin{equation}
    \mathbf{W}_{pro} = \theta_{2} . \mathbf{W}_{conv}, \mathbf{W}_{pro} \in \mathbb{R}^{C_{out} \times C_{in} \times 3 \times 3},
    \label{proconv}
\end{equation}
where $\theta_{2} \in [0, 1], \mathbb{R}^{1 \times 1}$ is another learnable parameter as the projection factor. 
Subsequently, the IBE modules perform four steps to refine the feature representation under the guidance of object boundary information. 
1) inputs ${x}_{in}$ will be convolved with the obtained three convolution operators, and the corresponding results are denoted as ${x}_{22}, {x}_{21}$, and ${x}_{23}$, respectively;
2) elements addition are conducted between ${x}_{22}$ and ${x}_{21}$;
3) the dual-normalization layer enables feature distribution normalization on ${x}_{21}$ and ${x}_{23}$;
4) the two outputs from the dual-normalization layer can be added together;
As a result, the aforementioned added features are convolved through pointwise convolution as the final output of the IBE module.

In addition, we leverage the homogeneity and additivity of the convolutions to fold IBM modules into the simpler depthwise separable convolution in the inference stage without performance degradation.
The details can be seen in Appendix \ref{sec:folding_ibe}.

\subsection{Recursive Warm-restart Training Strategy}
Strong augmentations (SA) are widely used for object detection, and \cite{ge2021yolox, redmon2018yolov3} shows well-designed SA effectively improves detectors.
However, we find that well-designed SA does not always benefit light-weight detectors, because the current training strategies cannot sufficiently exploit diverse training representations to boost the generalization ability on real validation data.
For example, YOLOX \cite{ge2021yolox} points out MixUp reduces YOLOX-nano's performance by about 5$\%$.
We argue that the limited capacity detectors fit the diverse data generated by SA as much as possible during training, such that the modules have no surplus capacity to adjust learned features to boost the generalization ability of modules on real validation data.
In other words, SA produces diverse representations that are equal to the data shifts, thus breaking the generalization ability of the module.
We also propose an effective training strategy to adjust these diverse features to improve the generalization ability, namely recursive warm-restart (RecWR).

As shown in Fig. \ref{RecWR} of the Appendix, the whole training procedure can be divided into four stages. 
Starting from the 1-${st}$ stage to 4-${th}$ stage training, the intensity of image augmentations gradually decreases. Specifically, in the 1-${st}$ training stage, some SA types will be combined, such as MixUp, Mosaic, and RandomAffine.
Starting from the 2-${nd}$ training stage, the above SA types are gradually unloaded in each training stage, until the 4-${th}$ training stage.
For the last training stage, only the random flip and random scales are performed on the training data.
In addition, before starting each training stage, the waiting training detectors load the trained weights of the previous training stage as the initialization.

We can see that in Appendix \ref{sec:recwar_mixup}, after training FemtoDet with RecWR, MixUp also help such extremely small detectors obtain better performance.
In other words, RecWR takes advantage of diversity features learned from SA to get FemtoDet out of sub-optimization.

\vspace{-5pt}
\section{Experiments}
We used VOC, COCO, TJU, and ImageNet datasets to conduct experiments and validate the proposed method. 
Specifically, whole experiments can be divided into two sections, ${i.e.}$, identifying low-energy components of detectors mentioned in Sec. \ref{sec311}, and validating the effectiveness of our designed FemtoDet, where the $\mathrm{Power}$ metric was measured in GTX 3090.
 \begin{table}[h]
 \vspace{-5pt}
 \small
 \centering
 \caption{Comparisons of energy-related metrics in MobileNetV2 0.25 (MBV2) \cite{sandler2018mobilenetv2} used different activation functions.
 We trained a classifier on ImageNet to evaluate relevant metrics.}
 \begin{tabular}{c|c|c|c|c}
 \hline
    Activation & Param & Power & Top1 & \multirow{2}{*}{mEPT} \\
    Functions  & (M)   & (W) &  (Acc) & \\\hline
    ReLU \cite{agarap2018deep} & \multirow{5}{*}{1.48} & \textbf{5.04} & 45.22 & \textbf{8.97} \\
    GELU \cite{hendrycks2016gaussian} &   & 5.67 & 47.21 & 8.33 \\
    Swish \cite{ge2021yolox} &   & 5.47 & 47.45 & 8.67 \\
    HSwish \cite{yu2021pp} &   & 5.33 & \textbf{47.60} & 8.93 \\
    SiLU \cite{elfwing2018sigmoid} &   & 5.73 & 47.25 & 8.25 \\\hline\hline
 \end{tabular}
 \label{energy_activation_functions}
\vspace{-5pt}
 \end{table}
\subsection{Benchmark to Find Low-energy Components}{\label{benchmarks}}
\subsubsection{Activation Functions}{\label{Activation Functions}}
Table \ref{energy_activation_functions} shows the energy-related metric results using different activation functions under the same architecture (MobileNetV2 0.25).
We observe that the activation functions can significantly improve models without extra parameters (Param).
On the other hand, such activation functions incur unacceptable overhead on some metrics which are inconvenient to measure.
For example, Swish improves ReLU-based models performance by 4.93$\%$, but produces more than 8.53$\%$ more energy cost.
The mEPT metric also demonstrate that ReLU results in the best energy versus performance tradeoffs.
 
\begin{table}[h]
\vspace{-5pt}
\small
\centering
\caption{Comparisons of energy-related metrics in MobileNetV2 0.75 and ResNet18 \cite{he2016deep}. 
The two CNNs can present the two typical convolution operators: vanilla convolutions and depthwise separable convolutions.
We compare different convolution operators with different kernel sizes on the same convolution operators.
Here, $\clubsuit$, and $\diamondsuit$ indicate two kernel sizes of $3\times$3, and $5\times$5.
In this experiment, we also evaluate models by training them on ImageNet.}
\begin{tabular}{c|c|c|c|c|c}
\hline
    \multirow{2}{*}{CNNs} & Convolution & Param & Power & Top1  & \multirow{2}{*}{mEPT}\\
                          & Operators   & (M)   & (W)   & (Acc) & \\\hline
    \multirow{2}{*}{Res18} & $\mathrm{vanConv}^{\clubsuit}$ & 12.46 & 170.64 & 69.93 & 0.41 \\
    & $\mathrm{vanConv}^{\diamondsuit}$ & 31.99 & 198.28 & 71.24 & 0.36 \\\hline\hline
    
    
    \multirow{2}{*}{MBV2} & $\mathrm{DSC}^{\clubsuit}$ & 2.29 & 5.56 & 69.51 & 12.50 \\
    & $\mathrm{DSC}^{\diamondsuit}$ & 2.35 & 6.07 & 70.96 & 11.69 \\\hline\hline
    

\end{tabular}
\label{energy_convolution_operators}
\vspace{-5pt}
\end{table}

\vspace{-5pt}
\subsubsection{Convolution Operators}
This sub-section compares energy-related metrics while training an image classifier on the ImageNet dataset based on the different convolution operators (vanCon and DSC), and the different kernel sizes (3$\times$3, and 5$\times$5) on the same convolution operator.
Experimental results in Table \ref{energy_convolution_operators} shows that:
1) When CNNs were constructed with the same kernel size, mobilenetv2 generated much lower energy costs than resnet18, as shown in Table \ref{energy_convolution_operators} of $\mathrm{vanConv}^{\clubsuit}$ and $\mathrm{DSC}^{\clubsuit}$.
In other words, DSC is more energy friendly than vanConv while achieving similar performance;
2) When CNNs were constructed with different kernel sizes and the same types of operators, the smaller kernel generated much lower energy costs than the larger kernel, as shown in Table \ref{energy_convolution_operators} of $\mathrm{vanConv}^{\clubsuit}$ and $\mathrm{vanConv}^{\diamondsuit}$.
However, the performance gain from the large kernels is not so impressive.

\vspace{-5pt}
\subsubsection{Necks of Detectors}
Here we evaluated the effects of different necks (including FPN, PAN, and SharedNeck) in FemtoDet.
Experimental results are shown in Table \ref{energy_fpn}, and some observation are summarized as follow:
1) Although FPN enables FemtoDet to achieve preferable results, it not only consumes more energy but also has a large parameter overhead;
2) Due to the limited representation of light-weight models, the top-down and bottom-up feature fusion of PAN is poor;
3) SharedNeck achieves the best results in the metrics of parameter, energy cost, object detection performance, and factors of mean energy versus performance tradeoffs by performing adaptive feature fusion.

\begin{table}[htp]
\vspace{-5pt}
\small
\centering
\caption{Comparisons of energy-related metrics in FemtoDet with different necks: FPN, PAN, and our proposed SharedNeck, respectively.
The detector was trained on the PASCAL VOC.}
\begin{tabular}{c|c|c|c|c}
\hline
    \multirow{2}{*}{Necks of Detectors} & Param & Power & \multirow{2}{*}{AP50} & \multirow{2}{*}{mEPT}\\
                                        & (k)  & (W)   &        & \\\hline
                                    FPN & 176.31 & 8.31& 40.04  & 4.82 \\
                                    PAN & 79.83  & 7.97& 39.91  & 5.01 \\
                             SharedNeck & \textbf{69.77}  & \textbf{7.83} & \textbf{42.50}  & \textbf{5.43} \\\hline\hline
    
\end{tabular}
\label{energy_fpn}
\vspace{-5pt}
\end{table}
\begin{table}[h]
\vspace{-5pt}
\small
\centering
\caption{PASCAL VOC object detection results with YOLOX, NanoDet Plus, and FemtoDet.}
\begin{tabular}{c|c|c|c|c}
\hline
    \multirow{2}{*}{Methods} & Param & Power & \multirow{2}{*}{AP50} & \multirow{2}{*}{mEPT}\\
                               & (k)  & (W)   &        & \\\hline
    YOLOX & 70.39  & 10.42      & 30.60  & 2.94 \\
    NanoDet Plus   & 69.60  & 11.29     & 38.44   & 3.40 \\
    FemtoDet & \textbf{68.77}  & \textbf{7.83} & \textbf{46.31} &\textbf{5.91}  \\\hline\hline

    Methods & mAP    & mAP-s & mAP-m & mAP-l\\\hline
    YOLOX   & 13.60  & 3.00  & 5.10  & 16.60 \\
    NanoDet Plus     & 20.78 & \textbf{4.02}  & 8.07   & 25.48 \\
    FemtoDet & \textbf{22.90} & 1.10  & \textbf{8.40}  & \textbf{28.50} \\\hline\hline
    
\end{tabular}
\label{results_voc}
\vspace{-5pt}
\end{table}
\begin{table}[h]
\vspace{-5pt}
\small
\centering
\caption{Inference speed (FPS) and power of trained detectors (YOLOX, NanoDet Plus, and FemtoDet were trained on PASCAL VOC) on Qualcomm Snapdragon 865 CPU platforms.}
\begin{tabular}{c|c|c|c|c}
\hline
    \multirow{2}{*}{Methods} & \multirow{2}{*}{Param (k)} & \multirow{2}{*}{AP50} & \multicolumn{2}{c}{Inference}\\\cline{4-5}
          &        &     & Power (W) & FPS \\\hline
    YOLOX & 70.39  & 30.60  & 1.41 & 38.04 \\
    NanoDet Plus   & 69.60  & 38.44 &  1.37 & 47.80 \\
    FemtoDet & \textbf{68.77}  & \textbf{46.31} & \textbf{1.11} & \textbf{64.47}  \\\hline\hline
    
\end{tabular}
\label{results_voc_inference}
\vspace{-5pt}
\end{table}

\vspace{-5pt}
\subsection{Validating the Effectiveness of FemtoDet}
We verify the effectiveness of FemtoDet on three datasets: PASCAL VOC, COCO, and TJU-DHD, while resizing inputs to 640$\times$640 for training and resizing inputs to 416$\times$416 for validating.
Two datasets, PASCAL VOC and TJU-DHD two datasets are converted to the COCO data types for evaluation.
In addition, we extract campus data from the TJU-DHD dataset to evaluate the pedestrian detection performance of our proposed extremely light detector, ${i.e.}$, FemtoDet.
Considering extremely light-weight detectors are challenging to fit the complex COCO dataset with poor detection performance, we pull the corresponding results to Appendix \ref{sec:coco_results}.
Also, we compare the detection performance of FemtoDet with YOLOX, and NanoDet Plus \cite{=nanodet}, in which they are at the same parameter level to ensure fairness.
All the detector backbones are pretrained on the ImageNet for 100 epochs, and the metric of mEPT presented in this section is calculated between AP50 and Power.

\vspace{-5pt}
\subsubsection{Results on PASCAL VOC}
Experimental results are presented in Table \ref{results_voc}, which shows light-weight detectors with eight metrics: parameter (Param), energy cost (Power), AP50, factors of mean energy versus performance tradeoffs (mEPT), mAP, mAP of small size objects (mAP-s), mAP of medium size objects (mAP-m), and mAP of large size of objects (mAP-l) respectively.
We can see that although YOLOX has a large number of parameters, it is much lower than FemtoDet in many metrics. 
NanoDet Plus has the parameter comparable to that of FemtoDet, but only outperforming FemtoDet in the metrics of mAP-s.
We believe that for such extremely light-weight detectors, their performances on large-scale objects are the most critical.
FemtoDet also presented better results than YOLOX and NanoDet Plus on the metrics of mAP-l.
Although the metrics ${i.e.}$ mEPT, used to evaluate performance and energy balance, FemtoDet achieves the best balance compared to the other two models.

In addition, we evaluated the inference speed (FPS) and power of trained detectors on Qualcomm Snapdragon 865 CPU platforms.
As shown in Tab. \ref{results_voc_inference}, our FemtoDet also achieves the smallest energy costs and inference speed on edge devices.
\begin{table}[h]
\vspace{-5pt}
\small
\centering
\caption{TJU-DHD pedestrian detection results with YOLOX, and FemtoDet.}
\begin{tabular}{c|c|c|c|c}
\hline
    \multirow{2}{*}{Methods} & Param & Power & \multirow{2}{*}{AP50} & \multirow{2}{*}{mEPT} \\
                               & (k)  & (W)   &        & \\\hline
    YOLOX & \textbf{67.37}  & 9.85      & 59.40  & 6.03 \\
    FemtoDet & 67.54  & \textbf{6.92} & \textbf{66.90} &\textbf{9.67}  \\\hline\hline

    Methods & mAP    & mAP-s & mAP-m & mAP-l \\\hline
    YOLOX   & 29.00  & 11.50  & 37.60  & 54.20  \\
    FemtoDet & \textbf{33.10} & \textbf{17.60}  & \textbf{50.20}  & \textbf{61.50} \\\hline\hline
    
\end{tabular}
\label{results_tju}
\vspace{-5pt}
\end{table}
\begin{table}[h]
\vspace{-5pt}
\small
\centering
\caption{Loose metric on TJU-DHD pedestrian detection with YOLOX, and FemtoDet.}
\begin{tabular}{c|c|c|c|c}
\hline
    \multirow{2}{*}{Methods} & Param & Power & \multirow{2}{*}{AR20} & \multirow{2}{*}{AP20} \\
                               & (k)  & (W)   &        & \\\hline
    YOLOX & \textbf{67.37}  & 9.85      & \textbf{90.40}  & 42.10 \\
    FemtoDet & 67.54  & \textbf{6.92} & 85.80 &\textbf{76.30}  \\\hline\hline

    Methods & AR20-m    & AP20-m & AR20-l & AP20-l \\\hline
    YOLOX   & 83.70  & 34.80  & 96.10  & 55.60  \\
    FemtoDet & \textbf{94.10} & \textbf{88.80}  & \textbf{98.60}  & \textbf{95.30} \\\hline\hline
    
\end{tabular}
\label{results_ap_ar_20}
\vspace{-5pt}
\end{table}

\vspace{-5pt}
\subsubsection{Results on TJU-DHD}
Extremely light-weight detectors like FemtoDet have broad application prospects in surveillance scenarios.
Hence, we evaluate the performance of FemtoDet in pedestrian detection using the campus dataset of TJU-DHD, and show the performance of the detectors on the less stringent metric AP20.
Table \ref{results_tju} and \ref{results_ap_ar_20} demonstrate that FemtoDet is competent for pedestrian detection in common surveillance scenarios: 1) FemtoDet has lower energy costs in deployment, which is 29.74$\%$ lower than YOLOX;
2) FemtoDet obtains 76.3$\%$ AP20, while YOLOX with the same parameters only obtain 71.80$\%$ AP20.

\begin{table}[h]
\vspace{-5pt}
\small
\centering
\caption{Ablation study on the effectiveness of our IBE modules and RecWR training strategy.
Where FemtoDet* denotes FemtoDet consists of pure DSC without the enhancement of IBE, 300e and 1200e mean the detector training with the same data augmentation with 300 or 1200 epochs.
}
\begin{tabular}{c|c|c|c|c|c}
\hline
    \diagbox{}{} & IBE & RecWR & 300e & 1200e & AP50 \\\hline\hline
    \multirow{5}{*}{FemtoDet*} & \XSolidBrush & \XSolidBrush & \CheckmarkBold & \XSolidBrush & 42.50\\
      & \XSolidBrush    & \XSolidBrush & \XSolidBrush & \CheckmarkBold & 42.11 \\
      & \XSolidBrush    & \CheckmarkBold & \XSolidBrush & \XSolidBrush & 45.13 \\
      & \CheckmarkBold  & \XSolidBrush   & \CheckmarkBold & \XSolidBrush & 45.78 \\
      & \CheckmarkBold  & \CheckmarkBold  & \XSolidBrush & \XSolidBrush & 46.31 \\\hline\hline
    
\end{tabular}
\label{ablation_studies}
\vspace{-5pt}
\end{table}

\vspace{-5pt}
\subsubsection{Ablation Studies}
We conduct a series of ablation studies on PASCAL VOC to demonstrate the effectiveness and efficiency of our IBE modules and RecWR training strategy.
As shown in Table \ref{ablation_studies}:
1) The second and third rows present that by training with 300 and 1200 epochs for FemtoDet*, such a longer epochs do not yield better gains.
2) The fourth and fifth rows present FemtoDet* was trained with RecWR and 300 epochs with the enhancement of IBE, respectively.
Results show that independently equipping IBE and RecWR can improve model performance.
3) The six rows demonstrate that jointly using IBE and RecWR enables FemtoDet to achieve the best performance.

\section{Conclusion}
This paper presented a baseline to encourage the research on energy and performance balance detectors. 
Our experimental results clearly showed that boosting performance can also cause problem of growth in energy consumption.
Conversely, the simple components like ReLU are suitable for building energy-oriented detectors. 
In addition, we also proposed a novel IBE module and RecWR training strategy to overcome the optimization problem of extremely light-weight detectors.
Compared to other state-of-the-art methods under the same parameters setting, IBE and RecWR support these baselines achieving the best performance on VOC, COCO, and TJU-DHD datasets, while consuming the least energy.
In the future, we will continue improving the detectors on the balance of energy and performance.

\section*{Appendix}
\appendix
\section{Training Settings}

\section{Architecture Details of the Backbone}
Table \ref{backbone_arch} presents the details of FemtoDet's backbone, which consists of 1 vanilla convolution and 13 depthwise separable convolutions (DSC).

\begin{table}[htp]
\vspace{-5pt}
\centering
\begin{tabular}{c|c|c|c|c|c}
\toprule[0.2em]
Input & Operator                           & $t$& $c$ & $n$ & $s$ \\
\toprule[0.2em]
$640^2\times3$ &    vanConv                  &  - &  8 & 1 & 2\\
$320^2\times8$ &   DSC    &  1 & 8   & 1 & 1 \\
$320^2\times8$ &   DSC     &  4 & 8   & 2 & 2 \\
$160^2\times8$ &    DSC     &  4 & 8   & 2 & 2 \\
$80^2\times8 $ &    DSC     &  4 & 16   & 3 & 2 \\
$40^2\times16$ &    DSC     &  4 & 24   & 2 & 1 \\
$40^2\times24$ &    DSC     &  4 & 40  & 2 & 2 \\
$20^2\times40$ &    DSC     &  4 & 80  & 1 & 1 \\
$20^2\times80$ &    DSC     &  - & -  & - & - \\
\toprule[0.2em]
\end{tabular}
\caption {
    \mbox{The backbones of FemtoDet}: Each line describes a sequence of 1 or more DSC layers, repeated $n$ times.
    Each layer has $c$ output channels.
    The first layer of each sequence uses stride $s$ and all the rest uses stride $1$.
    All spatial convolutions use $3\times 3$ kernels. The expansion factor $t$ is always applied to the input size.
}
\label{backbone_arch}
\vspace{-5pt}
\end{table}
\vspace{-5pt}

\begin{figure}[htp]
\subfloat{\includegraphics[width=7.6cm, height=3.4cm]{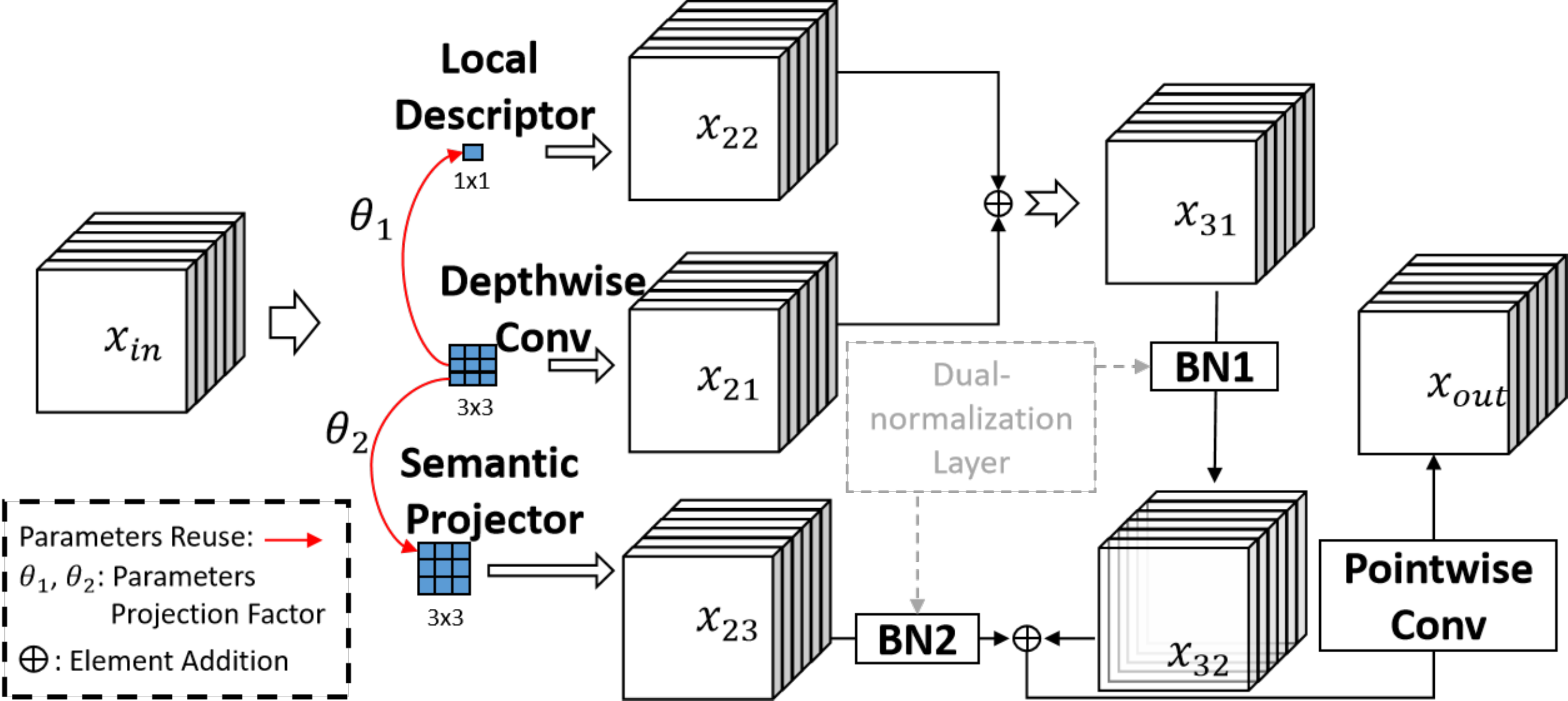}}\hspace{0.1mm}
\caption{
Overview of our proposed instance boundary enhancement (IBE) module.}
\label{ibe}
\vspace{-5pt}
\end{figure}

\begin{figure}[t]
\centering
\setcounter{subfigure}{0}
\subfloat[Before]{\includegraphics[width=4cm, height=3.2cm]{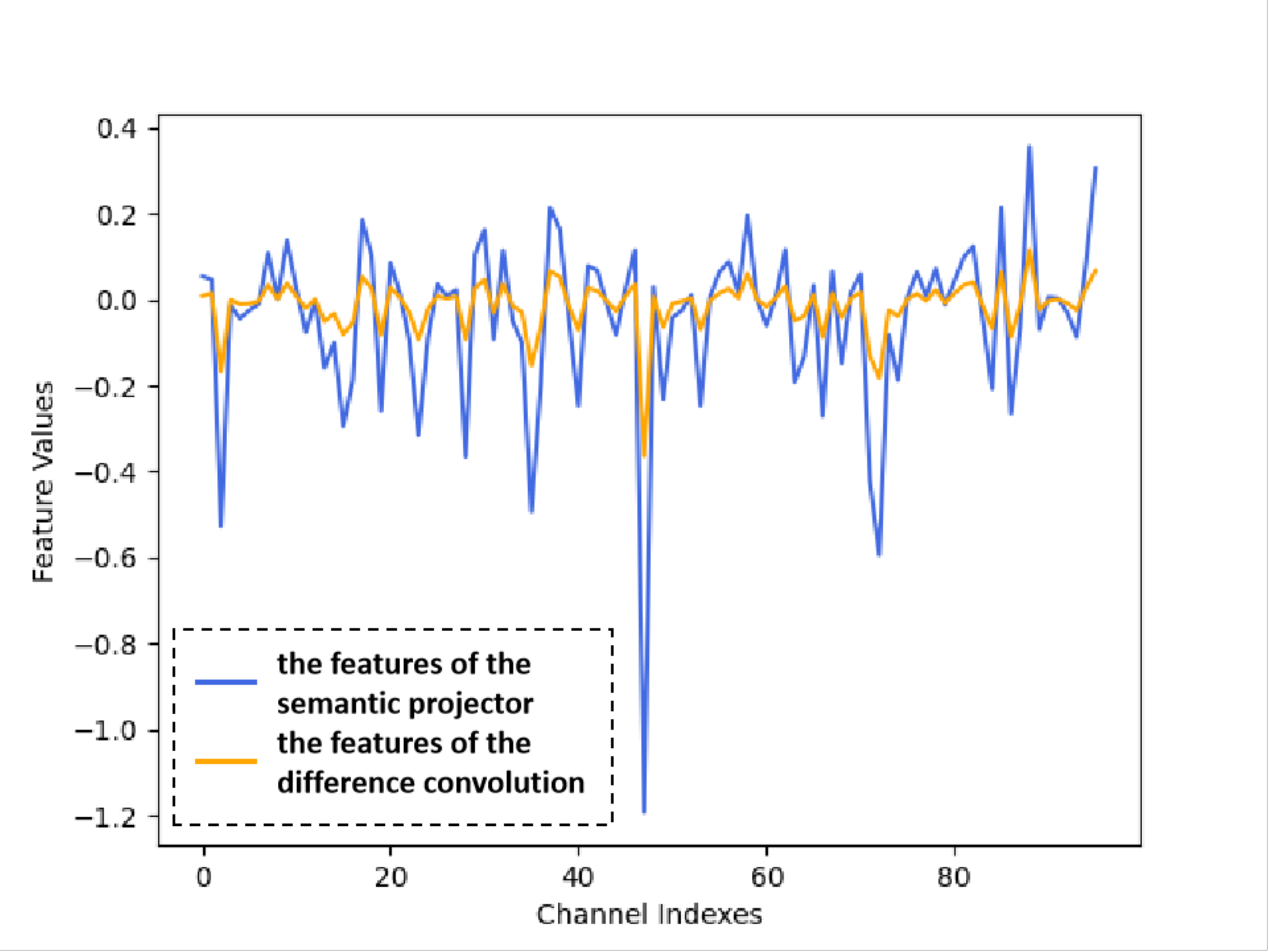}}\hspace{0.1mm}
\subfloat[After]{\includegraphics[width=4cm, height=3.2cm]{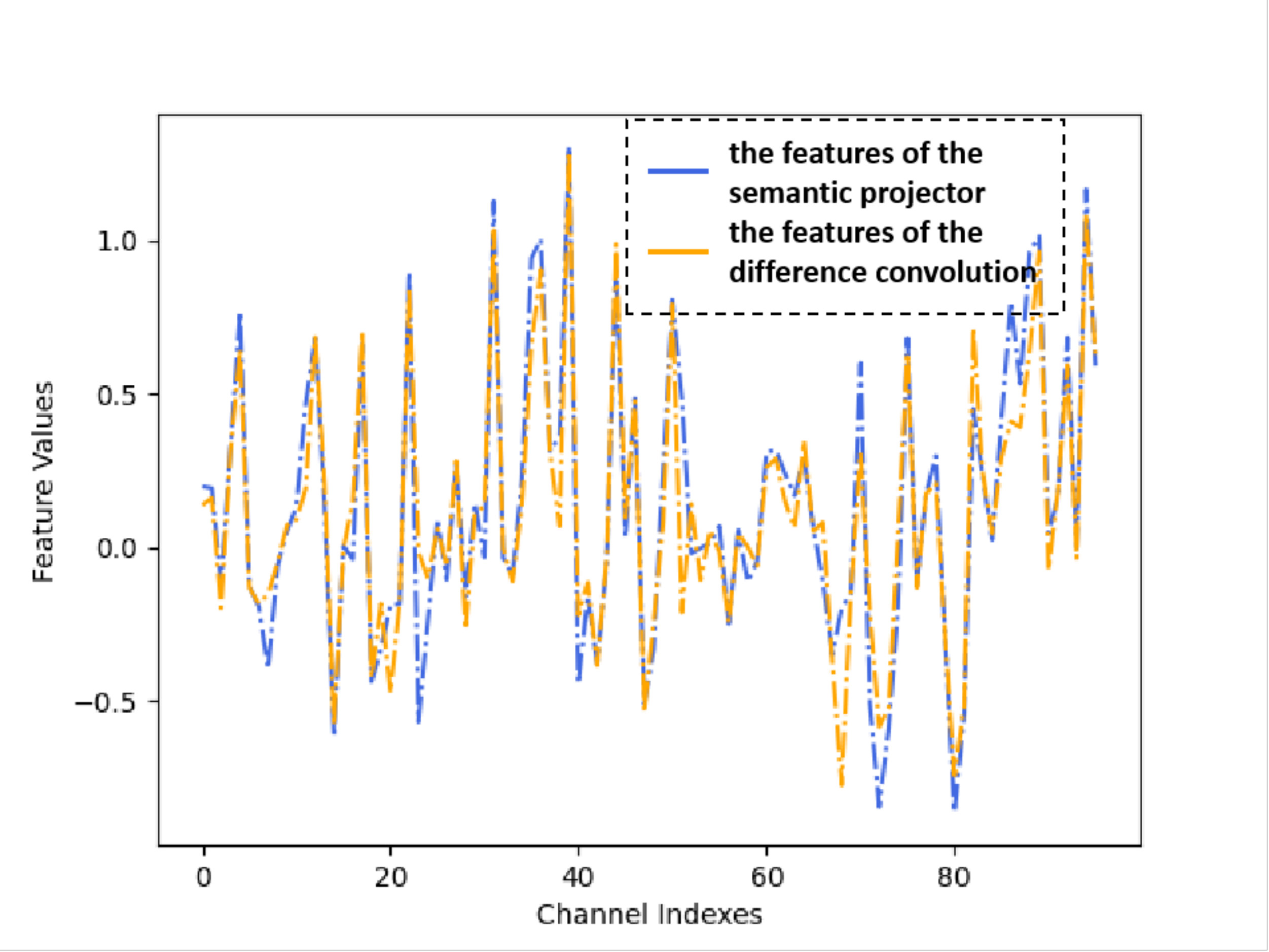}}\hspace{0.1mm}
\caption{We provided the comparison of activation values on an IBE module between the features from difference convolution and semantic projector.
The horizontal axis represents the channel indexes of features, and the vertical axis represents the values of the corresponding feature channels.
(a) Before: Feature distribution before the dual-normalization layer indicates that they are unaligned in the same space;
(b) After: Feature distribution after the dual-normalization layer.
It is obvious the dual-normalization layer effectively aligns two types of features.}
\label{unaligment_features}
\end{figure}

\begin{figure}[t]
\centering
\subfloat{\includegraphics[width=7.6cm, height=5.8cm]{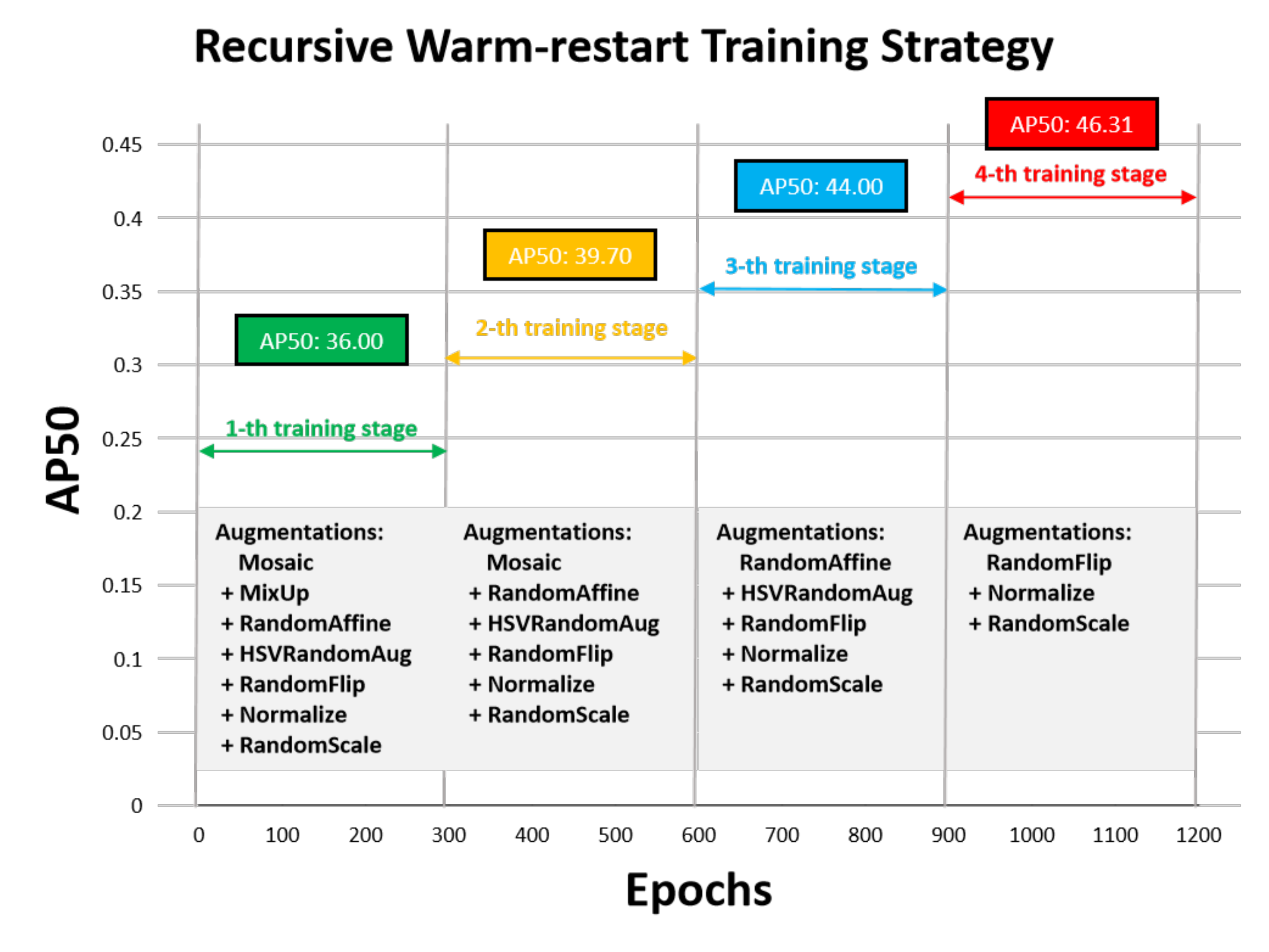}}\hspace{0.1mm}
\caption{
Four-stage training with recursive warm-restart (RecWR): From the first stage to the last stage of training, the augmentation strength of each stage training images is gradually weakening.
Before each training stage begins, the waiting training detectors will load the trained weights of the previous training stage as the model initialization.
The horizontal axis represents the number of training epochs, and the vertical axis represents the performance of the trained model on the PASCAL VOC validation set.
We can see that RecWR pushing detectors towards better performance.}
\label{RecWR}
\end{figure}

\section{Folding IBE to be DSC When Inferencing}\label{sec:folding_ibe}
According to the overview of IBE shown in Fig. \ref{ibe}, we represent the depthwise convolution, pointwise convolution, local descriptor, and semantic projector as $\mathrm{Conv_{1}}$, $\mathrm{Conv_{4}}$, $\mathrm{Conv_{2}}$, and $\mathrm{Conv_{3}}$ respectively.  
The output features of ${x}_{out}$ can be obtained by ${x}_{out}=\mathrm{Conv_{4}}(\mathrm{BN1}({x}_{31}) + \mathrm{BN2}({x}_{23}))$, where ${x}_{23}=\mathrm{Conv_{3}}({x}_{in})$, ${x}_{31}={x}_{21}-\mathrm{Sigmoid(\theta)}*{x}_{22}=\mathrm{Conv_{1}}({x}_{in})-\mathrm{Sigmoid(\theta)}*\mathrm{Conv_{2}}({x}_{in})$.
According to the homogeneity and additivity of the convolutions \cite{ding2019acnet}, we then show the process of folding all above complex operations into the depthwise convolution:

1) \textbf{Merging the $\mathrm{BN2}$ and $\mathrm{Conv_{3}}$}.
First, we express the weight matrix and bias of the $\mathrm{Conv_{3}}$ as $\omega$ and ${b}$, the variate of the $\mathrm{BN_{2}}$ as $\gamma$, $\sigma$, $\epsilon$ $\mu$ and $\beta$.
Then, we have ${x}_{23}=\mathrm{Conv_{3}}({x}_{in})=\omega * {x}_{in} + {b}$, $\mathrm{BN_{2}}({x}_{23})=\gamma*({x}_{23}-\mu)/\sqrt{\sigma^{2}+\epsilon} + \beta=\gamma*(\omega * {x}_{in} + {b}-\mu)/\sqrt{\sigma^{2}+\epsilon} + \beta={x}_{in} * (\gamma * \omega)/\sqrt{\sigma^{2} + \epsilon} + \beta + \gamma * ({b} - \mu)/(\sqrt{\sigma^{2} + \epsilon})$.
Finally, we obtain a new convolution $\hat{\mathrm{Conv}_{3}}$, which the weight matrix and bias of which is $\hat{\omega}=(\gamma * \omega)/\sqrt{\sigma^{2} + \epsilon}$, $\hat{b}=\beta + \gamma * ({b} - \mu)/(\sqrt{\sigma^{2} + \epsilon})$.
In other words, the two-step operation $\mathrm{{BN}_2}(\mathrm{Conv_3}({x}_{in}))$ is equivalent to the one-step operation $\mathrm{\hat{Conv_3}}({x}_{in}))$;

2) \textbf{Fold $\mathrm{Conv_{1}}({x}_{in})-\mathrm{Sigmoid(\theta)}*\mathrm{Conv_{2}}({x}_{in})$ to once convolution operation.}
First, based on the homogeneity of the convolutions, we can convert the multiplication between the constants and features to that between constants and convolution operators.
That means $\mathrm{Sigmoid(\theta)} * \mathrm{Conv_{2}}({x}_{in})=\mathrm{Sigmoid(\theta)} * (\omega * {x}_{in} + {b})=(\mathrm{Sigmoid(\theta)} * \omega) * {x}_{in} + \mathrm{Sigmoid(\theta)} * {b}=\mathrm{\hat{Conv_{2}}}({x}_{in})$.
And the weight matrix and the bias of the new convolution are $\mathrm{Sigmoid(\theta)} * \omega$ and $\mathrm{Sigmoid(\theta)} * {b}$, respectively.
Second, according to the additivity of the convolutions, we can convert the addition between features to that between convolution operators.
The $\mathrm{Conv_{1}}({x}_{in})-\mathrm{Sigmoid(\theta)}*\mathrm{Conv_{2}}({x}_{in})$ can be re-write as $\mathrm{Conv_{1}}({x}_{in})-\mathrm{Sigmoid(\theta)}*\mathrm{Conv_{2}}({x}_{in})=\mathrm{Conv_{1}}({x}_{in})-\hat{\mathrm{Conv_{2}}}({x}_{in})=(\omega_{1} * {x}_{in} + {b}_{1}) - (\hat{\omega_{2}} * {x}_{in} + \hat{{b}_{2}})=(\omega_{1}-\hat{\omega_{2}}) * {x}_{in} + {b}_{1} - \hat{{b}_{2}}=\hat{\mathrm{Conv}}({x}_{in})$.
Where $\omega_{1}$ and ${b}_{1}$ denotes the weight matrix and bias of $\mathrm{Conv_{1}}$, $\hat{\omega_{2}}$ and $\hat{{b}_{2}}$ is the weight matrix and bias of $\mathrm{Conv_{1}}$, and ($\omega_{1}-\hat{\omega_{2}}$) and (${b}_{1} - \hat{{b}_{2}}$) indicates the weight matrix and bias of $\hat{\mathrm{Conv}}$;

3) Similar to 1), $\mathrm{BN_{1}}(\mathrm{Conv_{1}}({x}_{in})-\mathrm{Sigmoid(\theta)}*\mathrm{Conv_{2}}({x}_{in}))=\mathrm{BN_{1}}(\hat{\mathrm{Conv}}({x}_{in}))$ can be merged into a new convolution as $\hat{\mathrm{Conv}}^{'}({x}_{in})=\mathrm{BN_{1}}(\hat{\mathrm{Conv}}({x}_{in}))$;
Like 2), $\mathrm{BN1}({x}_{31}) + \mathrm{BN2}({x}_{23})=\mathrm{BN1}(\hat{\mathrm{Conv}}({x}_{in})) + \hat{\mathrm{Conv_{3}}({x}_{in})}=\hat{\mathrm{Conv}}^{'}({x}_{in})+\hat{\mathrm{Conv_{3}}({x}_{in})}$ can be folded to be a single convolution operator.
After all the above operations, we can fold IBE to be DSC when inferencing.

\section{Explore the Impact of MixUp on Training FemtoDet}\label{sec:recwar_mixup}
\begin{table}[h]
\vspace{-5pt}
\small
\centering
\caption{In the ablation study on the impact of MixUp on training FemtoDet, where the RecWR is the original four-stage training strategy; $\mathrm{RecWR_{1}}$ uses the last three-stage data augmentations (without MixUp) to train FemtoDet; 300e means to train FemtoDet with 300 epochs on the second stage data augmentations (the same setting as when training YOLOX-tiny) of RecWR; $\mathrm{300e_{1}}$ denotes training FemtoDet with 300 epochs on the first stage data augmentations (with MixUp) of RecWR.
}
\begin{tabular}{c|c|c|c|c|c}
\hline
    \diagbox{}{} & RecWR & $\mathrm{RecWR_{1}}$ & 300e & $\mathrm{300e_{1}}$ & AP50 \\\hline\hline
    \multirow{5}{*}{FemtoDet} & \CheckmarkBold & \XSolidBrush & \XSolidBrush & \XSolidBrush & 46.31\\
      & \XSolidBrush    & \CheckmarkBold & \XSolidBrush & \XSolidBrush & 45.79 \\
      & \XSolidBrush    & \XSolidBrush & \CheckmarkBold & \XSolidBrush & 42.50 \\
      & \XSolidBrush  & \XSolidBrush   & \XSolidBrush & \CheckmarkBold & 36.00 \\\hline\hline

\end{tabular}
\label{exp_mixup}
\vspace{-5pt}
\end{table}

Table \ref{exp_mixup} shows the impact of MixUp \cite{zhang2017mixup} on training FemtoDet:
1) The second and third rows present the process of training FemtoDet with RecWR and $\mathrm{RecWR}_{1}$, demonstrating that might MixUp can help light-weight detectors achieve better performance when combined with RecWR;
2) The fourth and fifth rows show that MixUp hurts the light-weight detectors in the standard training strategy, which is consistent with the given by YOLOX;
3) Combining 1) and 2), we conclude that MixUp can only improve the light-weight detectors if it is used for light-weight detectors training under the RecWR.
In other words, RecWR takes advantage of diversity features learned from MixUp to get light-weight detectors out of sub-optimization.

\section{Results on COCO}\label{sec:coco_results}
\begin{table}[h]
    \vspace{-1.0mm}
    \small
    \centering
    \caption{COCO object detection results with YOLOX, and FemtoDet.}
    \begin{tabular}{c|c|c|c|c}
    \hline
        \multirow{2}{*}{Methods} & Param & Power & \multirow{2}{*}{AP50} & \multirow{2}{*}{mEPT}\\
                                   & (k)  & (W)   &        & \\\hline
        YOLOX & 79.93  & 11.58      & 9.50  & 0.82 \\
        NanoDet Plus   & \textbf{72.66}  & 12.80     & 11.79 & 0.92 \\
        FemtoDet & 72.67 & \textbf{8.28} & \textbf{12.60} &\textbf{1.52}  \\\hline\hline
    
        Methods & mAP    & mAP-s & mAP-m & mAP-l\\\hline
        YOLOX   & 4.50  & 0.70  & 3.70  & 7.90 \\
        NanoDet Plus   & \textbf{6.54}  & \textbf{1.72} & 5.75 & 10.39 \\
        FemtoDet & 6.20 & 0.70  & \textbf{6.00}  & \textbf{11.00} \\\hline\hline
        
    \end{tabular}
    \label{results_coco}
    \vspace{-1.0mm}
    \end{table}
COCO is another widely used object detection dataset with much greater, data complexity than PASCAL VOC.
This dataset contains 80 categories of around 160K images collected from the web.
We train detectors on the train2017 with 118K images and validate detectors on the val2017 with 41K images.
Table \ref{results_coco} compares the performance of YOLOX and FemtoDet on COCO.
Results on the metrics of detection seem terrible.
The reason is that detectors are too light-weight to accommodate such complex data.
Even so, FemtoDet performs better on COCO than YOLOX.

\section{Why is the IBE module able to capture the object boundary information?}
Classical edge detection \cite{torre1986edge, ziou1998edge} identifies sharp image brightness changes such as discontinuities in intensity, colour, or texture; image gradients or derivatives information are the first choices to extract such information.
Two perspectives enable IBE to capture the object boundary information:
1) First, the local descriptor is established to explore image gradient cues by integrating gradients around $3\times3$ depthwise convolution;
2) Second, the different convolutions in IBE based on gradient computing are used to encode important gradient information for edge detection by explicitly calculating pixel differences.

\begin{figure}[htp]
\centering
\subfloat{\includegraphics[width=7.6cm, height=5.5cm]{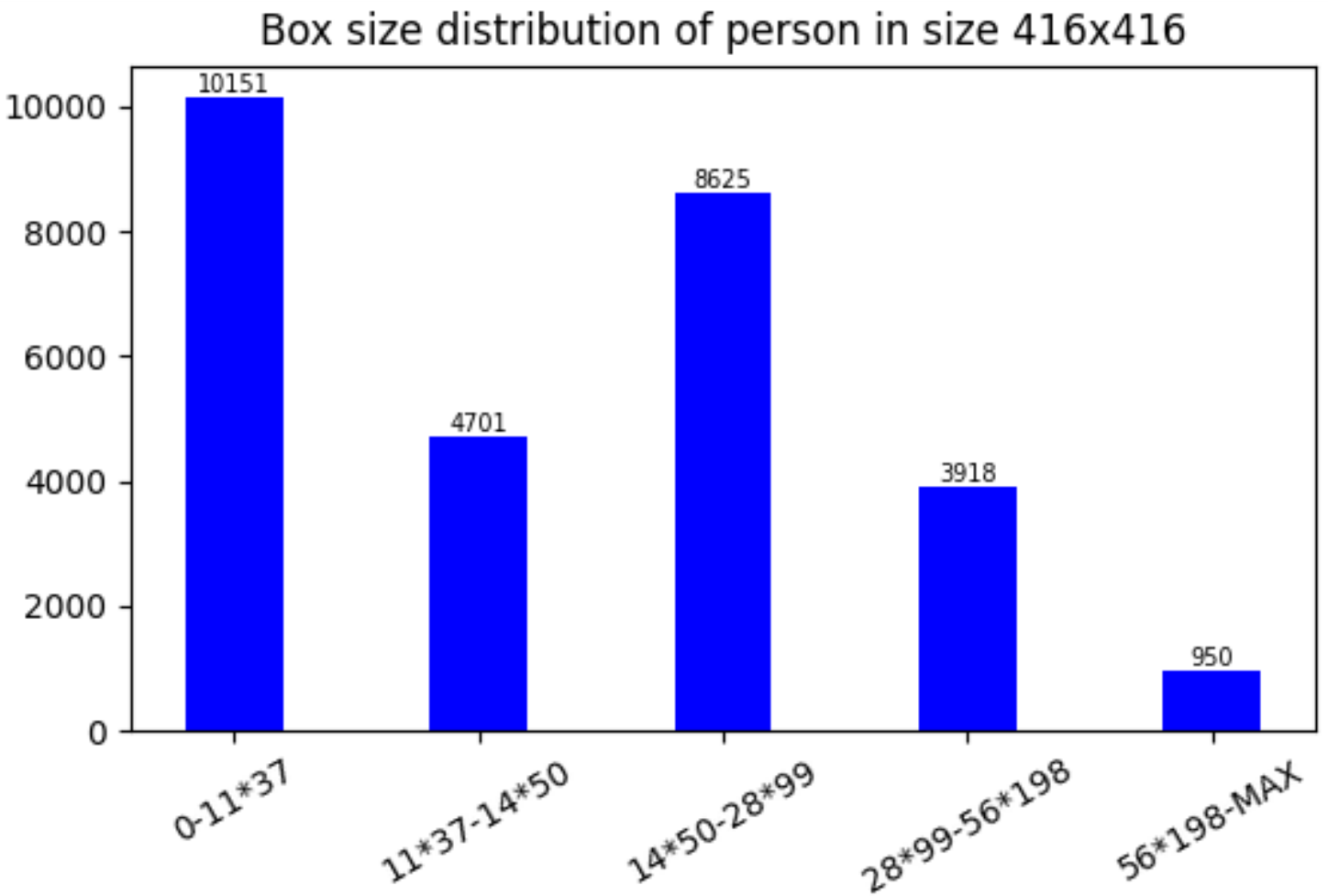}}\hspace{0.1mm}
\caption{The box size distribution of pedestrians in input size of 416$\times$416; 
It can be seen that the pedestrian are mostly small objects (0$\sim$11$\times$37), accounting for 35.81$\%$ of the total.
}
\label{person_box_distribution}
\vspace{-5pt}
\end{figure}
\begin{figure*}[htp]
\centering
\subfloat{\includegraphics[width=5.6cm, height=4.5cm]{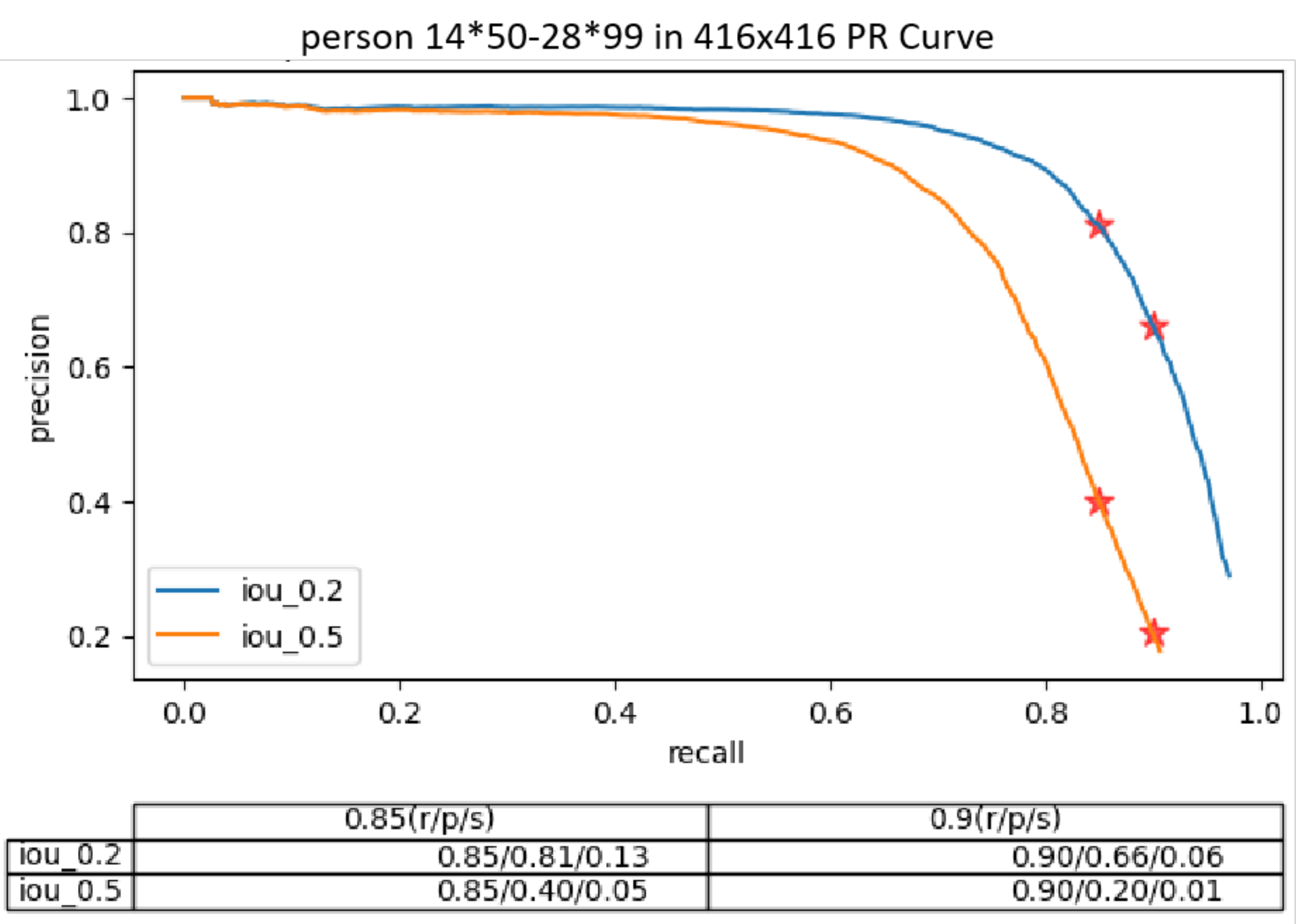}}\hspace{0.1mm}
\subfloat{\includegraphics[width=5.6cm, height=4.5cm]{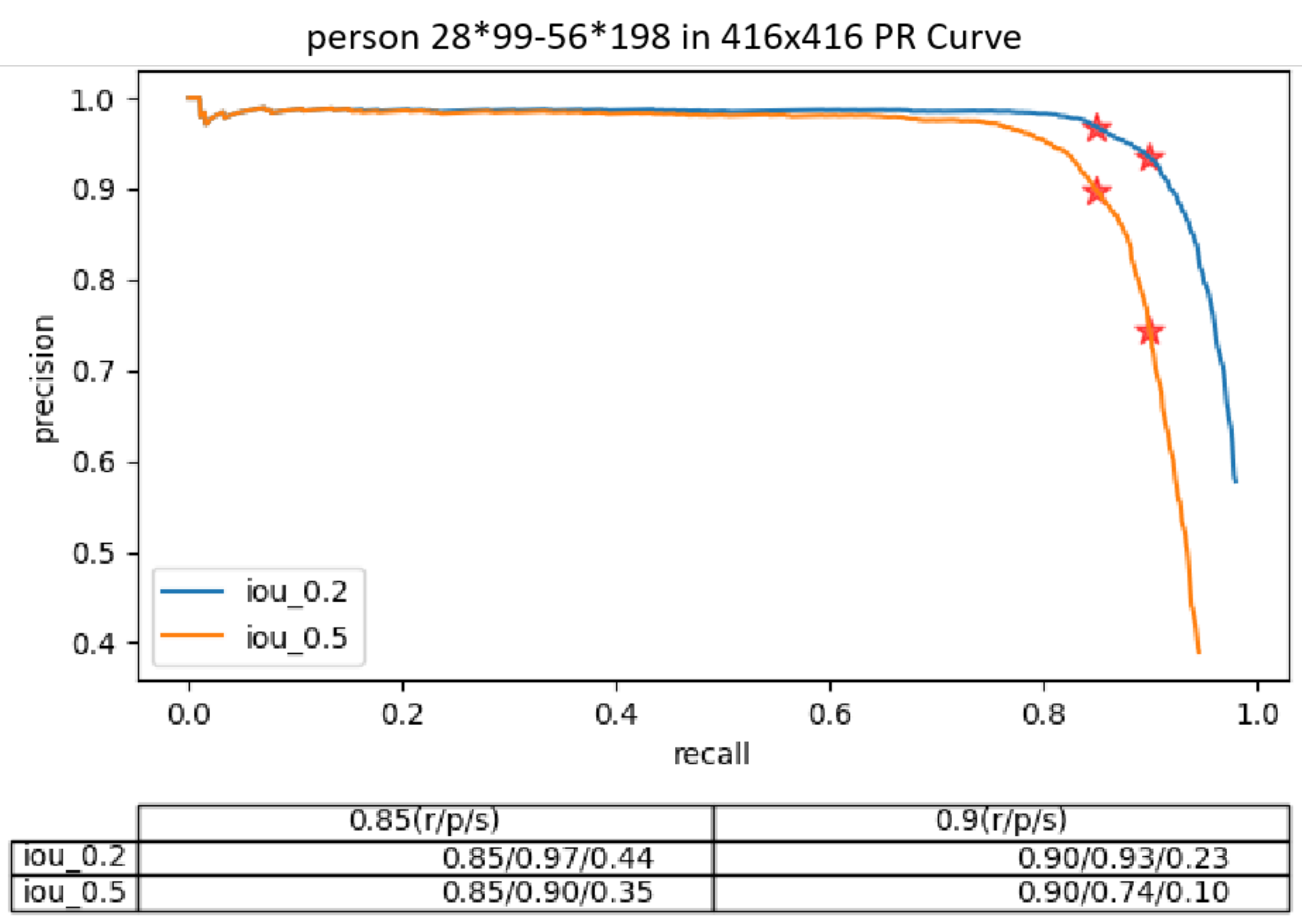}}\hspace{0.1mm}
\subfloat{\includegraphics[width=5.6cm, height=4.5cm]{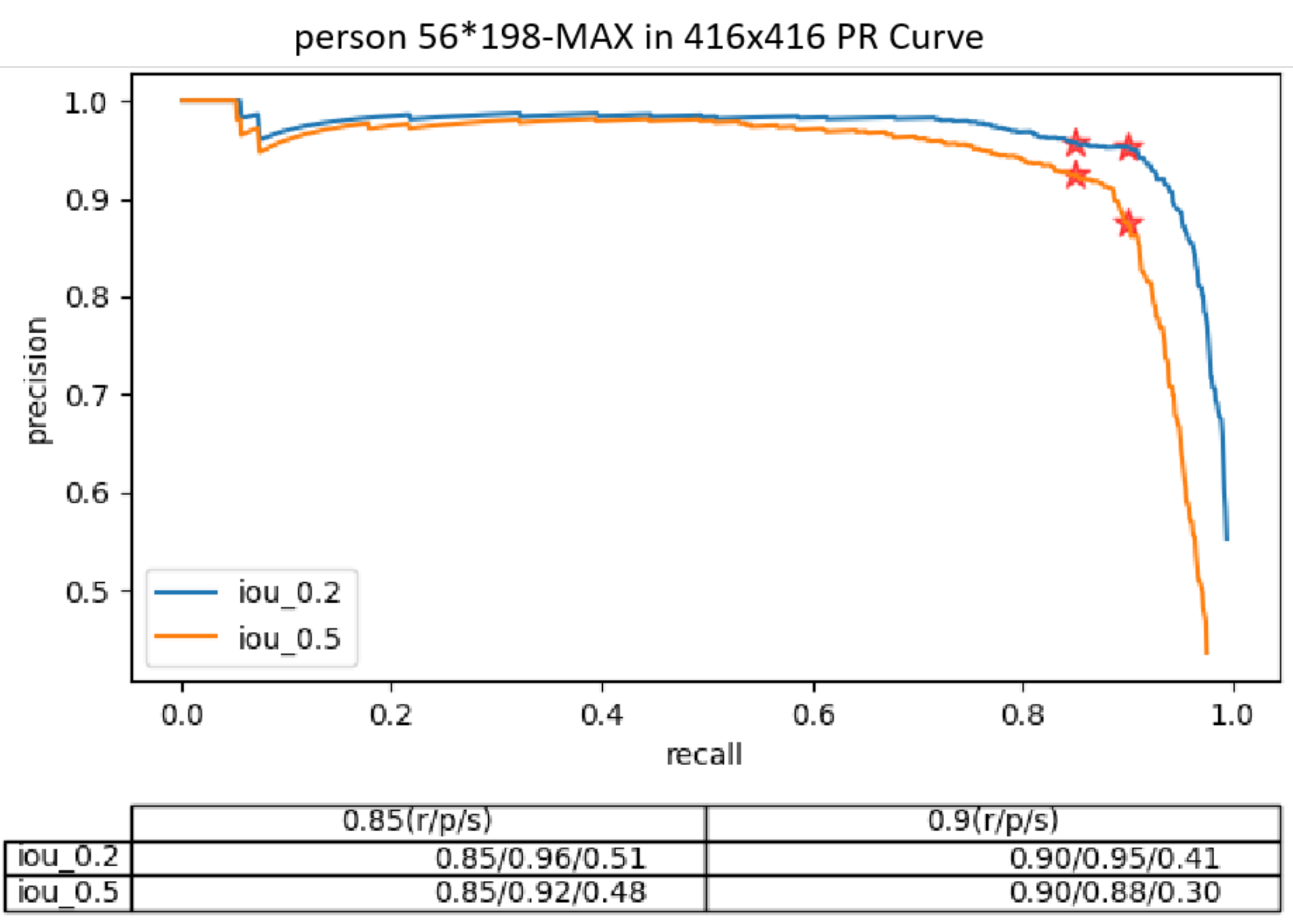}}

\setcounter{subfigure}{0}
\subfloat[14$\times$50$\sim$28$\times$99]{\includegraphics[width=5.6cm, height=4.5cm]{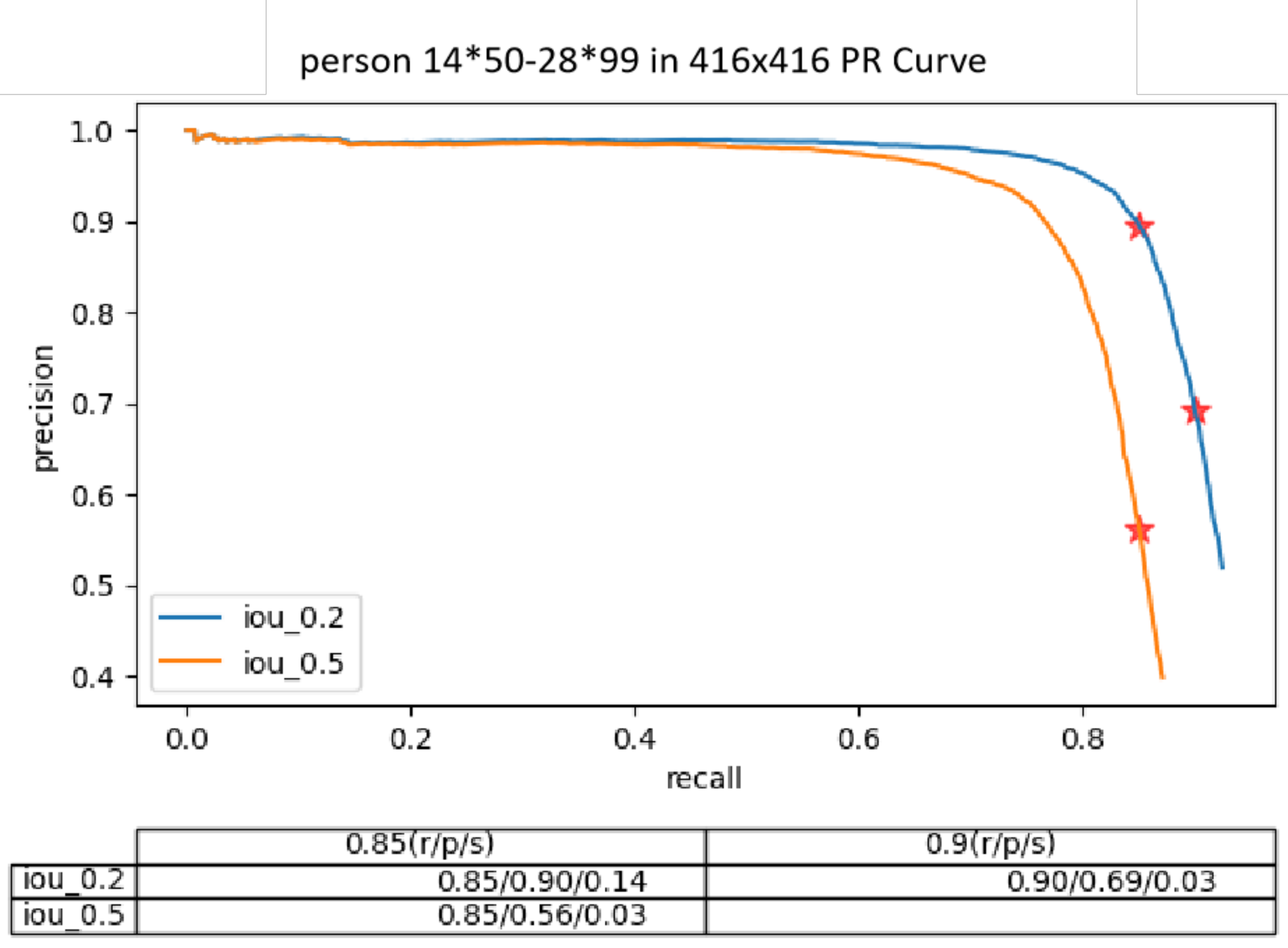}}\hspace{0.1mm}
\subfloat[28$\times$99$\sim$56$\times$198]{\includegraphics[width=5.6cm, height=4.5cm]{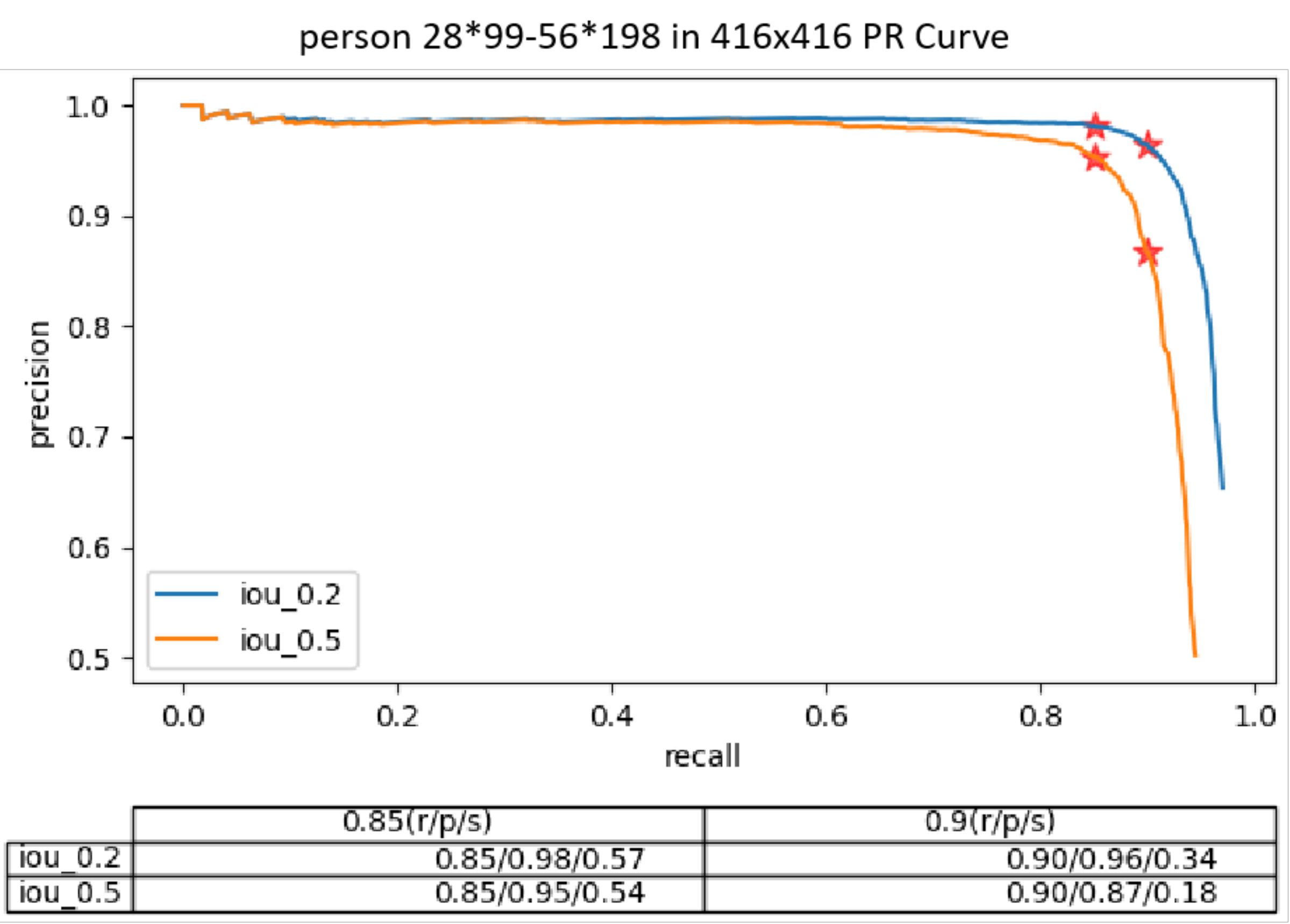}}\hspace{0.1mm}
\subfloat[56$\times$198$\sim$MAX]{\includegraphics[width=5.6cm, height=4.5cm]{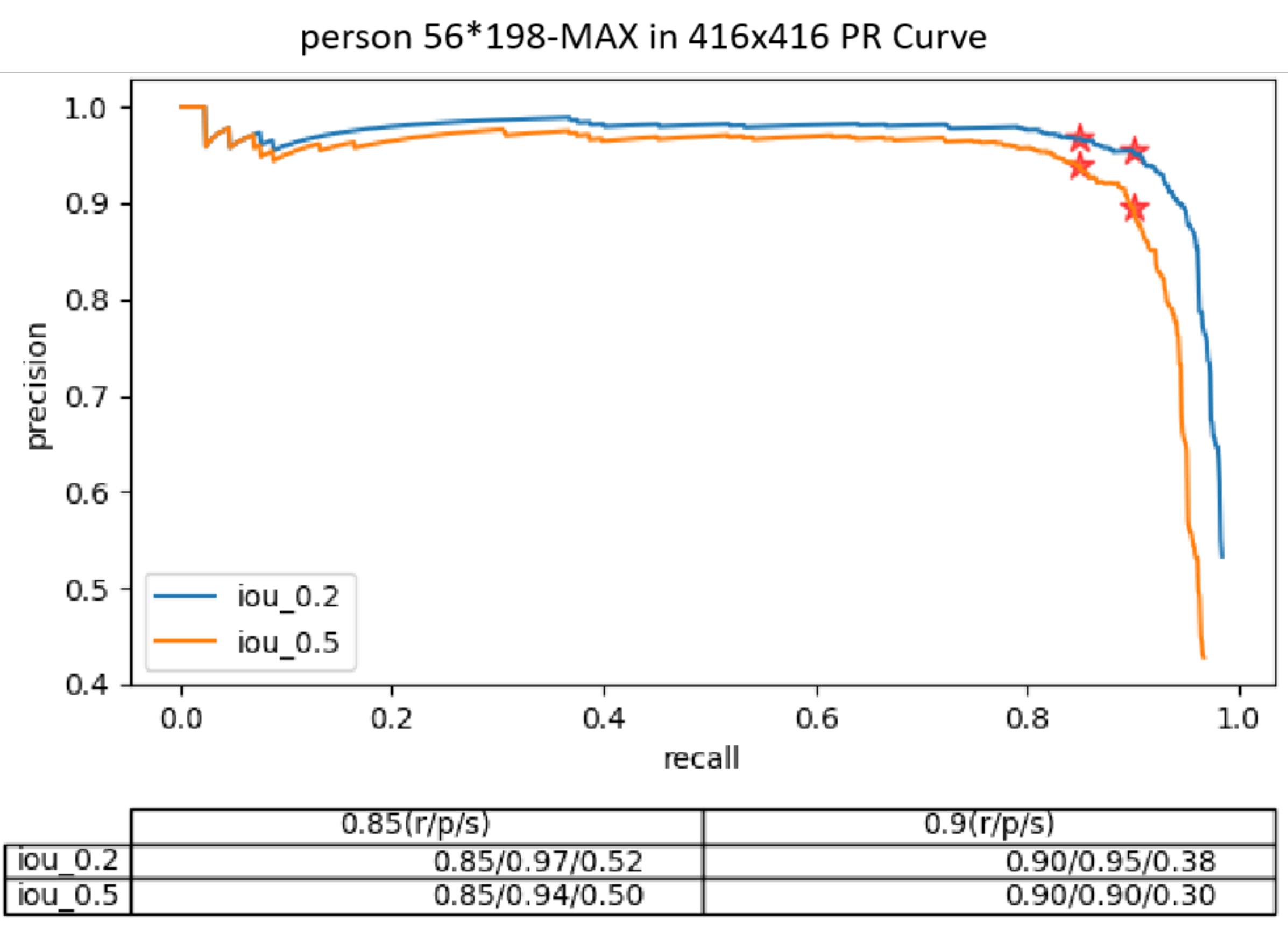}}
\caption{PR curve of three area ranges: (a), (b), (c) presents the PR curve of the area ranges between 14$\times$50$\sim$28$\times$99, 28$\times$99$\sim$56$\times$198, and 56$\times$198$\sim$MAX.
The first and second rows represent the PR curves of YOLOX and FemtoDet, respectively.
In addition, each sub-figure's horizontal axis represents recall and the vertical axis indicates precision; the table in the sub-figure shows the corresponding precision and confidence scores of AP20 and AP50 when the recall reaches 0.85 and 0.9.
}
\label{PRCurve}
\end{figure*}
\section{Analyze and Compare Pedestrian Detection Capabilities of FemtoDet and YOLOX}
In this section, We construct experimental results of this section on the TJU-DHD campus dataset as mentioned in Sec. 4.2.2.
The box size distribution of pedestrians with input size of 416$\times$416 is shown in Fig. \ref{person_box_distribution}.
According to the different sizes of pedestrians in the validation set, we divide it into five area ranges (0$\sim$11$\times$37, 11$\times$37$\sim$14$\times$50, 14$\times$50$\sim$28$\times$99, 28$\times$99$\sim$56$\times$198, and 56$\times$198$\sim$MAX) to evaluate detectors.
It can be found that the pedestrians are mostly small objects (0$\sim$11$\times$37), accounting for 35.81$\%$ of the total.
Therefore, it is unreasonable for extreme detectors to display only their average precision (mAP, AP50, or AP20) on the whole validation set.
Next we will focus on the detection results of the detectors at the three largest scales (14$\times$50$\sim$28$\times$99, 28$\times$99$\sim$56$\times$198, and 56$\times$198$\sim$MAX). 

Table \ref{PRCurve} presents the PR curve of the above three area ranges.
In the area with smaller range of 14$\times$50$\sim$28$\times$99:
1) Although the FemtoDet's highest recall of AP50 can't reach 0.90, its precision is higher (0.56 $\&$ 0.40) when the recall reaches 0.85;
2) FemtoDet has better precision when its recall is 0.85 (0.90 $\&$ 0.81) and 0.90 
 (0.69 $\&$ 0.66) uneder AP20.
In the area with medium range of 28$\times$99$\sim$56$\times$198:
FemtoDet not only achieves better precision, but also has higher confidence with higher recall, both under the evaluation matrics of AP20 (recall/precision/scores: 0.85/0.98/0.57 $\&$ 0.85/0.97/0.44, 0.90/0.96/0.34 $\&$ 0.90/0.93/0.23) and AP50 (recall/precision/scores: 0.85/0.95/0.54 $\&$ 0.85/0.90/0.35, 0.90/0.87/0.18 $\&$ 0.90/0.74/0.10).
This means FemtoDet is much robust.
In the area with maximal range of 56$\times$198$\sim$MAX:
While both extremely light-weight detectors demonstrate high performance on the detection of large objects, FemtoDet hold up relatively better, achieving a recall of 0.85 as AP5o with a  precision improvement of about 2.17$\%$ (0.94 $\&$ 0.92) over the comparison method.

\begin{figure}[htp]
\centering
\subfloat{\includegraphics[width=3.8cm, height=2.5cm]{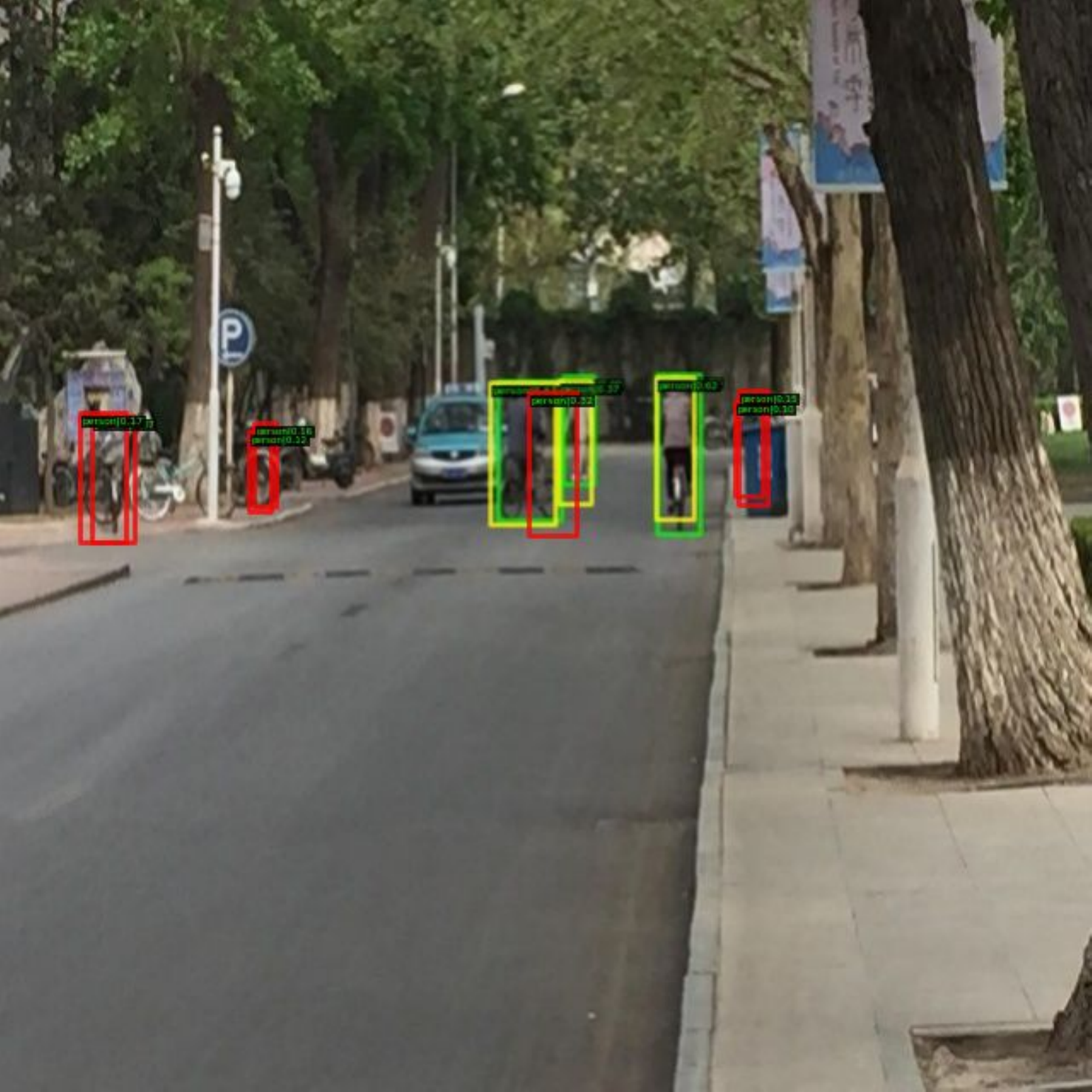}}\hspace{0.2mm}
\subfloat{\includegraphics[width=3.8cm, height=2.5cm]{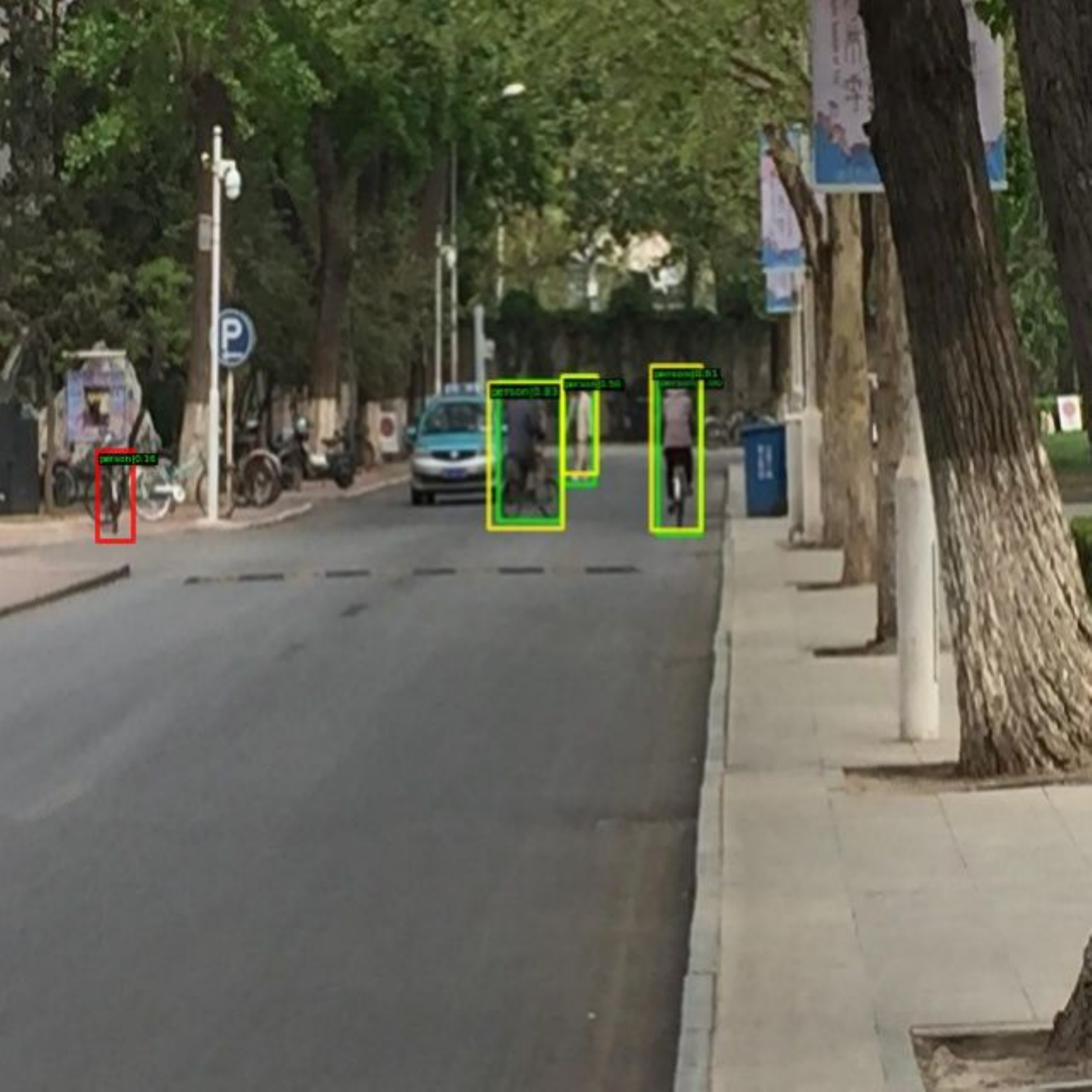}}\\\vspace{0.1mm}

\subfloat{\includegraphics[width=3.8cm, height=2.5cm]{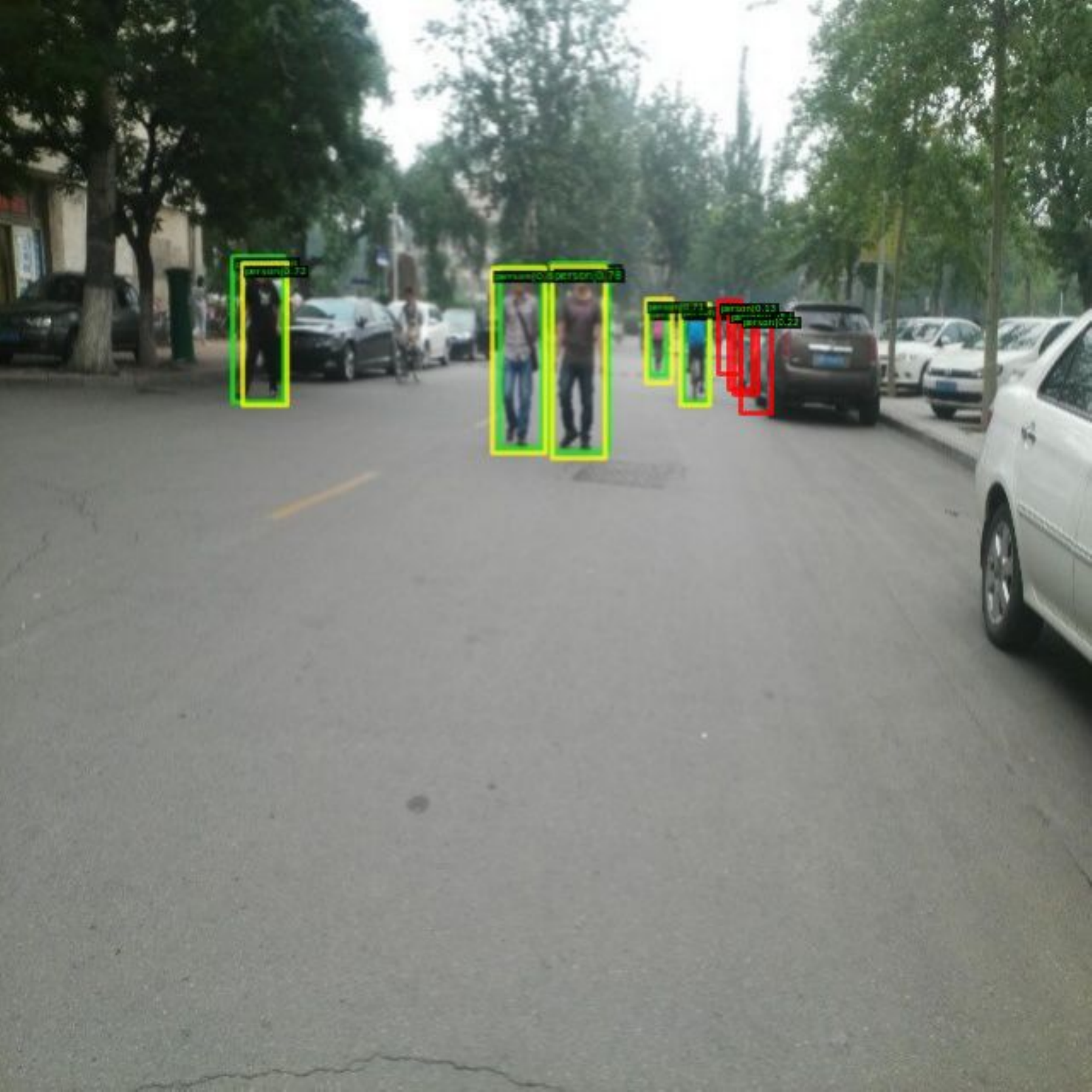}}\hspace{0.2mm}
\subfloat{\includegraphics[width=3.8cm, height=2.5cm]{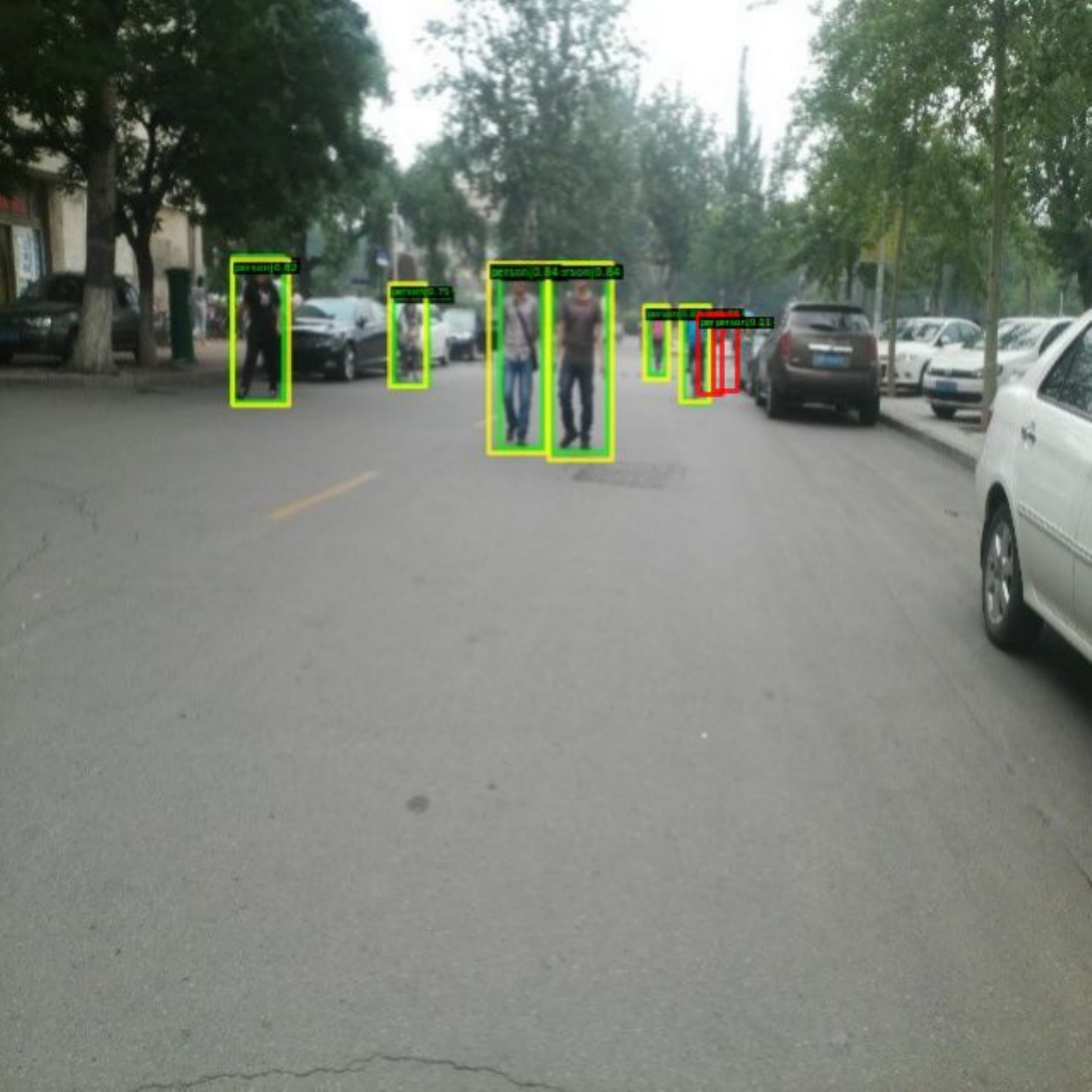}}\\\vspace{0.1mm}

\subfloat{\includegraphics[width=3.8cm, height=2.5cm]{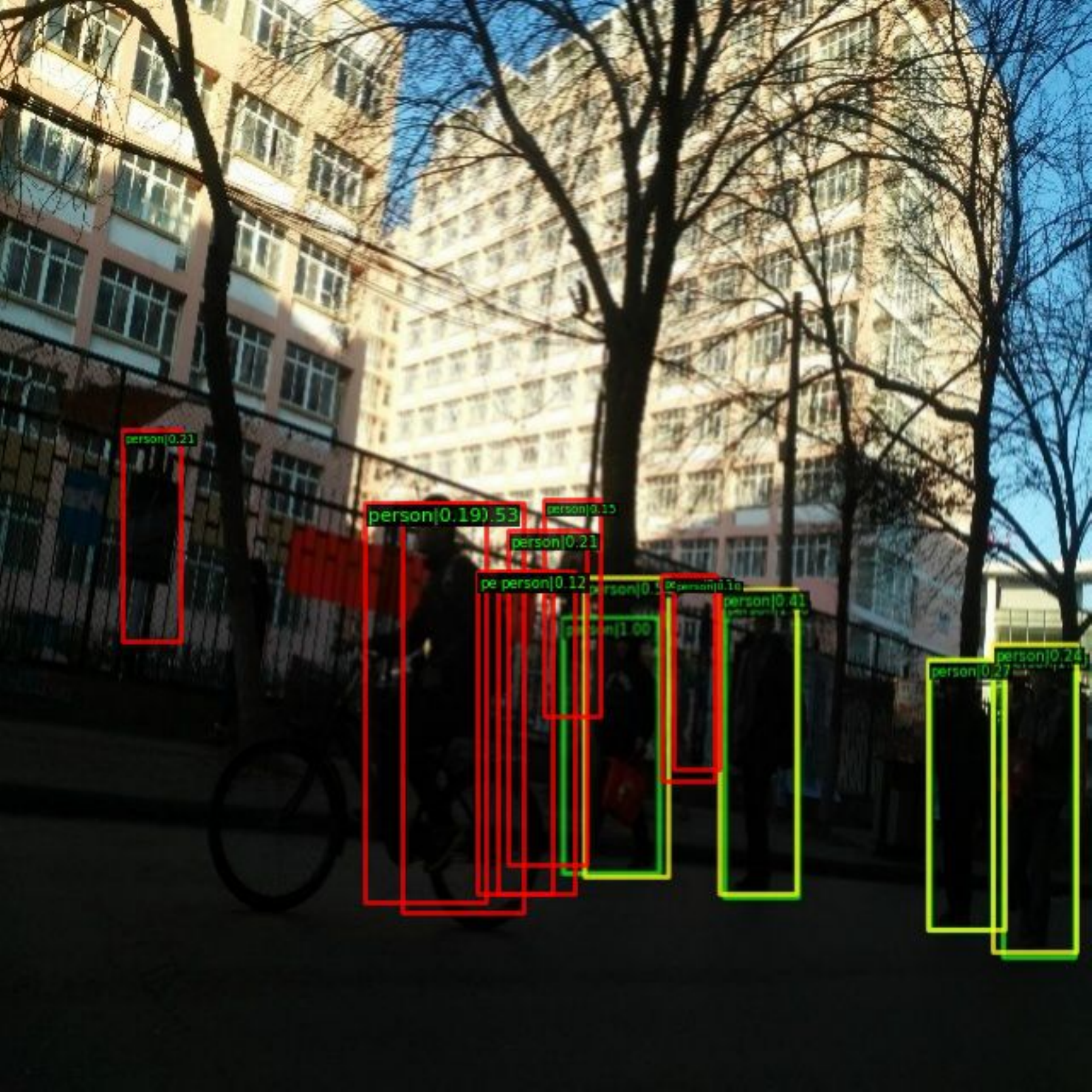}}\hspace{0.2mm}
\subfloat{\includegraphics[width=3.8cm, height=2.5cm]{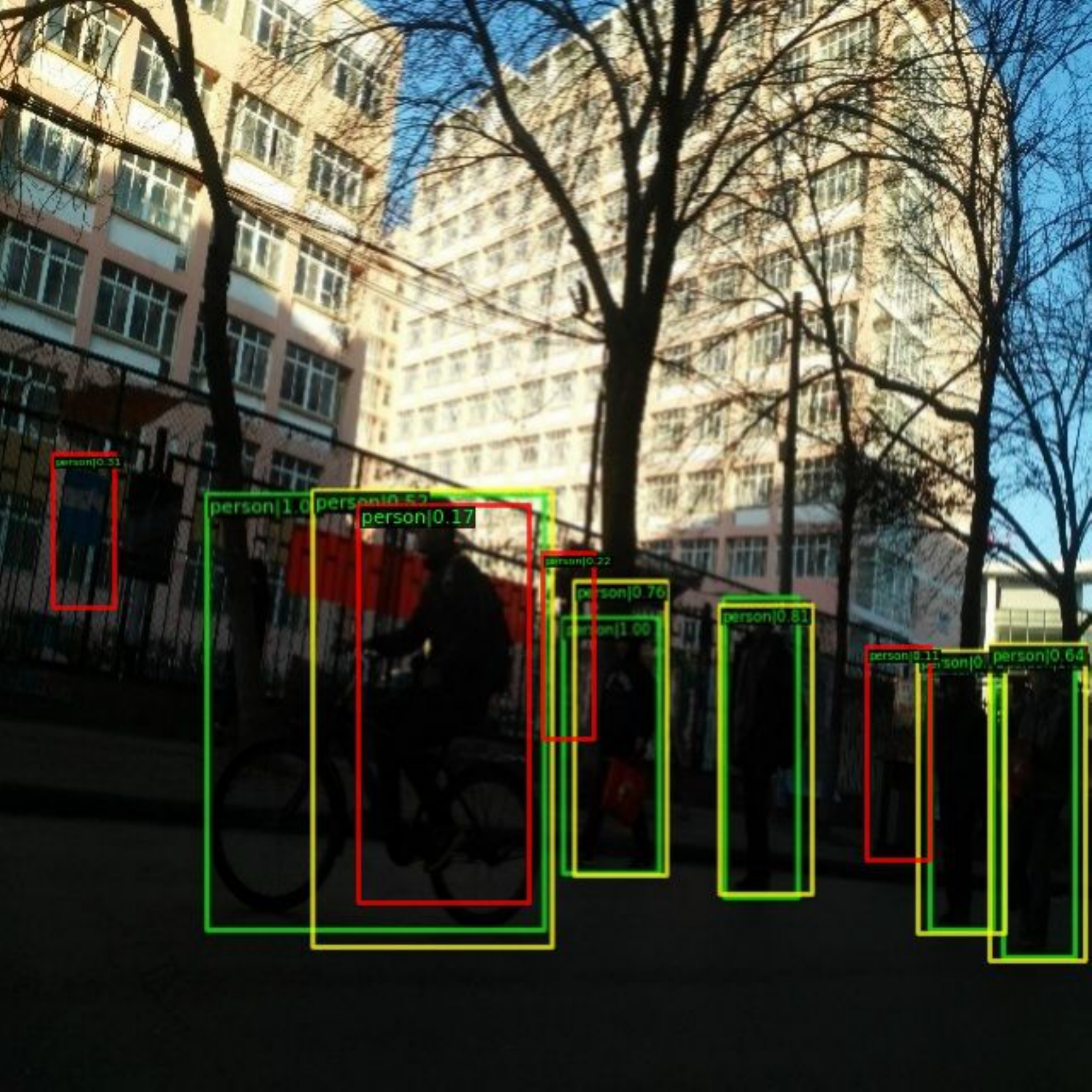}}\\\vspace{0.1mm}

\subfloat{\includegraphics[width=3.8cm, height=2.5cm]{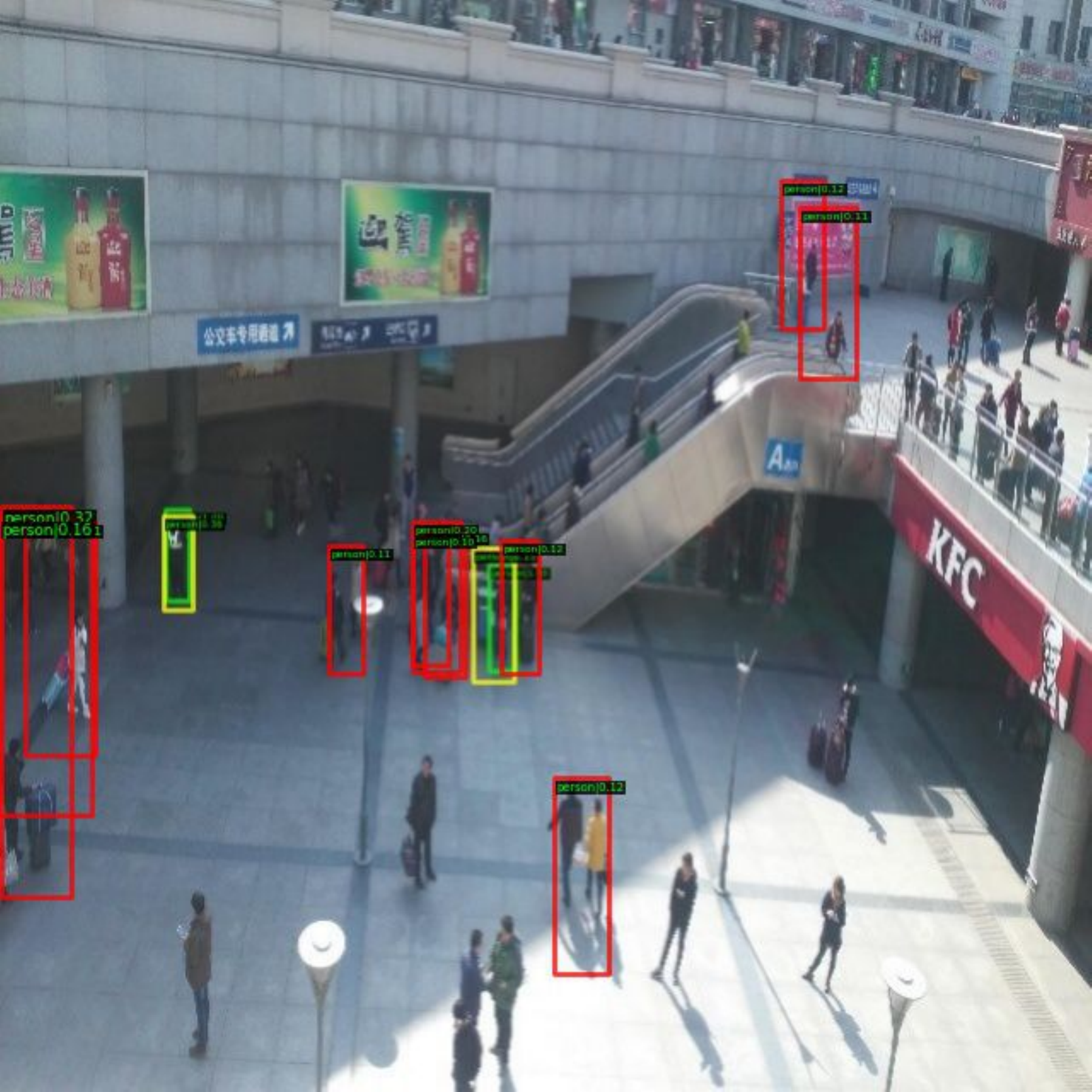}}\hspace{0.2mm}
\subfloat{\includegraphics[width=3.8cm, height=2.5cm]{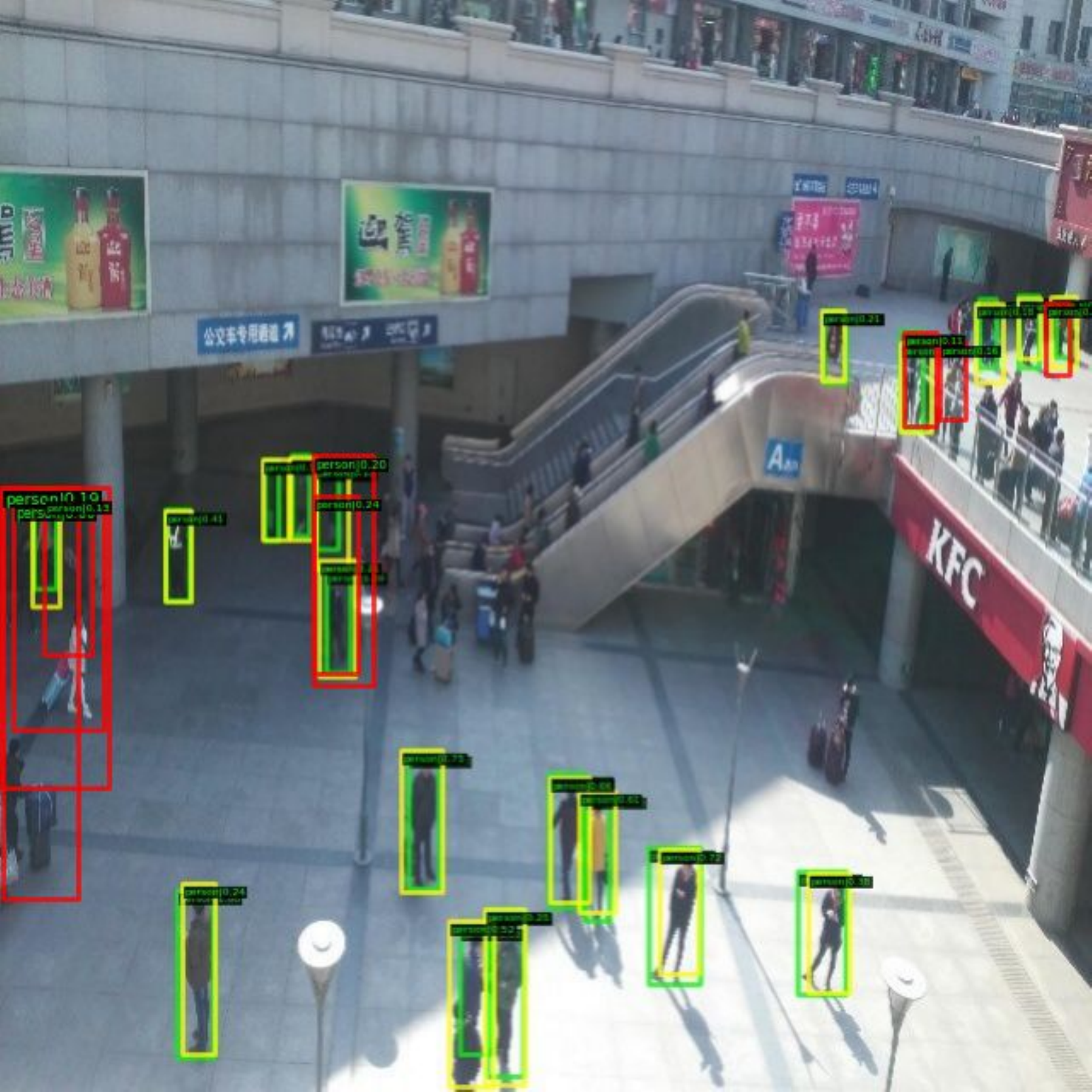}}\\\vspace{0.1mm}

\subfloat{\includegraphics[width=3.8cm, height=2.5cm]{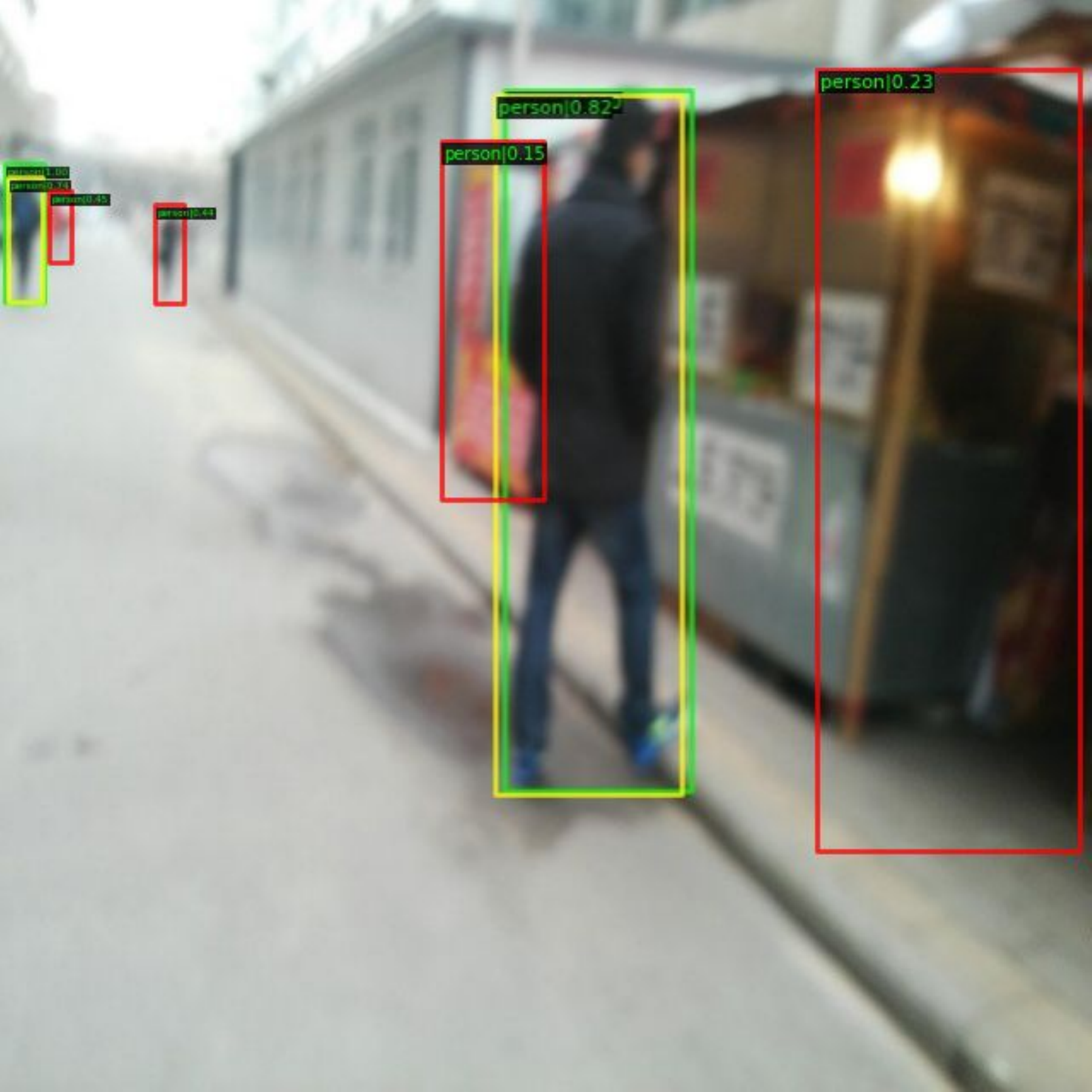}}\hspace{0.2mm}
\subfloat{\includegraphics[width=3.8cm, height=2.5cm]{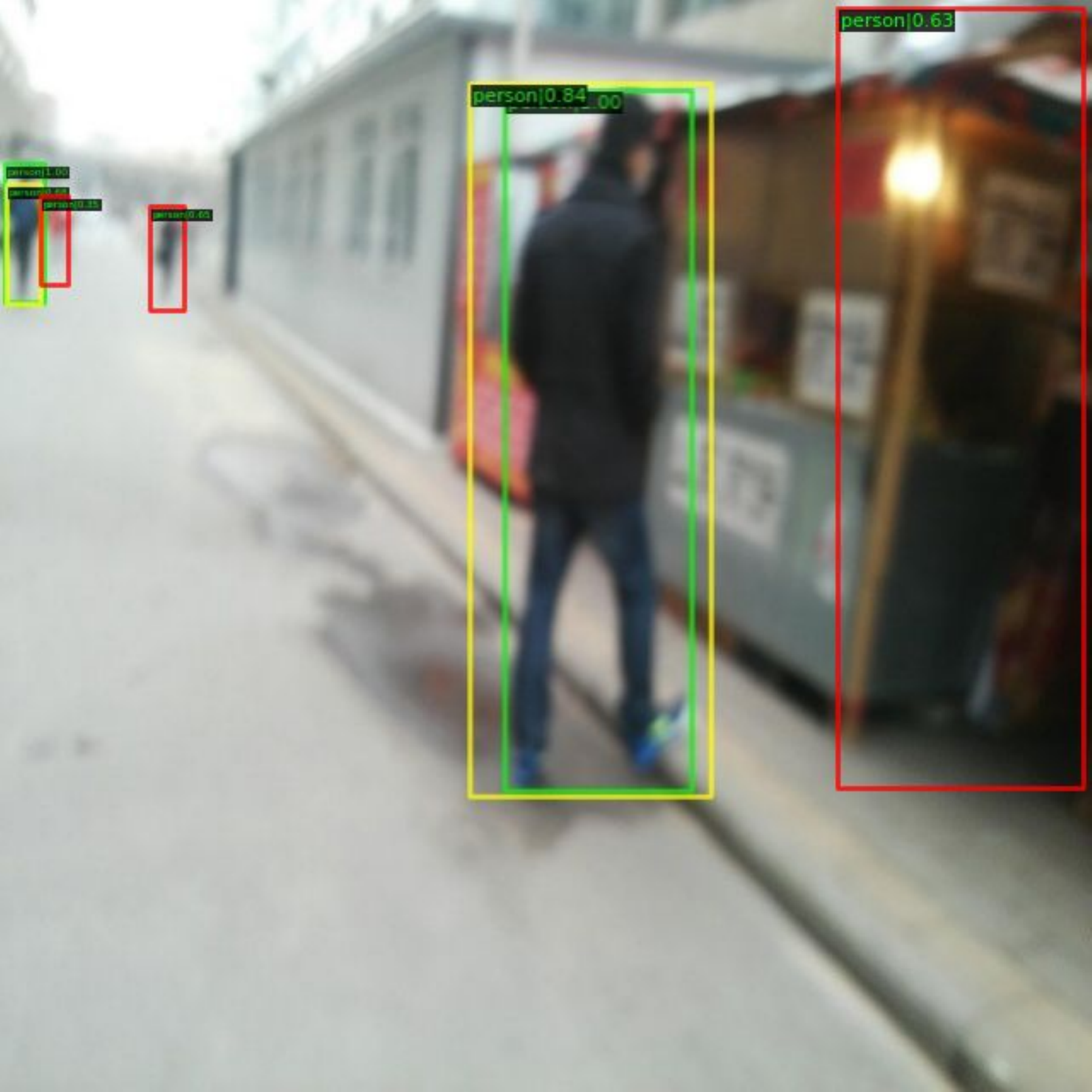}}\\\vspace{0.1mm}

\setcounter{subfigure}{0}
\subfloat[YOLOX]{\includegraphics[width=3.8cm, height=2.5cm]{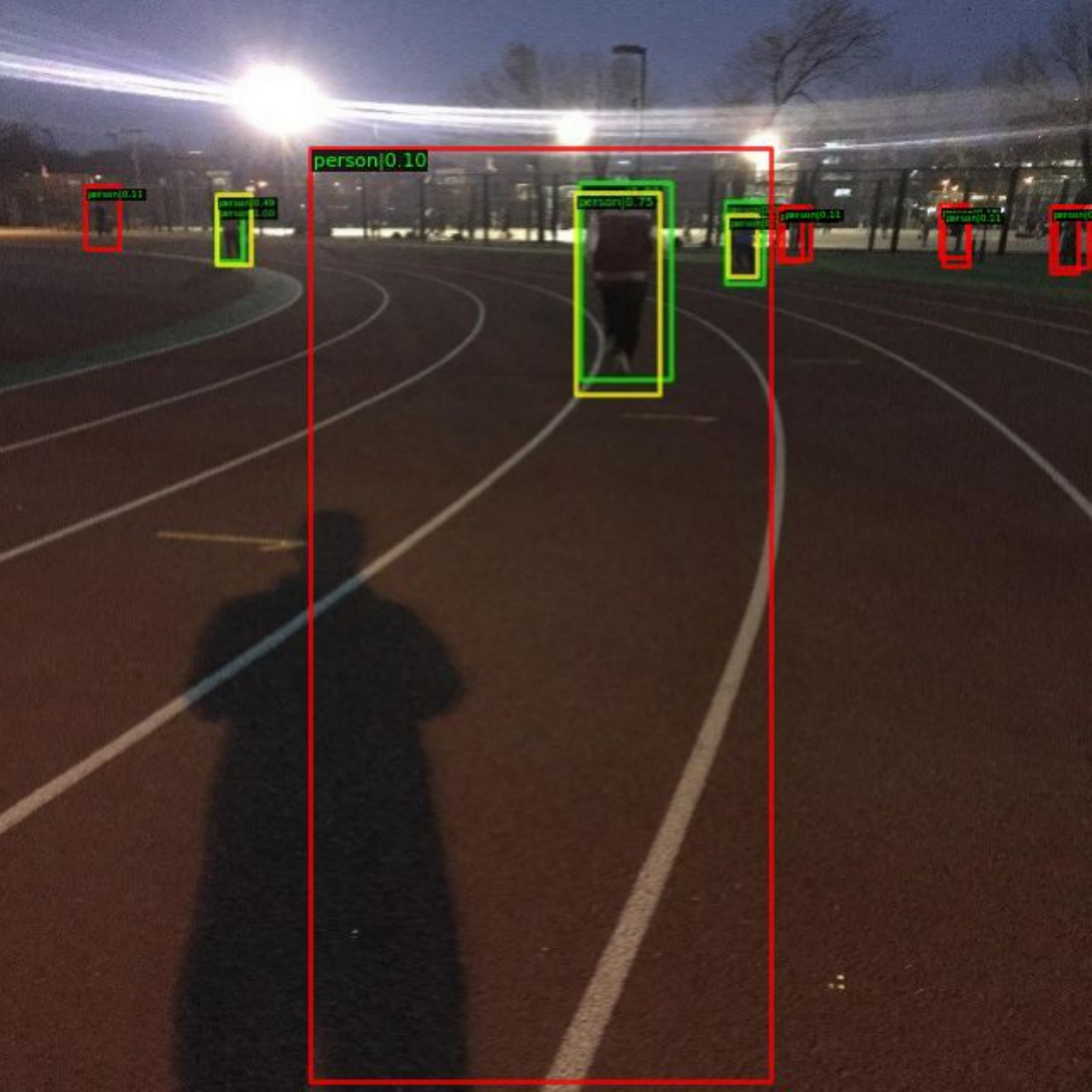}}\hspace{0.2mm}
\subfloat[FemtoDet]{\includegraphics[width=3.8cm, height=2.5cm]{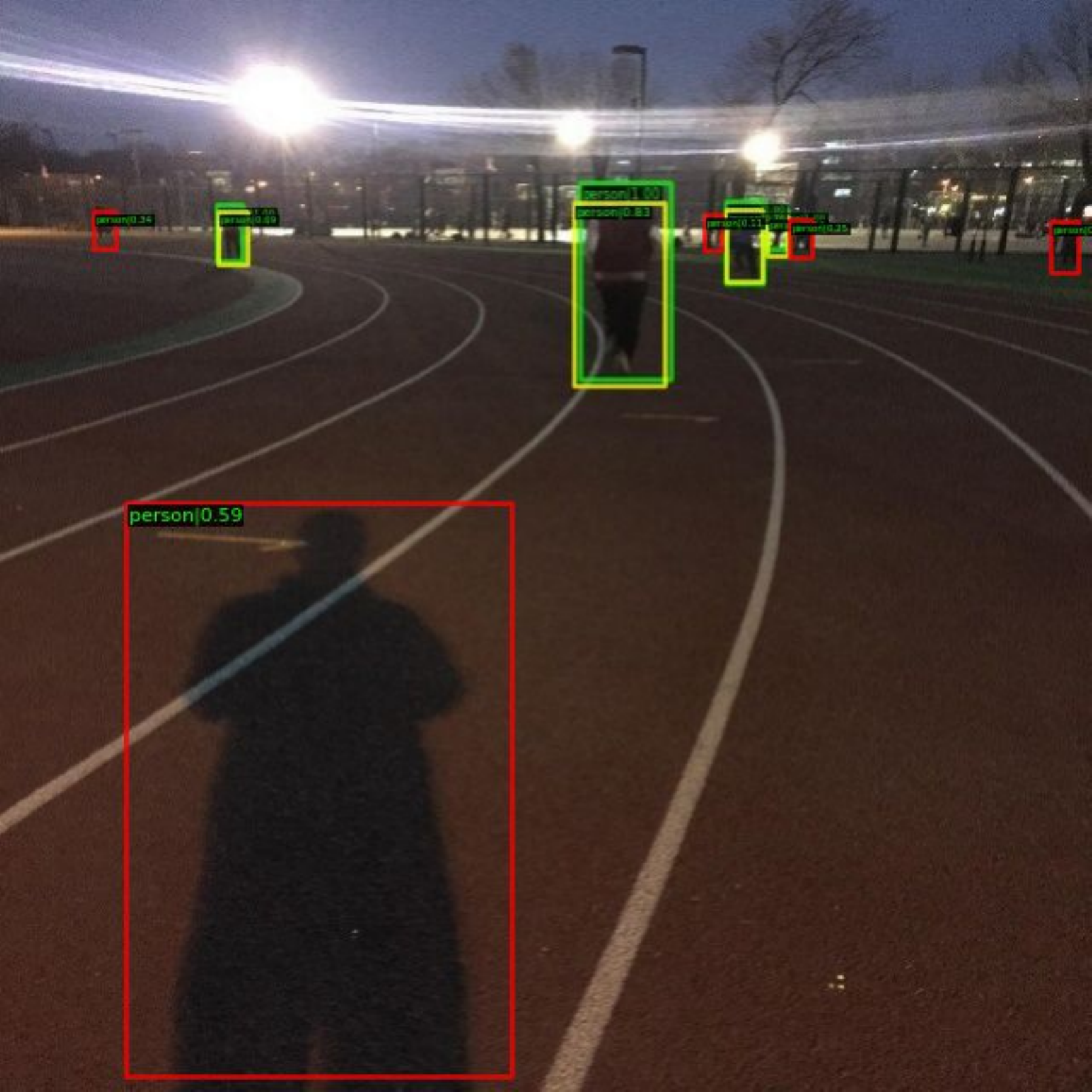}}\\
\caption{Visualization of false positive predictions.
The red boxes represent the false positive boxes, the green boxes represent the GT (ground truth), and the yellow boxes represent the correct prediction boxes.}
\label{vis_fp}
\end{figure}

\begin{figure}[htp]
\centering
\subfloat{\includegraphics[width=3.8cm, height=2.5cm]{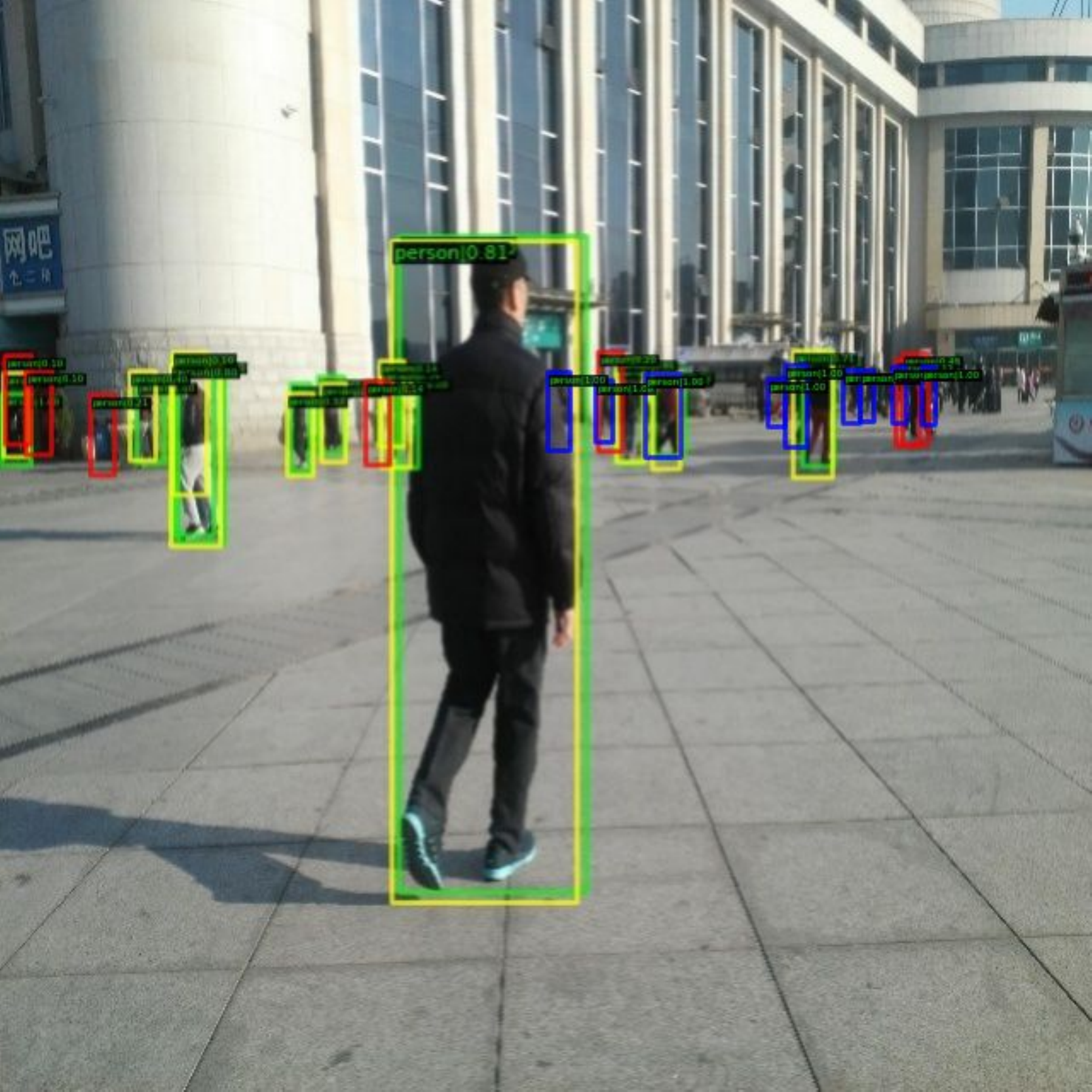}}\hspace{0.2mm}
\subfloat{\includegraphics[width=3.8cm, height=2.5cm]{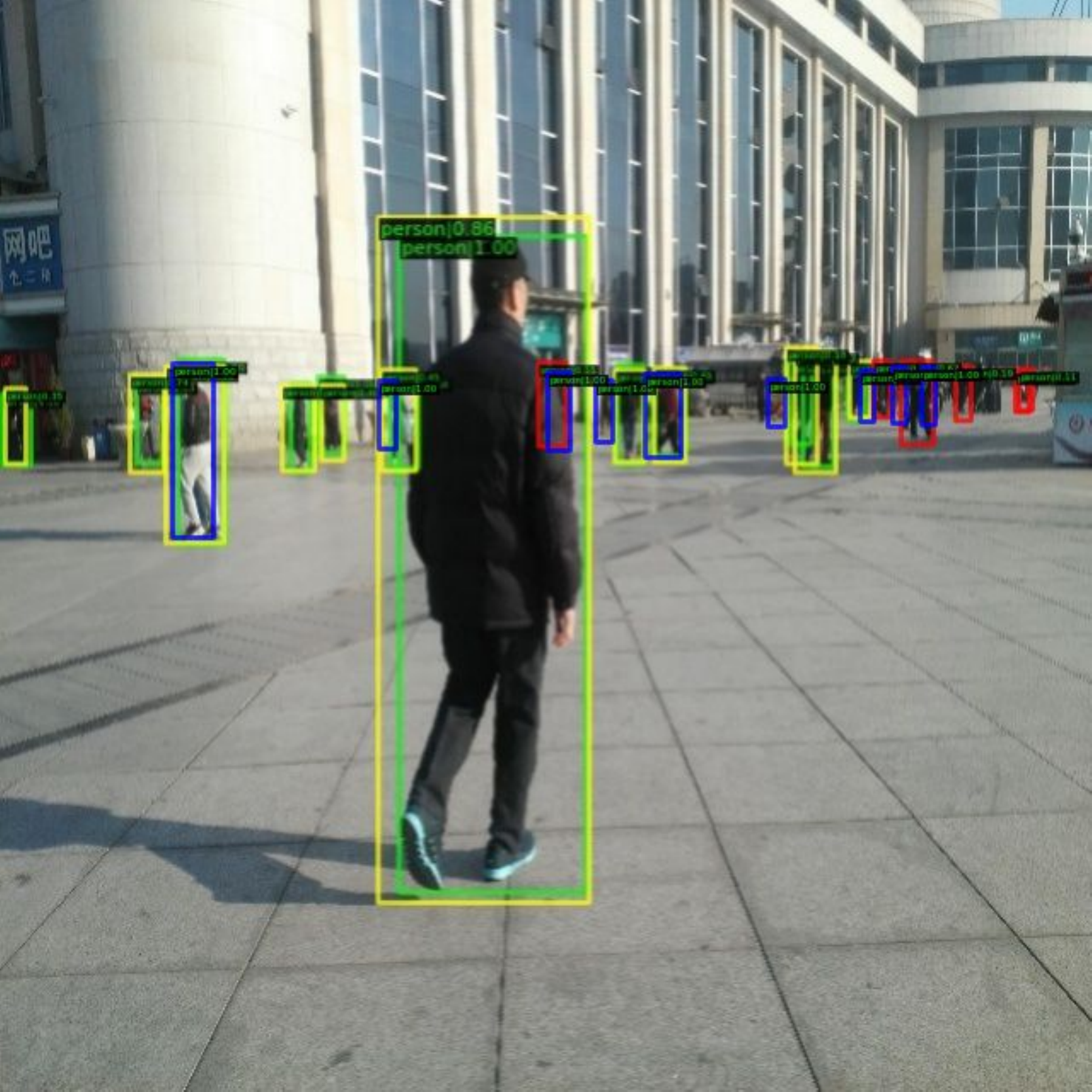}}\\\vspace{0.1mm}

\subfloat{\includegraphics[width=3.8cm, height=2.5cm]{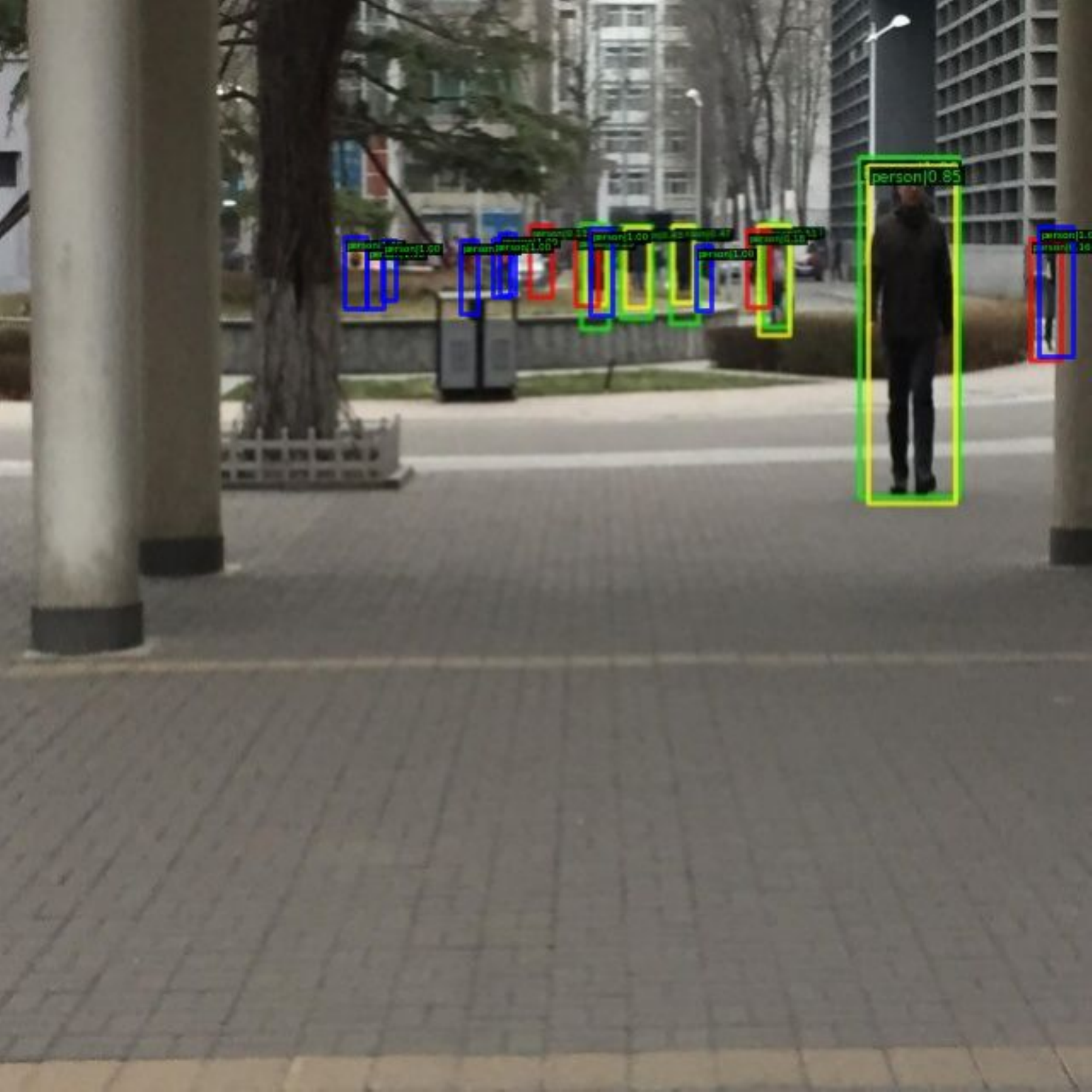}}\hspace{0.2mm}
\subfloat{\includegraphics[width=3.8cm, height=2.5cm]{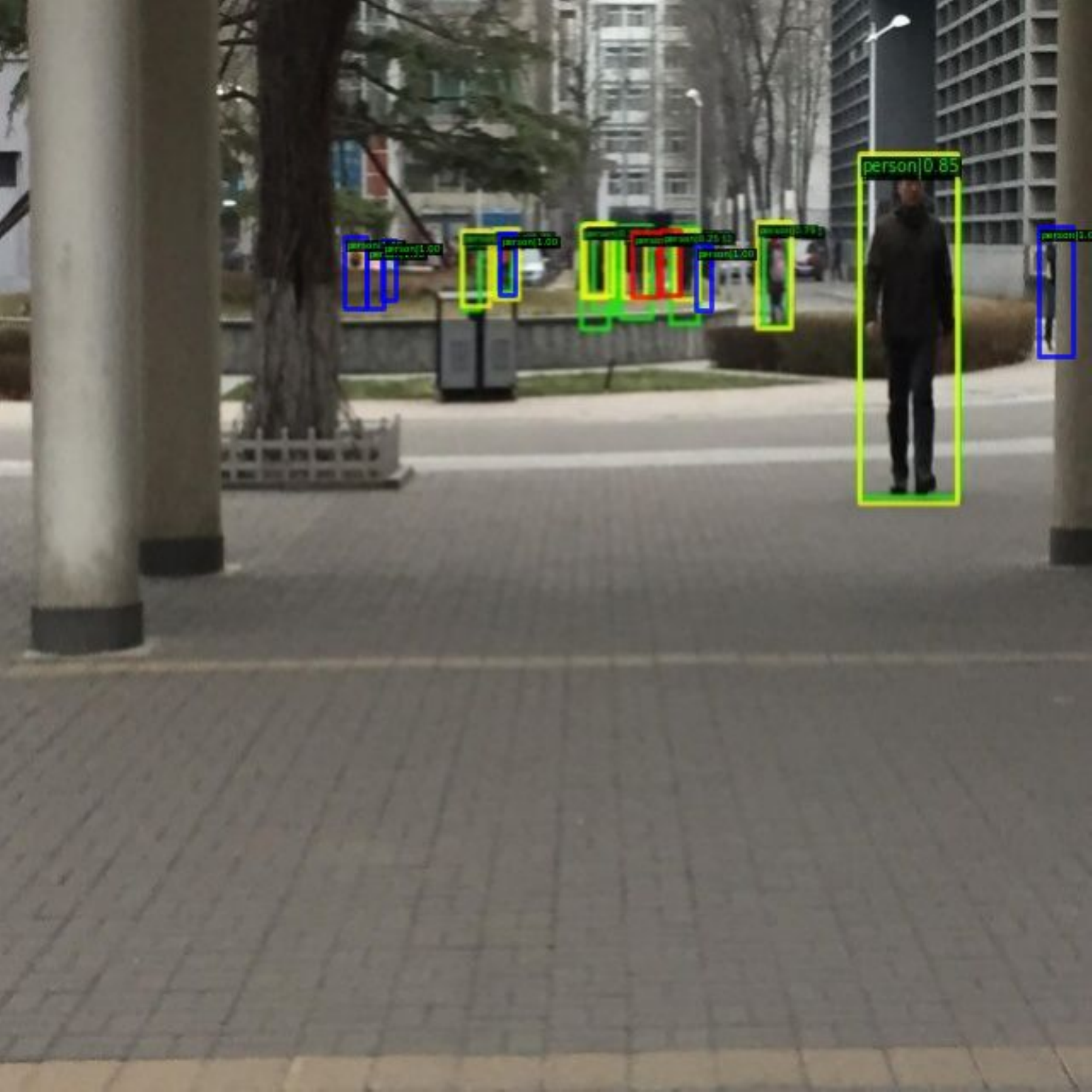}}\\\vspace{0.1mm}

\subfloat{\includegraphics[width=3.8cm, height=2.5cm]{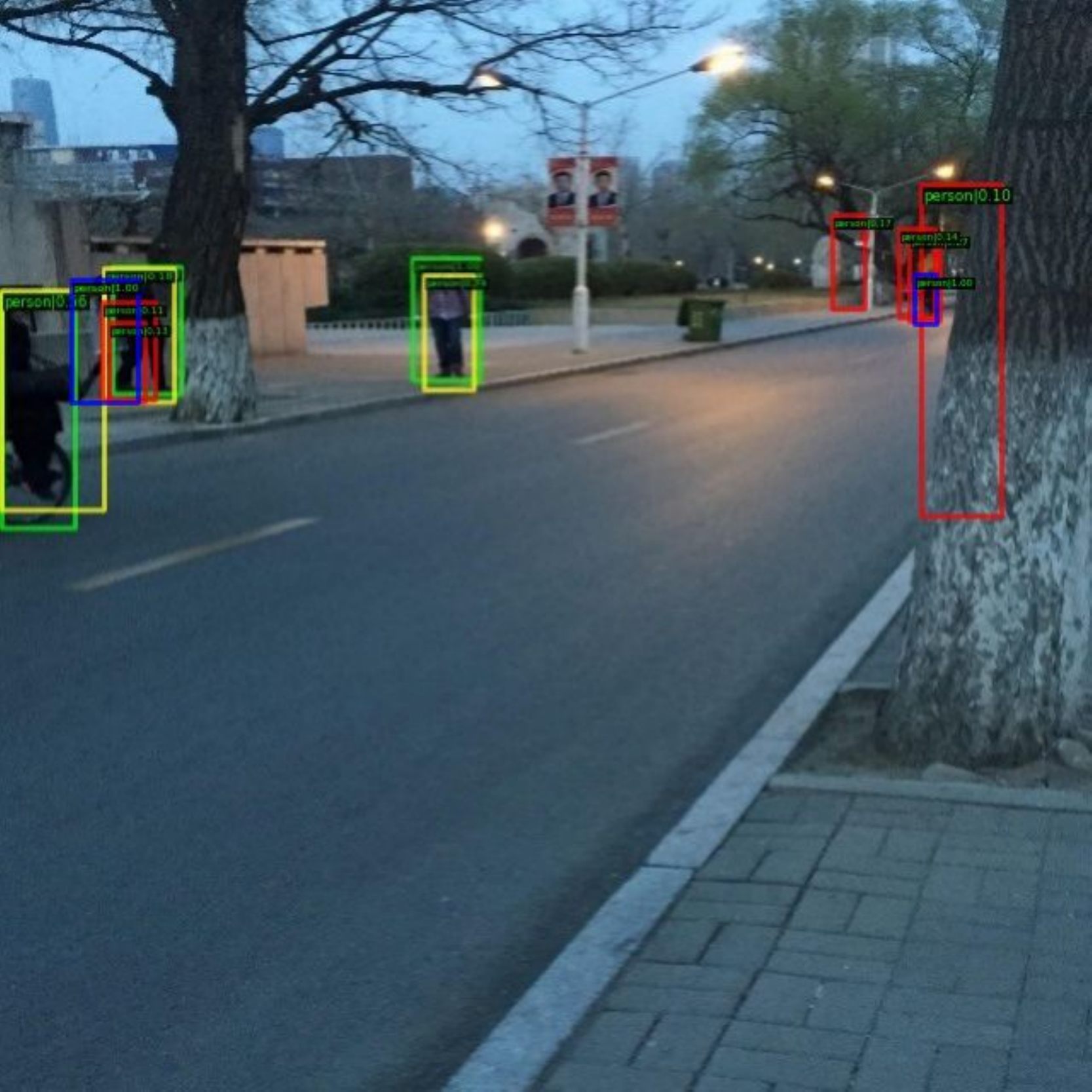}}\hspace{0.2mm}
\subfloat{\includegraphics[width=3.8cm, height=2.5cm]{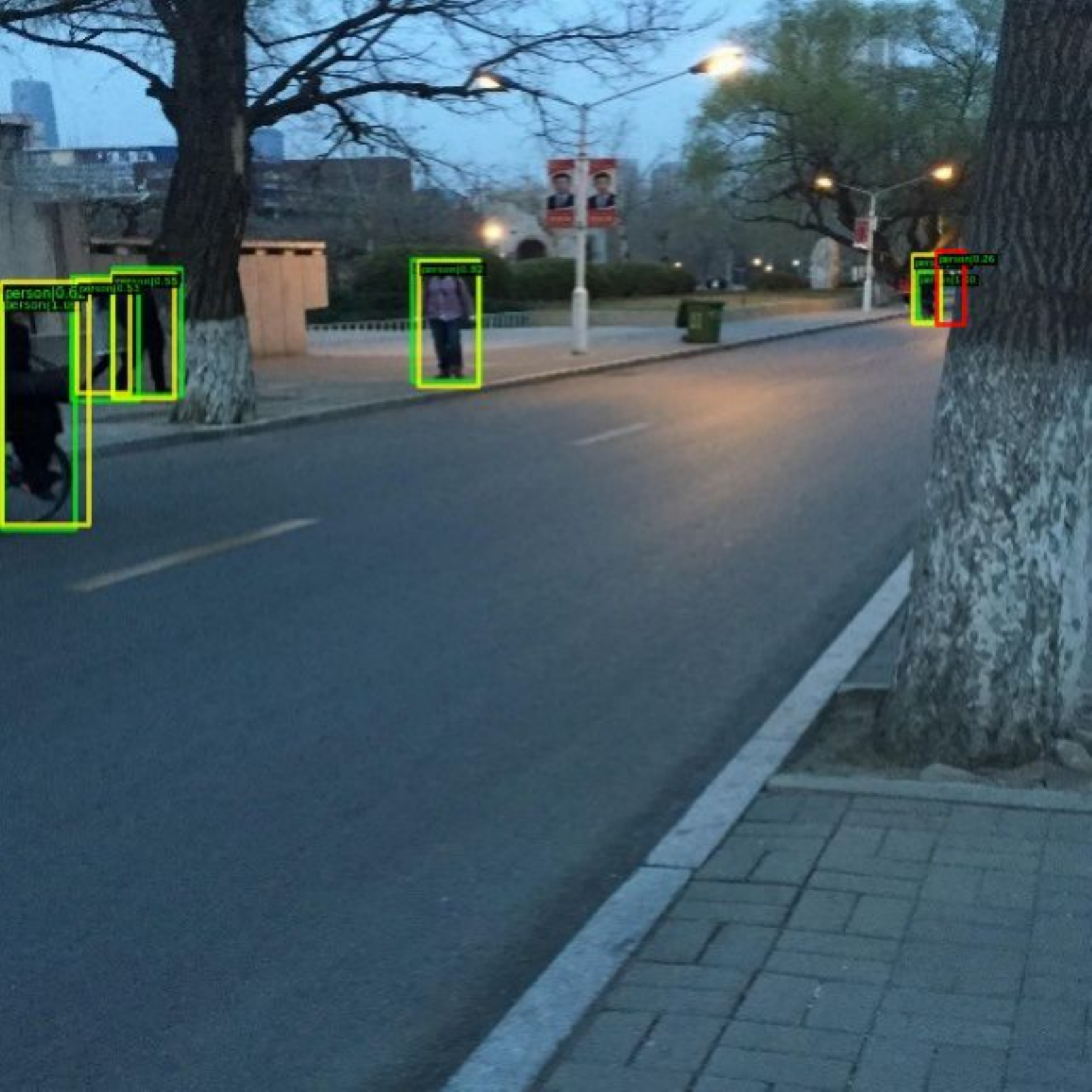}}\\\vspace{0.1mm}

\subfloat{\includegraphics[width=3.8cm, height=2.5cm]{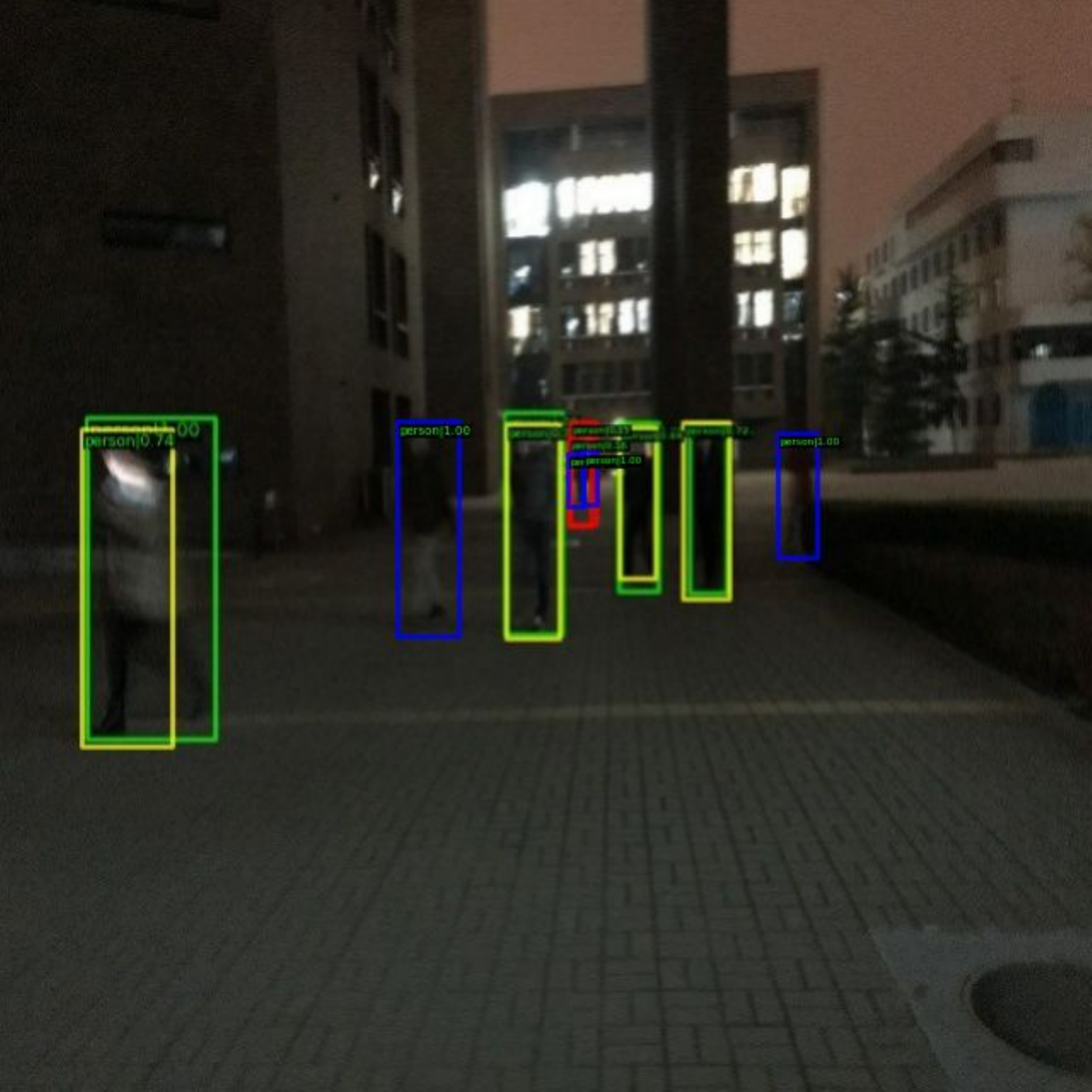}}\hspace{0.2mm}
\subfloat{\includegraphics[width=3.8cm, height=2.5cm]{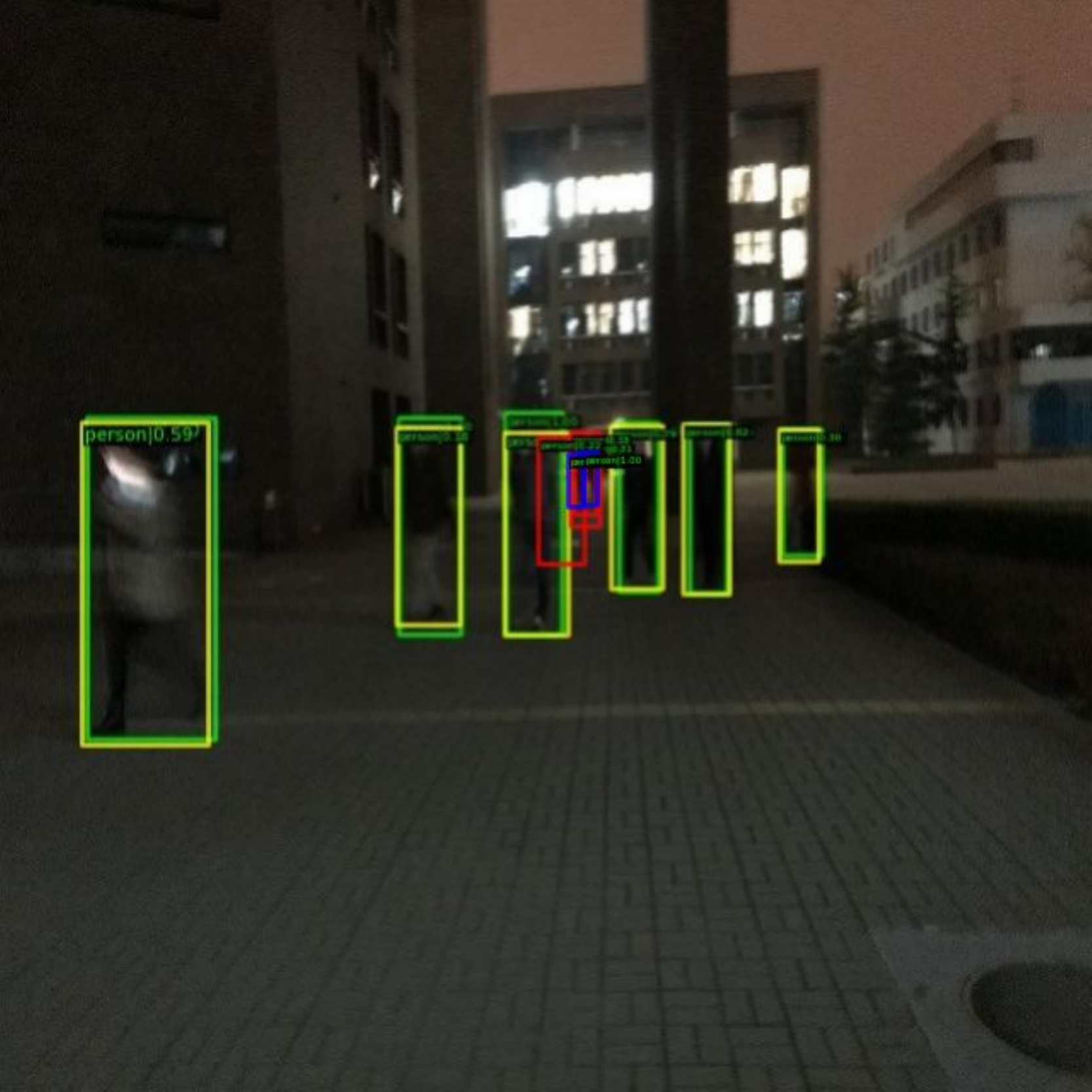}}\\\vspace{0.1mm}

\subfloat{\includegraphics[width=3.8cm, height=2.5cm]{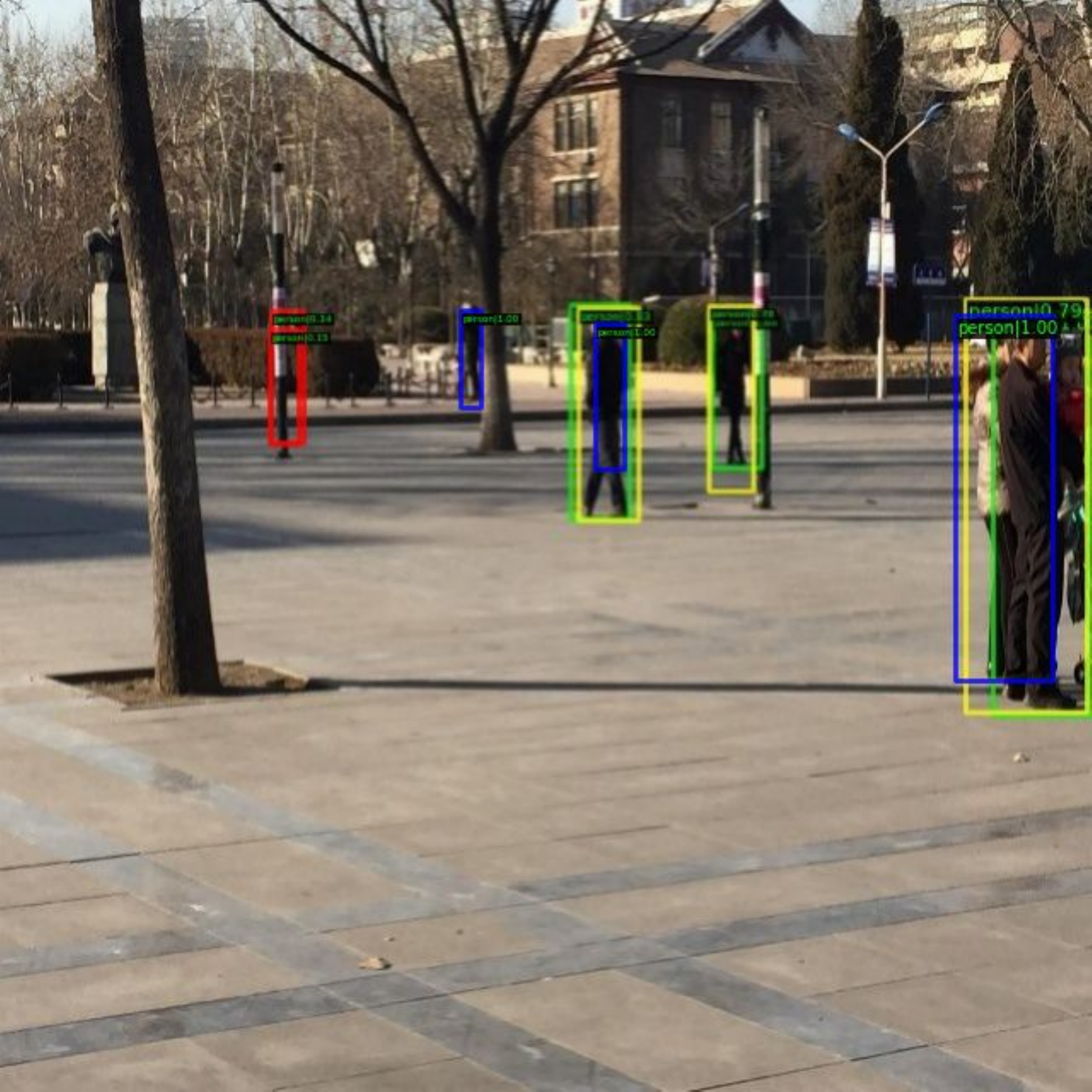}}\hspace{0.2mm}
\subfloat{\includegraphics[width=3.8cm, height=2.5cm]{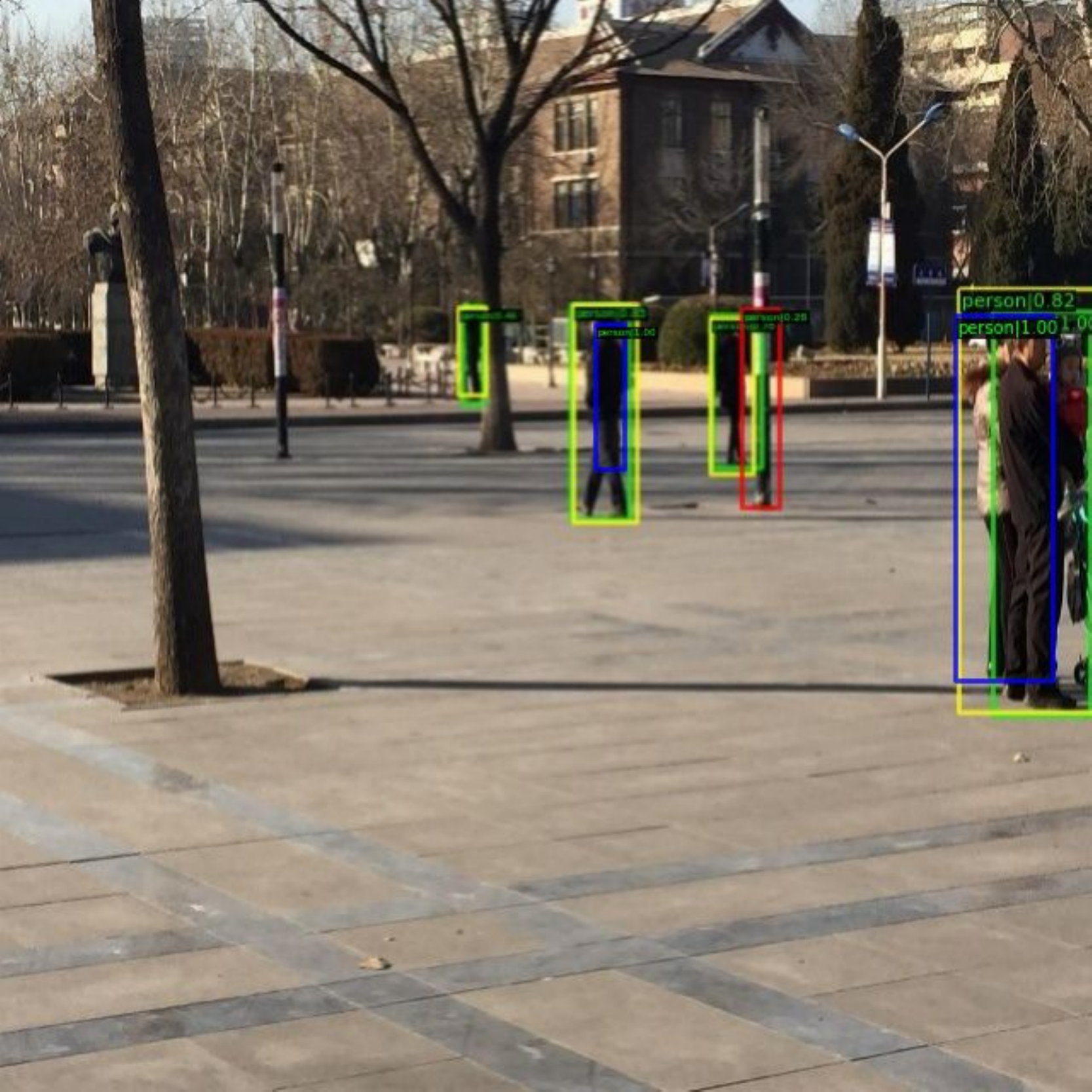}}\\\vspace{0.1mm}

\setcounter{subfigure}{0}
\subfloat[YOLOX]{\includegraphics[width=3.8cm, height=2.5cm]{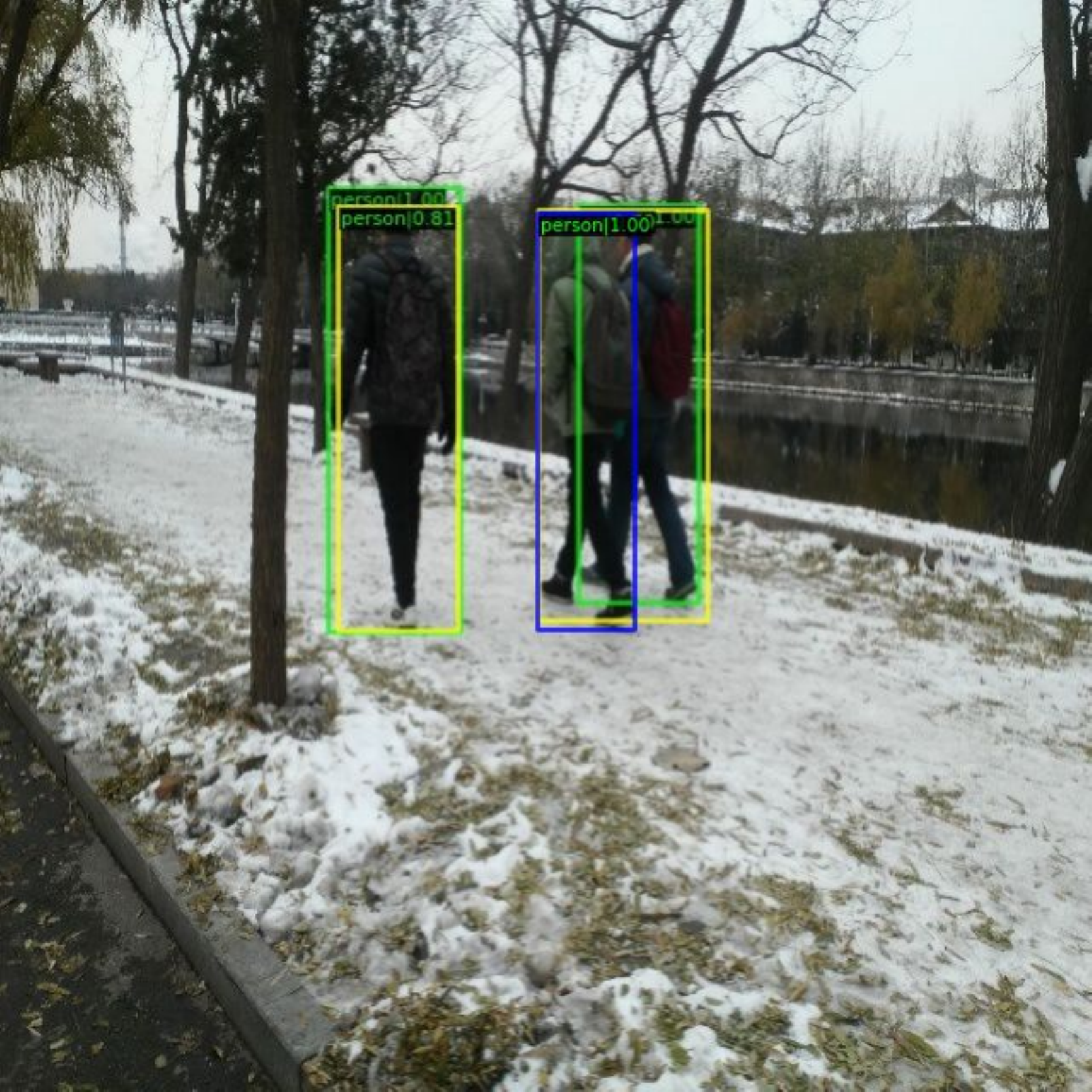}}\hspace{0.2mm}
\subfloat[FemtoDet]{\includegraphics[width=3.8cm, height=2.5cm]{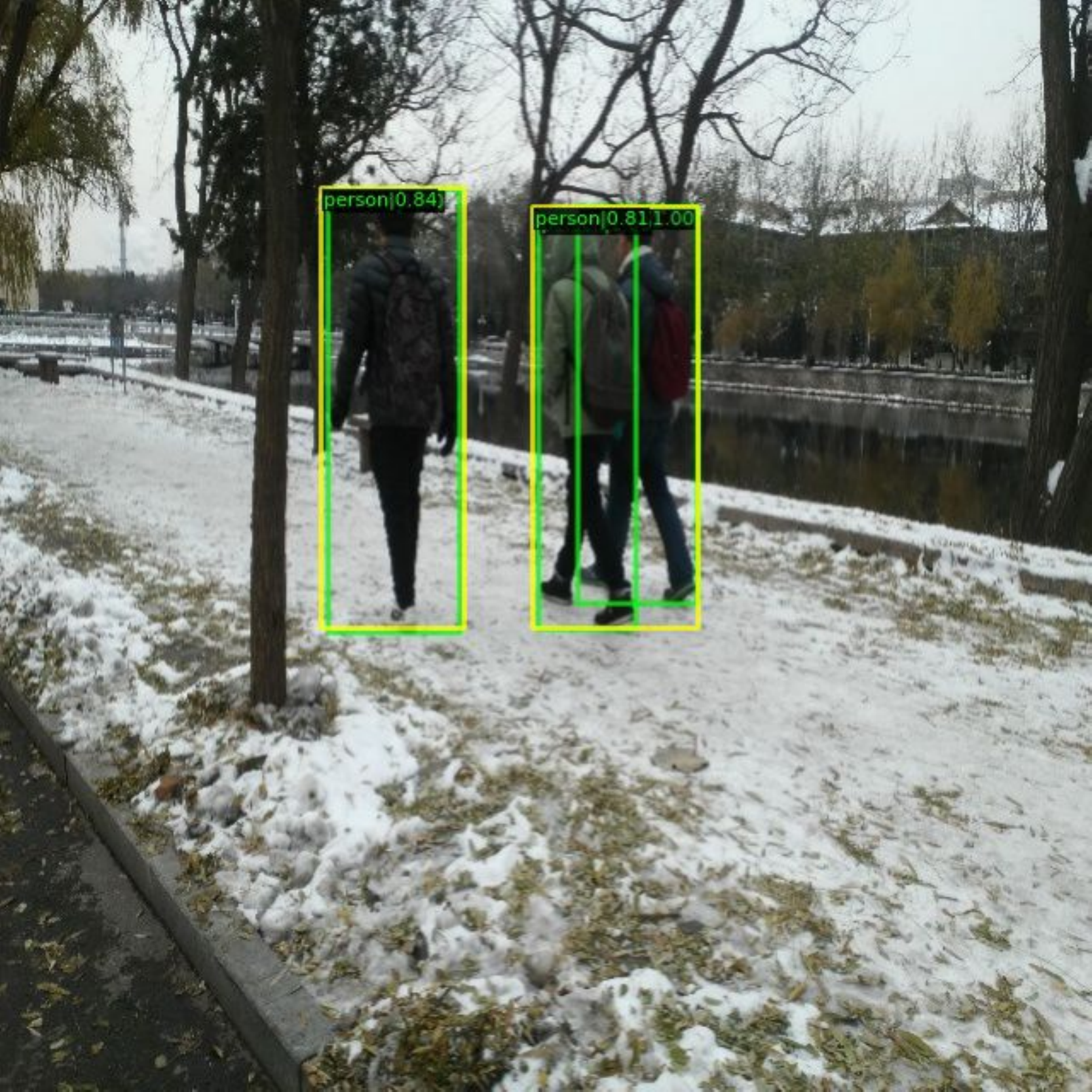}}\\
\caption{Visualization of miss detection.
The blue boxes represent the miss detection boxes, the green boxes represent the GT (ground truth), and the yellow boxes represent the correct prediction boxes.}
\label{vis_ms}
\end{figure}

As we all know, false positive detections affect the precision of object detection, and the miss detection will affect the recall of object detection.
Specially, we visualize the false positive detections and miss detection of the object detection results under AP50 in Fig. \ref{vis_fp} and Fig. \ref{vis_ms}, respectively.

\section{Experiments On Bigger Models}
Some bigger models, such as DeformableDETR (ResNet50, 39.8 parameters), achieve 44.50 mAP on COCO.
In comparison, FemtoDet (72.7k parameters) only achieves 12.30 mAP on COCO due to differences in settings.
FemtoDet is trained and tested on COCO with an input size of 416$\times$416, while DeformableDETR is trained and tested on COCO with an input size of 800$\times$1333.
To address this issue, we compared FemtoDet and DeformableDETR (ResNet50) on PASCAL VOC under the same conditions.
FemtoDet (8.3W on 3090Ti) achieved 46.3 AP50 / 22.0 mAP, while DeformableDETR (ResNet50, 134.0W on 3090Ti) achieved 70.7 AP50 / 26.8 mAP when tested on 416$\times$416 inputs.
We also applied the proposed IBE to other larger models and observed consistent improvement in performance.
For instance, MBV2 1.0 without IBE obtained 71.9 top1 Acc, while with IBE it achieved 72.2 top1 Acc on ImageNet.
Similarly, yolox nano (MBV2 1.0 as backbone) without IBE achieved 53.1 mAP, while with IBE it achieved 53.5 mAP on VOC.
These results demonstrate that FemtoDet achieves a reasonable tradeoff, and the IBE module consistently improves performance in larger models.

\section{Why not use "R RS HO R+HO A" metric to present the results on TJU-DHD?}
The missing rate ("R RS HO R+HO A") used on TJU-DHD mainly focuses on challenging examples, such as small objects (RS) and heavy occlusion objects (HO).
However, FemtoDet is better suited for detecting more accessible objects rather than these difficult examples.
Therefore, we have presented AP-related results in Table 6 of the regular paper.
    

{\small
\bibliographystyle{ieee_fullname}
\bibliography{egbib}
}

\end{document}